\documentclass[10pt, nonatbib]{journal} 
\usepackage[a4paper, total={5in, 8in}]{geometry}

\usepackage[utf8]{inputenc}
\usepackage{amsmath}
\usepackage{latexsym}

\usepackage{xcolor}
\usepackage{soul}
\usepackage{tabularx}
\usepackage[symbol]{footmisc}
\renewcommand{\thefootnote}{\fnsymbol{footnote}}
\usepackage[pdftex]{graphicx}
\usepackage{cite}
\usepackage[numbers,sort&compress]{natbib}
\newcommand{\shortcite}[1]{\cite{#1}}

\newcounter{savefootnote}
\usepackage{threeparttable}

\usepackage{enumitem}
\usepackage{ulem}
\usepackage{tikz}
\usepackage{booktabs}
\usepackage{etoolbox}
\newtoggle{final}
\toggletrue{final}
\newtoggle{newfinal}
\toggletrue{newfinal}

\definecolor{ferngreen}{rgb}{0.31, 0.47, 0.26}
\newcommand{\cg}[1]{\iftoggle{final}{#1}{{\color{ferngreen} #1}}}
\newcommand{\pw}[1]{\iftoggle{final}{#1}{{\color{blue} #1}}}
\newcommand{\npw}[1]{\iftoggle{newfinal}{#1}{{\color{blue} #1}}}
\newcommand{\ncg}[1]{\iftoggle{newfinal}{#1}{{\color{orange} #1}}}

\newcommand{\mz}[1]{\iftoggle{final}{#1}{{\color{olive} #1}}}

\usepackage{cleveref} 
\date{}
\begin{document}
\title{A Survey on Interpretable Reinforcement Learning}

\author{Claire Glanois \footnotemark[2] \quad 
       Paul Weng \footnotemark[2] \quad 
       Matthieu Zimmer \footnotemark[2]\\
       Dong Li \footnotemark[3] \quad 
       Tianpei Yang \footnotemark[4] \quad 
       Jianye Hao \footnotemark[3] \quad 
       Wulong Liu \footnotemark[3] \quad 
       }
       
\maketitle

\begin{abstract}
Although deep reinforcement learning has become a promising machine learning approach for sequential decision-making problems, it is still not mature enough for high-stake domains such as autonomous driving or medical applications. 
In such contexts, a learned policy needs for instance to be interpretable, so that it can \mz{be} inspected before any deployment (e.g., for safety and verifiability reasons). 
\npw{This survey provides an overview of various approaches to achieve higher interpretability in reinforcement learning (RL).
To that aim,}
\pw{we distinguish interpretability (as a property of a model) and explainability (as a post-hoc operation, with the intervention of a proxy) and discuss them in the context of RL with an emphasis on the former notion.
In particular, w}e argue that interpretable RL may \cg{embrace} different facets: interpretable inputs, interpretable (transition/reward) models, and interpretable decision-making.
Based on this \cg{scheme}, we summarize and analyze recent work related to interpretable RL with an emphasis on papers published in the past 10 years.
We also discuss briefly some related research areas and point to some potential promising research directions.
\end{abstract}

\tableofcontents

\footnotetext[2]{University of Michigan-Shanghai Jiao Tong University Joint Institute, China}
\footnotetext[3]{Huawei Noah's Ark Lab, China}
\footnotetext[4]{College of Intelligence and Computing (CIC), Tianjin University, China.}
\renewcommand*{\thefootnote}{\arabic{footnote}}
\setcounter{footnote}{0}
\section{Introduction}

Reinforcement learning (RL) \citep{sutton_reinforcement_2018} is a general machine learning framework for designing systems with automatic decision-making capabilities.
Research in RL has soared since its combination with deep learning, called deep RL (DRL), achieving several recent impressive successes (e.g., AlphaGo \citep{silver_mastering_2017}, video game \citep{vinyals_grandmaster_2019}, or robotics \citep{openai_solving_2019}).
These attainments were made possible notably thanks to the introduction of the powerful approximation capability of deep learning \cg{and its adoption} for sequential decision-making and adaptive control. 

However, this combination has simultaneously brought all the drawbacks of deep learning to RL. 
Indeed, as noticed by abundant recent work in DRL, policies learned using a DRL algorithm may suffer from various weaknesses, e.g.:
\begin{itemize}
    \item They are generally hard to understand because of the blackbox nature of deep neural network architectures \citep{zahavy_graying_2016}.
    
    \item They are difficult to train, require a \npw{large} amount of data, and DRL experiments are often difficult to replicate \citep{henderson_deep_2018}.
    
    \item They may overfit the training environment and may not generalize well to new situations \citep{zhang_study_2018}. 
    
    \item Consequently, they may be unsafe and \cg{vulnerable} to adversarial attacks \citep{huang_adversarial_2017}.
\end{itemize}
These observations reveal why DRL \npw{is} currently not ready for real-world high-stake applications such as autonomous driving or healthcare, and explain why interpretable and explainable RL has recently become a very active research direction. 

\cg{For most of real-world deployments of RL algorithms,} it is crucial that learned policies are \cg{intelligible} 
\pw{as they}
provide an answer (or a basis for an answer) to various concerns \npw{encompassing ethical, legal, operational, or} usability viewpoints:
\begin{description}[wide,nosep]
    \item[Ethical concerns] 
    When designing an autonomous system, \cg{it is essential to} ensure that its behavior follows some 
    ethical and fairness principles \cg{discussed and agreed upon beforehand by the stakeholders according to the context} \citep{crawford_ai_2016,yu_building_2018,leslie_understanding_2020,morley_what_2020,lo_piano_ethical_2020, dwork_fairness_2012,friedler_impossibility_2021}.

    \pw{The growing discussion about bias and fairness in machine learning \citep{mehrabi_survey_2019} suggests that mitigating measures must be taken in every aspect of an RL methodology as well.
    In this regard, intelligibility is essential to help assess the embedding of moral values into autonomous systems, and contextually evaluate and debate their \cg{equity and social impact}.}
    \item[Legal concerns]
    As autonomous systems start to be deployed, legal issues arise regarding notably safety \pw{\citep{amodei_concrete_2016}}, accountability \pw{\citep{doshi-velez_accountability_2019,european_commission_ethics_2019}}, or privacy \pw{\citep{horvitz_data_2015}}.
    For instance, fully-autonomous driving cars should be permitted in the streets only once proven safe with high confidence.
    \cg{The question of risk management \citep{bonnefon_trolley_2019} but also responsibility,

    in the case of an accident involving such systems, has become a more \npw{pressing} and complex problem.}
    Verification, accountability, but also privacy can only be ensured with more transparent systems.
    
    \item[Operational concerns]
    Since transparent systems are inspectable and verifiable, they can be examined before deployment to encourage that their decision-making is based on meaningful (ideally causal) relations and not on spurious features, ensuring higher reliability and increased robustness.
    \cg{From the vantage point of \pw{machine learning} researchers or engineers, such systems have the advantage of being more easily debugged and corrected}.
    Moreover, one may expect that such systems are easier to train, more data efficient, and more generalizable and transferable to new domains thanks to interpretability inductive biases.
    
    \item[Usability concerns]
    \npw{From an end-user's point of view, interpretable and explainable models can form an essential component for building more interactive systems, where a user can request more information about the outcome or the decision-making process.
    In particular, explainable systems would arguably be more trustworthy, which is a key requirement for their integration and acceptance \citep{mohseni_multidisciplinary_2020}. Although the question of trust touches on many other contextual and non-epistemic factors (e.g., risk aversion or goal) beyond intelligibility.}
\end{description}
\pw{In addition to this high-level list of concerns, we refer the interested reader to \cite{whittlestone_societal_2021} for a more thorough discussion about the potential societal impact of the deployment of deep RL-based systems.} 
\ncg{Although interpretability is a pertinent instrument \npw{to achieve} more accountable AI-systems, the debate around their real-life implementation should stay active, and include diverse expertise from legal, ethical, and socio-political fields, whose coverage goes beyond the scope of this survey.}


    
    
    
    
    
    

Motivated by \npw{the importance of these concerns}, the number of publications in DRL specifically tackling interpretability issues has increased significantly in recent years.
The surging popularity of this topic also explains the recent publication of three survey papers \citep{puiutta_explainable_2020,alharin_reinforcement_2020,heuillet_explainability_2021} on interpretable and explainable RL.
\npw{In \cite{puiutta_explainable_2020} and \cite{heuillet_explainability_2021}, a short overview is provided with a limited scope,} notably in terms of surveyed papers, while
\cite{alharin_reinforcement_2020} covers more studies, organized and categorized into explanation types.
The presentation of those surveys generally leans towards explainability \npw{as opposed to} interpretability (see \Cref{sec:interpretability} for the definitions adopted in this survey) and focuses on understanding the decision-making part of RL.

\npw{In contrast, t}his survey aims at providing a more comprehensive view of what \cg{may} constitute interpretable RL, which we here specifically distinguish from explainable RL (see \Cref{sec:interpretability}).
In particular, while decision-making is indeed an important aspect of RL, we believe that achieving interpretability in RL \cg{should involve} a more encompassing discussion of every \cg{component involved in these algorithms}, and should stand on three pillars:
interpretable inputs (e.g., percepts or other information provided to the agent), interpretable (transition/reward) models, and interpretable decision-making \pw{(policies or value functions)}.

\npw{Based on this observation, we organize previous work that proposes methods for achieving greater interpretability in RL, along those three components, with an emphasis on deep RL papers published in the last 10 years.}
Thus, in contrast to the previous three surveys, we cover additional work that belongs to interpretable RL such as relational RL or neuro-symbolic RL and also draw connections to other work that naturally falls in this designation, such as object-based RL, physics-based models, or logic-based task descriptions.
\pw{One goal of this proposal is to discuss the work in (deep) RL that is specifically identified as belonging to interpretable RL and to draw connections to previous work in RL that is related to interpretability.
Since such latter work covers a very broad research space, we can only provide a succinct account for it.}

The \pw{remaining of this} survey is organized as follows.
In the next section, we recall the necessary definitions and notions related to RL.
Next, we discuss the definition of interpretability (and explainability) in the larger context of artificial intelligence (AI) and machine learning (\Cref{sec:interpretability definition}), and apply it in the context of RL (\Cref{sec:interpretability RL}).
\cg{In the following sections, we present the studies related to interpretable inputs \pw{and} models in \Cref{sec:inputs} \pw{and} \Cref{sec:model} \pw{respectively}.
The work tackling interpretable decision-making, which constitutes the core part of this survey, is discussed in \Cref{sec:decision-making}.}
%
For \cg{the sake of} completeness, we also \cg{sketch a succinct review} of explainable RL in \Cref{sec:XRL}, \cg{although this area has been more thoroughly explored in the existing literature.}
Based on this overview, in \Cref{sec:open}, we provide a list of open problems and future research directions, which we deem\cg{ed particularly relevant.
Finally, we conclude in \Cref{sec:conclusion}}.

\section{Background} \label{sec:background}

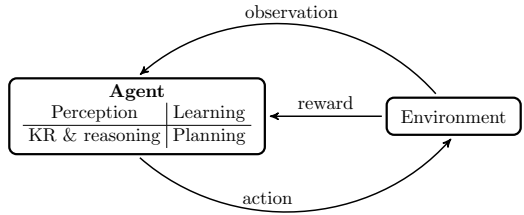
\begin{figure}[t]
\usetikzlibrary{arrows,positioning} 
\tikzset{
    >=stealth',
    punkt/.style={
           rectangle,
           rounded corners,
           draw=black, very thick,
           text width=6.5em,
           minimum height=2em,
           text centered},
    pil/.style={
           ->,
           thick,
           shorten <=2pt,
           shorten >=2pt,}
}
\centering
\scalebox{.7}{
\begin{tikzpicture}[node distance=1cm, auto,]
 \node[] (dummy) {};
 \node[punkt,left=of dummy,text width=13em] (agent) {\textbf{Agent}\\
 \begin{tabular}{c|c}
 Perception & Learning\\
 \hline
 KR \& reasoning & Planning
 \end{tabular}
 };
 \node[punkt,right=of dummy] (environment) {Environment}
   edge[pil,bend right=45] node[above] {observation} (agent.north) 
   edge[pil] node[above] {reward} (agent.east); 
   
 \path[] (agent.south) 
   edge[pil, bend right=45] node[auto] {action} (environment.south);
\end{tikzpicture}
}
\caption{Interaction loop in RL.}
\label{fig:RL}
\end{figure}

In reinforcement learning (RL), an agent interacts with an environment through an interaction loop \pw{(see Figure~\ref{fig:RL})}.
The agent repeatedly receives some observations from the environment, 
chooses an action, and receives some new observations and \cg{commonly} some immediate reward.
\pw{Although most RL methods solve this problem by considering the RL agent as reactive (i.e., given an observation, choose an action), \Cref{fig:RL} lists some other potential problems that an agent may tackle on top of decision-making: perception if the input is high-dimensional (e.g., an image), learning from past experience, knowledge representation (KR) and reasoning, and finally planning if the agent has a model of its environment.
}

This \pw{RL} problem is generally modeled as a Markov decision process (MDP) \pw{or one of its variants, notably partially observable MDP (POMDP)}\npw{\footnote{For the formal definitions of all those models, see~\cite{puterman94}.}}.
An MDP is defined as a tuple with a set of states, a set of actions, a transition function, and a reward function.
The sets of states and actions, which \cg{may} be finite, infinite, or even continuous, specify respectively the possible world configurations for the agent and the possible \cg{response} that it can perform.
In a partially observable MDP, the agent does not observe the state directly, but has access to an observation that probabilistically depends on the hidden state.
The difficulty in RL is that the transition and reward functions are not known \cg{to the agent}.
The goal of the agent is to learn to choose actions (i.e., encoded in a policy) such that it maximizes its expected (discounted) sum of rewards (i.e., represented by a \textit{value function}).
\pw{A policy may choose actions based on states or observations in a deterministic or randomized way.
In RL, the value function often takes the form of a so-called Q-function, which measures the value of an action followed by a policy in a state.}
To solve this RL problem, model-based and model-free algorithms have been proposed, depending \mz{on whether} a model of the environment (i.e., transition/reward model) is \cg{explicitly} learned or not.

The success of deep RL is explained partly by the use of neural networks to approximate value functions or policies, but also by \cg{various} algorithmic progress. 
Deep RL algorithms can be categorized in two main categories: value-based methods and policy gradient methods, in particular in their actor-critic version. 
For the first category, the model-free methods are usually variations of the DQN algorithm \citep{mnih_human-level_2015}. 
For the second one, the current state-of-the-art model-free methods are PPO \citep{schulman_proximal_2017} for learning a stochastic policy, TD3 \citep{fujimoto_addressing_2018} for learning a deterministic policy, and SAC \citep{haarnoja_soft_2018} for entropy-regularized learning of a stochastic policy.
Model-based methods can span from simple approaches such as first learning a model and then applying a model-free algorithm \npw{using the learned model as a simulator}, to more sophisticated methods that \cg{leverage} the learned model to accelerate solving an RL problem \pw{\shortcite{francois-lavet_combined_2019,veerapaneni_entity_2020,scholz_physics-based_2014}}.

\pw{
For complex decision-making tasks, hierarchical RL (HRL) \citep{barto_recent_2003} has been proposed to exploit temporal and hierarchical abstractions, which may facilitate learning and transfer, but also promote intelligibility.
Although various architectures have been proposed, decisions in HRL are usually made at (at least) two levels.
In the most popular framework, a higher-level controller (also called meta-controller) chooses temporally-extended macro-actions (also called options), while a lower-level controller chooses the primitive actions.
Intuitively, an option can be understood as a policy with some starting and ending conditions.
When it is known, it directly corresponds to the policy applied by the lower-level controller.
An option can also be interpreted as a subgoal chosen by the meta-controller for the lower-level controller to reach.
}

\section{Interpretability and Explainability} \label{sec:interpretability}

In this section, we first discuss the definition of interpretability and explainability \pw{as proposed} in the explainable artificial intelligence (XAI) literature. 
Then, we focus on the instantiations of those notions in reinforcement learning.

\subsection{Definitions} \label{sec:interpretability definition}

Various terms have been used in the literature to qualify the capacity of a model to make itself understandable, such as interpretability, explainability, intelligibility, comprehensibility, transparency, or understandability.
Since no consensus about the nomenclature in XAI has been reached yet, they are not always distinguished and are sometimes used interchangeably in past work or surveys on XAI.
Indeed, interpretability and explainability are for instance often used as synonyms \citep{ribeiro_model-agnostic_2016,miller_explanation_2019,molnar_interpretable_2019}.
For better clarity and specificity, in this survey, we only employ the two most common terms, \textit{interpretability} and \textit{explainability}, and clearly distinguish those two notions, which we define below.
\pw{This distinction allows us to provide a clearer view of the different work in interpretable and explainable RL.}
Moreover, we use \textit{intelligibility} as a generic designation that encompasses those two notions.
For a more thorough discussion of the terminology in the larger context of machine learning and artificial intelligence, we refer the interested readers to surveys on XAI 
\citep{lipton_mythos_2017,barredo_arrieta_explainable_2020,chari_directions_2020,gilpin_explaining_2019}. 

Following \cite{barredo_arrieta_explainable_2020}, we \npw{simply} understand interpretability as a passive quality of a model, while explainability here refers to an active notion that corresponds to any external, usually post-hoc, \pw{methodology} or proxy aiming at providing insights into the \cg{working and decisions} of a \pw{trained} model, \npw{although both notions are epistemologically inseparable from both the observer and the context}. 

%
While the former approach is achieved by resorting to \cg{more transparent} models,
the latter \cg{approach} is carrying out additional processing steps \cg{to explicitly provide a kind of explanation aiming} to clarify, \cg{justify, or rationalize the decisions of} a trained black-box model.
At first sight, it seems that \pw{interpretability} is involving an \textit{objectual and mechanistic}  understanding of the model, whereas explainability mostly restrict\pw{s} itself to a more \textit{functional}\footnote{A functional understanding ``relies on an appreciation for functions, goals, and purpose'' while a mechanistic understanding ``relies on an appreciation of parts, processes, and proximate causal mechanisms'' \citep{paez_pragmatic_2019}.}---and often model-agnostic---understanding of the outcomes of a model.
\pw{Yet}, as advocated by \cite{paez_pragmatic_2019}, post-hoc \pw{intelligibility} in AI should require some degree of objectual understanding\footnote{Some objectual understanding is particularly beneficial when considering legal accountability and public responsibility.} of the model, since a thorough understanding of a model's decisions, also encompasses the ability to think counterfactually (``What if...'') and contrastively (``How could I alter the data to get outcome X?'').
\pw{Note that both notions, interpretability and explainability, may depend on the observer. 
Indeed, what is intelligible and what constitutes a good explanation may be completely different for an end-user, a system designer (AI engineer or researcher), or a legislator.
Except in our discussion on explainable RL, we will generally take the point of view of a system designer.}


Since the main focus of this review is  interpretability\footnote{One may argue that model-based interpretability is more desirable than post-hoc explainability \cite{rudin_stop_2019}.}, we further clarify this notion \pw{by recalling three potential definitions as proposed by \cite{lipton_mythos_2017}:}
simulatability, decomposability, and algorithmic transparency.

A model is \textit{simulatable} if its inner working can be simulated \pw{by} a human.
Examples of simulatable models are small linear models or decision trees.
\cg{The concept of simplicity, and quantitative aspects, consequently underlie any definition of simulatability.}
\pw{In that sense, a hypothesis class is not inherently interpretable with respect to simulatability.
Indeed, a decision tree may not be simulatable if its depth is huge, whereas a neural network may be simulatable if it has only a few hidden nodes.}
%
%
%
A model is \textit{decomposable} if each of its parts (input, parameter, and calculation) can be understood intuitively.
Since a decomposable model assumes its inputs to be intelligible, any simple model based on complex highly-engineered features is not decomposable.
Examples of decomposable models are linear models or decision trees using interpretable features.

\pw{While the other two definitions focuses on the model, the third one shifts the attention to the learning process and requires it to be intelligible.
Thus, a}n algorithm is \textit{transparent} if its properties are well-understood (e.g., convergence).
\pw{In that sense, standard learning methods for linear regression or support vector machine may be considered transparent.
However,} since the training of deep learning models is currently still not very well-understood, \cg{it results in a regrettable lack of transparency of these algorithms.}

\begin{figure*}[t]
\centering
\includegraphics[trim=0 125pt 0 0, clip,width=.9\textwidth]{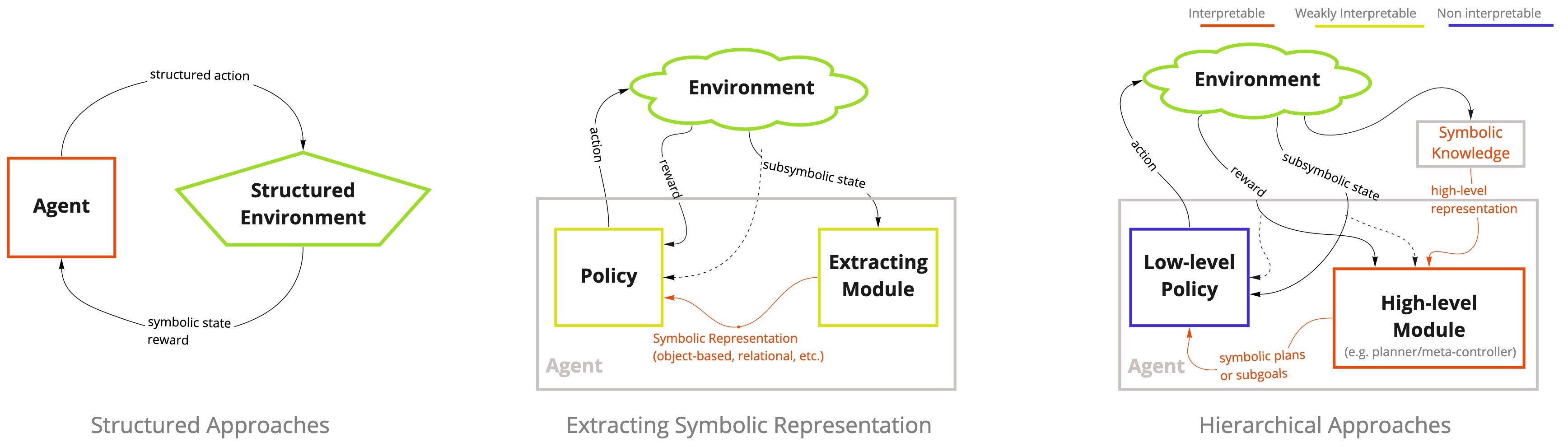}
\caption{Illustrations of the different approaches for interpretable inputs: (Left) Structured Approach, (Center) Extracting Symbolic Representation, (Right) Hierarchical Approach. 
Dashed lines represent optional links, depending on the methods.}
\label{fig:interpretable inputs}
\end{figure*}

\pw{

Although algorithmic transparency is important, the most relevant notions for this survey are the first two since our main focus is the intelligibility of trained models.
More generally, it would be useful and interesting to try to provide finer definitions of those notions, however this is out of the scope of this paper, which aims at providing an overview of work related to interpretable (deep) RL.}



\subsection{Interpretability in RL} \label{sec:interpretability RL}

Based on the previous discussion of interpretability in the larger context of artificial intelligence, we now turn to the reinforcement learning (RL) setting.
To solve an RL problem, the agent may need to solve different AI tasks (notably perception, knowledge representation, reasoning, learning, planning) depending on the assumptions made about the environment and the capability of the agent.


For this whole process to be interpretable, all its components have arguably to be intelligible, such as:
(1) the inputs (e.g., observations or any other information the agent may receive) and its processing, 
(2) the transition and preference models, and 
(3) the decision-making model (e.g., policy and value functions).
The preference model describes which actions or policies are preferred. 
It can simply be based on the usual reward function, but can also take more abstract forms such as logic programs.
\npw{Note that making those components more intelligible to us supposes a certain disentanglement of the underlying factors, and entails a certain representation structure. 
With this consideration, this survey can be understood as discussing methods to achieve \textit{structure} in RL, which consequently enhances interpretability.}

The three \pw{definitions} of \pw{interpretability} (i.e., simulatability, decomposability, and algorithmic transparency) discussed previously can be applied in the RL setting.
For instance, for an RL model to be simulatable, it has to involve simple inputs, simple preference (possibly also transition) models, and simple decision-making procedures, which may be hard to achieve in practical RL problems.
\cg{In this regard}, applying those definitions of \pw{interpretability} at the global level in RL does not lead to any interesting insights in our opinion.
However, because RL is based on different components, \cg{it may be judicious to apply} the \pw{different definitions} of \pw{interpretability} to them, possibly in a differing way.
A \cg{more modular} view provides a more revealing analysis framework to understand previous and current work related to interpretable RL.
Thus, an RL approach can be categorized for instance, as based on a non-interpretable input model, a simulatable reward and decision-making models \citep{penkov_learning_2019} or as based on simple inputs, a simulatable reward model, and decomposable transition and decision-making models \citep{degris_learning_2006}.

Next, we overview and discuss RL approaches that are related to or focus on interpretable inputs \pw{(\Cref{sec:inputs})}, interpretable transition and preference models \pw{(\Cref{sec:model})}, and interpretable decision-making \pw{(\Cref{sec:decision-making})}. 
\pw{We argue that those categories are essential aspects of interpretable RL since they each pave the way towards higher interpretability in RL.
As in any classification attempt, the boundary between the different clusters is not completely sharp.
Indeed, some propositions could arguably belong to several categories.
However, to avoid repetition, we generally discuss them only once with respect to their most salient contributions.}











\section{Interpretable Inputs} \label{sec:inputs}

\pw{A first step towards interpretable RL regards the inputs that an RL agent uses to learn and make its decisions.
It includes the agent's observations, but also other information that the agent may have at its disposal, such as relational structure or hierarchical structure of the problem.
Arguably, these inputs must be intelligible if one wants to understand the decision-making process later on.

Furnishing extra-structure and being typically lower-dimensional, interpretable inputs can result in faster learning, better generalizability and transferability but also be smoothly integrated with reasoning and planning.

\pw{Diverse approaches have been investigated to provide interpretable inputs to the agent (see \Cref{fig:interpretable inputs}).}
They may be pre-given as seen in the literature of structured RL (\Cref{sec:structured RL}), or may need to be extracted from high-dimensional observations (\Cref{sec:extracting symbolic}).
Tangentially, additional interpretable knowledge can be provided to help the RL agent, in addition to the observations (\Cref{sec:symbolic knowledge}).
}


\subsection{Structured \cg{Approaches}} \label{sec:structured RL}

The literature explored in this \pw{sub}section assume\pw{s} a pre-given \pw{structured} representation of the environment which may be modelled through a collection of objects, and their relations as in \textit{object-\pw{oriented}} or \textit{relational} RL (RRL, \cite{dzeroski_relational_1998}).
\pw{Thus}, the MDP is assumed to be structured and the problem is solved within that structure:
task, reward, state transition, policies and value function are defined over objects and their interactions\pw{---}e.g. using first-order logic (FOL) \cite{barwise_introduction_1977} as in RRL.
\cg{A non-exhaustive overview of this work is provided in \Cref{tab:structured_RL}.}


\cg{A first delicate question, tied to knowledge representation \citep{swain_knowledge_2013}, is the choice of the specific structured representations for the different elements (e.g., state, transition, rewards, value function, policy).} 
\ncg{For instance, a first step in this literature was to depart from \textit{propositional representation}, and turn towards relational representations}, which not only seems to better fit the way we reason about the environment---in terms of objects and relations
---but may bring other benefits, \pw{such} as the smooth incorporation of logical background knowledge.
Indeed, in propositional representations, the number of object\mz{s} is fixed, all relations have to be grounded\pw{---a computationally heavy operation---}but most importantly it is not \pw{suitable} to generalize over object\mz{s} and relations
, and is unable \pw{to} capture the structural aspect of the domain
\cg{(e.g., Blocks World \shortcite{slaney_blocks_2001}).}
\ncg{Several variations of this structure are presented below.}

\begin{table}[t]
\footnotesize
    \centering
    \caption{Overview of approaches for Structured Approaches}
    \label{tab:structured_RL}
    
    \begin{tabular}{@{}llll@{}}
    \toprule

        Relational Representations & Approach & Interpretability & Refs \\
    \midrule


     
       Relational, & Linear Programming & Simulatable\footnote{}\setcounter{savefootnote}{\value{footnote}} &  \citep{guestrin_generalizing_2003} 
       \\
       Q linear approx.   & & 
       & \\
       
       Relational, & Q-learning  & Simulatable  & \citep{dzeroski_relational_2001}\\
       FOL Q$-$decision tree  & &  & \\
       Relational, hierarchical, & Q-learning &Simulatable  & \citep{driessens_learning_2001}\\
        FOL Q-decision tree  &  & 
        & \\

        Relational, & Q-learning & Decomposable  & \citep{cole_symbolic_2003}\\
       HOL Q$-$decision tree  &  & 
       & \\
       
     Relational, feature-based,   & Q-learning & Partial decomp. & \citep{walker_relational_2004}\\
      Q linear approx. &  & & \\
        
    Relational, feature-based, & Q-learning & Partial decomp.  & \citep{sanner_simultaneous_2005} \\
    Q w/ rel. Naive Bayes Net &  & & \\
    
    Relational, & Q-learning w/ & Partial decomp.  & \citep{gartner_graph_2006} \\
    Q w/ graph kernels & Gaussian processes & & \\
    
    Relational & Actor-Critic & Partial decomp.  & \npw{\citep{garg_symbolic_2020,janisch_symbolic_2021}} \\
    GraphNN State rep.,  &  & & \\
    Neural policy/value &  & & \\
 
    
      
         
    \bottomrule
    \end{tabular}
\end{table}
\footnotetext[\value{savefootnote}]{Simulatability holds assuming small domains and also implies decomposability here.\stepcounter{savefootnote}
}

\paragraph{Relational MDP}
\cg{Relational Markov Decision Process (RMDPs) \citep{guestrin_generalizing_2003} are first-order representation of factored MDP  \citep{boutilier_stochastic_2000}, expressed via objects, predicates and functions, and based on probabilistic relational model (PRM, \cite{koller_probabilistic_1999})
\footnote{PRMs may be understood as ``relational'' extensions of ``propositional'' Bayesian networks. 
}.
The representation involves different classes of objects (over which binary relations are defined), each having attributes attached to a specific domain.
Transition and reward models are assumed given, e.g. as a dynamic Bayesian network (DBN, \cite{dean_model_1990}) 
, although the specific representational language may vary\footnote{Relational Dynamic Influence Diagram Language (RDDL, \shortcite{sanner_relational_2011}), extended DBN using state-dependent rewards aggregated over objects are able to model parallel effects. 
In contrast, Probabilistic Planning Domain Definition Language (PPDDL) \citep{younes_ppddl10_2004} employs action-transition-based rewards and models correlated effect. Noe that \cite{guestrin_generalizing_2003} assumes static representations, which are unfit for real-world dynamics or relational environments such as Blocks World.}.
Closely related, Object Oriented-MDPs (see \shortcite{diuk_object-oriented_2008}, presented in \Cref{sec:transition model})---later extended to deictic representations \citep{marom_zero-shot_2018}---use similar state-representation yet differ in \pw{the way their} transition dynamics are described: transitions are assumed deterministic and learned within a specific propositional form in the first step of the\npw{ir proposed} algorithm.
}

\paragraph{Relational RL}
\npw{Following the initial work on Relational RL  \cite{dzeroski_relational_1998}, a consequent line of work summarized below extends previous work dealing with MDPs modelled in a relational language to the learning setting, at the crossroad of RL and logical machine learning---such as inductive logic programming (ILP) and probabilistic logic learning.
}


\npw{In \cite{dzeroski_relational_2001}, the Q-function is learned with a relational regression tree using Q-learning  extended to situations where states, actions, and policies are represented using first-order logic.%
}
\cg{\npw{However,} explicitly representing value functions in relational learning is difficult, partly due to \textit{concept drift} \citep{otterlo_survey_2005}, \npw{which occurs} since the policy providing examples for the Q-function is being constantly updated.
It may motivate to turn towards policy learning (as \shortcite{dzeroski_relational_1998} relying on P-trees), and employ approximate policy iteration methods which would keep explicit representation of the policy but not the value function, for larger probabilistic domains.}

\cg{Diverse extensions of relational MDPs and of Relational RL (RRL) have been proposed, either exact or approximate methods, in model-free and in model-based, with more or less expressive representations and within a plain or more hierarchical approach \citep{driessens_learning_2001}.
Regarding the representations, previous model-free RRL work is based on explicit logical representation such as logical (FOL or more rarely Higher Order Logic (HOL, e.g., \shortcite{cole_symbolic_2003})) regression trees, which, in a top-down way, recursively partition the state space; in contrast, other bottom-up and feature-based approaches \citep{walker_relational_2004,sanner_simultaneous_2005} aim to learn useful relational features which they would combine to estimate the value function, either by feeding them to a regression algorithm \citep{walker_relational_2004}, or into a relational naive Bayes network \citep{sanner_simultaneous_2005}. Other alternatives to regression trees have been implemented such as through Gaussian processes---incrementally learnable Bayesian regression---with graph kernels, defined over a set of state and action \citep{gartner_graph_2006}.
Finally, other work in quest of more expressivity turns towards neural representations. 
For instance, after extracting a graph instance from a given Relational Dynamic Influence Diagram Language (RDDL, \shortcite{sanner_relational_2011}), \cite{garg_symbolic_2020} computes nodes embedding via graph propagation steps, which are then fed to value and policy decoders (\pw{multi-layer perceptrons,} MLP) attached to each action symbol.
\npw{In \cite{janisch_symbolic_2021}, graph neural networks are similarly used} to build a relational state representation in relational problems. 
The authors resort to auto-regressive policy decomposition \citep{vinyals_starcraft_2017} to tackle multi-parameter actions (attached to unary or binary predicates).%
}

As work within RRL extends beyond the scope of this survey, we refer the curious reader to the generous overview in \cite{otterlo_book_2009} or \shortcite{otterlo_wiering_2012}, Chapter $8$.
Let us point out that despite the ``reinforcement'' appellation, a significant proportion of work in RRL assumes that environment models (transitions and reward structures) are known to the agent, which may be unrealistic.
RRL ha\mz{s} also been applied to diverse domains, such as for efficient exploration within robotics \citep{martinez_relational_2017}.



\paragraph{Conclusion}
In structured \pw{MDP} 
and \pw{relational RL}, by borrowing \pw{from} symbolic reasoning, most work leads to agents that can learn and reason about objects. 
Such an explicit, logical representation of learned structures may help both generalization or transfer learning efficiently and robustly in similar representational frameworks---e.g.\pw{,} \textit{combinatorial generalization}. 
However, some major drawbacks are that these approaches necessitate the symbolic representation to be hand-designed, and often rely on non-differentiable operations. 
They are therefore not very flexible over framework variations (e.g., task or input) and not well suited for more complex tasks, or noisy real-life environments. 

\subsection{Learning Symbolic Representations} \label{sec:extracting symbolic}

When input\pw{s} are given as high-dimensional raw data, it seems judicious---although challenging---to extract explicit symbolic representations on which we can arguably reason and plan in a more efficient and intelligible way. \cg{This process of abstraction, which is very familiar to \pw{human cognition}\footnote{Physical theories are a typical example of this practice, where laws---\pw{such} as laws of motions---are reused across instantiations and scenes with various primitive entities.}, \npw{reduces} the complexity of an environment to low dimensional, discrete, abstract features.
By abstracting away lower-level details and irrelevant variations, this paradigm brings undeniable advantage and could greatly leverage the learning and generalization abilities of the agent. 
Moreover, it provides the possibility of reusing high-level features through environments, space and time.}

Some promising new research directions---tackling the key problem of symbol grounding \pw{\citep{harnad_symbol_1990}}---are adopting an end-to-end training,
therefore tying the semiotic emergence not only to control but also to efficient high-level planning \citep{konidaris_constructing_2014,konidaris_symbol_2015,andersen_active_2017}, or model-based learning \citep{francois-lavet_combined_2019}, 
to encourage more meaningful abstractions. 
\cg{Work presented below \pw{(see \Cref{tab:extracting_symbolic})} ranges from extracting symbols to relational representations\npw{, which are in turn used for control or planning}. 
In the next two sections, we distinguish actual \textit{\pw{high-level (HL)} decomposability}---meaning the HL module is decomposable—--from \textit{HL modularity}, which denotes the gain in interpretability brought by the task decomposition, which may be seen as a partial high-level decomposability.}


\paragraph{Symbol grounding}

Some previous approaches \pw{\citep{konidaris_constructing_2014,konidaris_symbol_2015,andersen_active_2017,konidaris_skills_2018}} have tackled the problem \mz{of learning} symbolic representations adapted for high-level planning from raw data.  
As they are concerned about evaluating the feasibility and success probability of a high-level plan, they only need to construct symbols
both for the initiation set and the termination set of each successive option. 
There, the state-variables are gathered into factors, which can be seen as sub-goals, and are tied to a set of symbols; through unsupervised clustering, each option is attached to a partition of the symbolic state; it leads to a \npw{probabilisitc} distribution over symbolic options \citep{sutton_between_1999} which guides the higher-level policy to evaluate the plan.

\begin{table}[t]
\footnotesize
    \centering
    \caption{Overview of approaches for Learning Symbolic Representation}
    \label{tab:extracting_symbolic}
    \begin{tabular}{@{}llll@{}}
    
    \toprule

    Representations & Approach & Interpretability & Refs \\
    \midrule


Probabilistic symbols &  Unsupervised clustering  & HL \pw{modular} & \citep{konidaris_symbol_2015}\\
for planning &(e.g., DBSCAN algorithm) && \citep{konidaris_skills_2018}\\

Probabilistic symbols & Unsupervised clustering  &HL \pw{modular} &  \citep{andersen_active_2017}\\
for planning & (Bayesian hierarchical)&&\\


Symbols as classifiers  & Human-teaching &HL \pw{modular} &  \citep{kulick_active_2013}\\

Symbols as classifiers & Program-guided & HL \pw{modular} & \citep{penkov_learning_2019}\\

Symbols as classifiers & Program-guided AE &  HL \pw{modular} & \citep{sun_program_2020}\\

Objects & Object recognition,  & Partial 
\pw{decomp.} & \citep{li_object-sensitive_2017}\\
&w/ template matching&&\\

Objects, relational & Unsupervised object extraction & Partial decomp. & \citep{garnelo_towards_2016}  \\
 & w/ activation spectrum &&\\

Objects & Unsupervised video &  - & \citep{goel_unsupervised_2018} \\
& segmentation w/ optical flow&&\\

Relational & Relational MLP-modules & Weak & \citep{adjodah_symbolic_2018}\\

Objects & CNN w/ attention  & Weak & \citep{zambaldi_deep_2019}\\

    \bottomrule
    \end{tabular}
\end{table}

Instead of random or greedy exploration, \cite{andersen_active_2017} proposes an online active exploration algorithm, with a learned Bayesian symbolic model guiding the exploration.

In a different direction, some researchers have involved human teaching \citep{kulick_active_2013} or programs \citep{penkov_learning_2019,sun_program_2020} to guide the learning of symbols.
For instance, in the work by \cite{penkov_learning_2019}, the pre-given program mapping the perceived symbols (from an auto-encoder network) to actions, imposes semantic priors over the learned representations, and may be seen as a regularization which structures the latent space. Meanwhile \cite{sun_program_2020} presents a perception module which aim\cg{s} to answer the conditional queries (``if'') within the program, which accordingly provides a symbolic goal to the low-level controller.

However, these studies, by relying on a human-designed program, or human-teaching, partly bypass the problem of autonomous and enacted symbol extraction.


\paragraph{Object-Recognition}

When the raw input is given as \mz{an} image, diverse techniques within computer vision and within Object Recognition (OR) or Instance Segmentation (combining semantic segmentation and object localization) are beneficial to extract a symbolic representation.
Such extracted information, fed as input \pw{to} the policy or Q-network, should arguably lead to more interpretable networks.
OR aims specifically to find and identify objects in an image or video sequence, despite possible changes in sizes, scales or obstruction when objects are being moved; it ranges from classical techniques -- such as template matching \cite{brunelli_template_2009}, or Viola-Jones algorithm \cite{viola_robust_2001}-- to more advanced ones.

As an example of recent work within RL learning symbolic representation, O-DRL \cg{\citep{li_object-sensitive_2017,iyer_transparency_2018}}
 can exploit and incorporate object characteristics such as the presence and positions of game objects, which are extracted with template matching before being fed into the Q-network.
In the case of moving rigid objects, techniques have been developed to exploit information from object movement to further improve object recognition.
For instance, \cite{goel_unsupervised_2018} first autonomously detects moving objects by exploiting structure from motion, and then uses this information for action selection. 

Aiming to delineate a general end-to-end reinforcement learning framework, \npw{Deep Symbolic RL (DSRL) \cite{garnelo_towards_2016} combines} a symbolic front end with a neural back end learning to map high-dimensional raw sensor data into a symbolic representation in a lower-dimensional conceptual space.
\cg{Such proposition may be understood within emblematic neuro-symbolic approaches' scheme (as presented in \shortcite{bader_dimensions_2005}, Fig.4), where a symbolic system and a connectionist system share information back and forth.}
The authors also reflect on a few key notions for an ideal implementation such as \textit{conceptual abstraction} (e.g., how to detect high level similarity), \textit{compositional structure},
\textit{common sense priors} or \textit{causal reasoning}.
However, their first prototype proposal is relatively limited, with a symbolic front end carrying out very little high-level reasoning, and a simple neural back end for unsupervised symbol extraction.
Further work has questioned the generalization abilities of DSRL \citep{dutra_comparison_2017} or aimed to incorporate common sense within DSRL \citep{davila_garcez_towards_2018} to improve learning efficiency and accuracy, albeit \cg{still within} quite restricted settings.

\paragraph{Relational Representations}

Within the \pw{goal of having} more interpretable inputs for the policy or Q-value network, some work aims \pw{at computing} specifically relation-centric representations, e.g., graph-based representation. 
Such relational representation can be leveraged for decision-making, e.g. once fed to the value or policy network or even in a hierarchical setting.
Within this line of research, graph networks \cite{battaglia_relational_2018} 
stand out as an effective way to compute interactions between entities and can support combinatorial generalization to some extent \citep{scarselli_graph_2009,gilmer_neural_2017,cranmer_discovering_2020,sanchez-gonzalez_graph_2018,li_gated_2017,battaglia_relational_2018}.
Roughly, the inference procedure is a form of propagation process similar to a message passing system. 
Having high capacity, graph networks have been thoroughly exploited in a diverse range of problem domains, either for supervised, unsupervised or in model-free or model-based RL, for tasks ranging from visual scene understanding, to physical systems dynamics via chemical molecule properties, image segmentation, point clouds data, combinatorial optimization, or dynamic of multi-agent systems. 


Aiming to represent relations between objects, relational modules have been designed to inform the Q-network \citep{adjodah_symbolic_2018}, and/or the policy network \citep{zambaldi_deep_2019}. In a work by \cite{zambaldi_deep_2019}, the pairwise interactions are computed via a self-attention mechanism \citep{vaswani_attention_2017}, and used to update each entity representation which---\pw{according to the authors'} claim---is led to reflect important structure about the problem and \pw{the} agent's intention.
Unlike most prior work in relational inductive bias (e.g., \shortcite{wang_nervenet_2018}), it does not rely on a priori knowledge of the problem and the relations, yet is hard to scale to large input space, suffering from quadratic complexity.
Other relational networks modules previously developed could be easily incorporated into any RL framework; e.g., \cite{santoro_simple_2017,chang_compositional_2017} which aim to factorize dynamics of physical systems into pairwise interactions.

\paragraph{Conclusion}
Extracting symbolic and relational representations from high dimensional raw data is crucial as it would avoid the need of hand-designing the symbolic domain, and could therefore unlock enhanced adaptability \pw{when facing} new environments.
Indeed, the symbolic way of representing the environment inherently benefit\pw{s} from \pw{its} compositional and modular perspective, and enable\pw{s} the complexity of the state-space \pw{to be reduced} thanks to abstraction.
Moreover, most of the modules presented in this section may be conveniently incorporated as a preprocessing step for any object-oriented or relational RL, either being trained beforehand or more smoothly end-to-end with the subsequent model.

Nevertheless, despite a certain history of work in this area, in the absence of pre-given hand-crafted schemes,
it remains a consequent challenge for an agent to autonomously extract relevant abstractions model from a high-dimensional continuous complex and noisy environment.
Advanced techniques currently in used in computer vision could be leveraged in the context of reinforcement learning, notably in Multiple Object Tracking (e.g., YOLO or Deep Sort \shortcite{mandal_object_2020}) but also to extract structured representation out of visual scenes (e.g., scene graphs \shortcite{bear_learning_2020}).

For scene interpretation, going beyond the traditional spatial or subsumption (''part-of'') relations, some logic-based approaches have emerged, using Description Logics (DL)---or fuzzy DL to handle \pw{uncertainty}---in order to derive new facts in the scene, given basic components (object type or spatial relation), e.g., redefining labels.
For instance, \cite{donadello_logic_2017} \npw{extending} \cite{serafini_logic_2016} employs a FOL interpreted in the real numbers.
\cg{Another idea would be to first learn---ideally causally---\textit{disentangled} subsymbolic representations from low-level data (e.g. \cite{higgins_towards_2018}), to bootstrap the subsequent learning of higher-level symbolic representations, on which more logical and reasoning-based frameworks can then be deployed.}

\npw{As a side note}, this line of work touches sensitive questions on how symbols acquire their meanings; e.g. with the well-known \textit{symbol grounding problem}, inquirying on how to ground the representations and symbolic entities from raw observations\footnote{A common assumption in contemporary cognitive science is that these representations have to emerge in strong dependency to the actions and goals of the agent (enacted), and the environment (situated).
}. On top of the challenges of ``when'' and ``how'' to invent a new symbol, enters also the question of how to assess of its quality.

\subsection{Hierarchical Approaches} \label{sec:symbolic knowledge}

\cg{Instead of working entirely within a symbolic structure as in \Cref{sec:structured RL}, many researchers have tried to incorporate elements of symbolic knowledge with sub-symbolic components, with notable examples within hierarchical RL (HRL, cf. \cite{Hengst2010}). \ncg{Their aim was notably to leverage both symbolic and neural worlds, with a more structured high-level and a more flexible low-level.}
\Cref{tab:hierarchical} introduces these approaches, focusing notably on their high-level interpretability, as their lower-level components rarely claim to be interpretable.
}
Indeed, different levels of temporal and hierarchical abstractions within human-decision-\pw{making} arguably participate to make it more intelligible. For instance, \cite{beyret_dot--dot_2019} demonstrates how a meta-controller providing subgoals to a controller leverage both performance and interpretability for robotic tasks.


\begin{table}[t]
\footnotesize
    \centering
    \begin{threeparttable}
    \caption{Overview of approaches for Hierarchical RL}
    \label{tab:hierarchical}
    \begin{tabular}{@{}llll@{}}
   
    \toprule

   Symbolic Knowledge & Learned Components  & HL Interpretability & Refs\\
    \midrule

\pw{Subgoals} & Controllers (LL, HL)$^{\star}$ & Modular &\citep{beyret_dot--dot_2019}\\

HL Domain & Controller, HL RL Agent & Partial Decomp. & \citep{sridharan_reba_2019}\\

Symbolic Plans & Full \& Subpolicies & Modular  &   \citep{andreas_modular_2017}\\

(STG)\footnote{}\setcounter{savefootnote}{\value{footnote}} & Selector, Subpolicies   & Modular & \citep{shu_hierarchical_2018}\\

Model Primitive& Gate, Subpolicies   & Modular &\citep{wu_model_2019}\\

HL domain, SP$^\ddag$ &  RL Agent & Simulatable  & \citep{leonetti_synthesis_2016}\\
 
HL domain, SP &  RL Agent, SP & Simulatable  & \citep{yang_peorl_2018}\\

HL domain, SP & Controllers (LL, HL)$^{\star}$, & Partial. Decomp. & \citep{jiang_integrating_2018}\\
&Task  \& Motion Planner&&\\

HL domain, SP & Controllers (LL, HL)$^{\star}$, SP& Partial. Decomp. & \citep{lyu_sdrl_2019}\\

HL domain & Subgoal FSA$^\dag$, RL Agent  &  Simulatable &\citep{furelos-blanco_induction_2021}\\

HL domain, FSA$^\dag$ & FSA-guided RL Agent &Simulatable &\citep{li_formal_2019}\\
 
HL filtering rules & RL Agent  & Partial decomp. &\citep{zhang_faster_2019}\\
HL rules & Model-based agent  &Partial decomp. &\citep{lu_robot_2018}\\

HL domain, FSA & RL Agent, RS$^\S$ &  Partial decomp. & \citep{camacho_ltl_2019} \\

HL domain, SP & Controller, RS$^\S$ & Partial decomp. & \citep{grzes_plan-based_2008} \\

\bottomrule
\end{tabular}
\begin{tablenotes}
    \footnotesize
    \item[]$^{\star}$ Low-level, High Level $^\dag$Finite State Automaton $^\ddag$Symbolic Planner $^\diamond$Knowledge Representation $\&$ Reasoning $^\S$Reward Shaping 
\end{tablenotes}
\end{threeparttable}
\end{table}

Working at a higher level of abstraction than the controller, it seems reasonable that the high-level module (e.g., meta-controller, planner) manipulates symbolic representations and handles the reasoning part, while the low-level controller could still benefit from the flexibility of neural approaches.
Yet, we could also imagine less dichotomic architectures: for instance, another neural module may coexist at the high-level along with logic-based module to better inform the decisions under uncertainty (as in \shortcite{sridharan_reba_2019}).


\footnotetext[\value{savefootnote}]{Stochastic Temporal grammar may be given as a prior, or trained.\stepcounter{savefootnote}}


\normalsize 

\cg{Various domain-specific symbolic knowledge may be incorporated to HRL frameworks, in order to leverage both learning and high-level reasoning: high-level domain-knowledge (as w/ high-level transitions, and mappings from low-level to high level), or high-level plans, task-decomposition or also specific decisions rules (e.g. for safety filtering). High-level domain knowledge may be described through  symbolic logic-based or \textit{action language} such as PDDL or RDDL \citep{sanner_relational_2011} or even \npw{via} temporal logic \citep{camacho_ltl_2019,li_formal_2019}.
}



\paragraph{Modularity}
Modular approaches in HRL have been relevantly applied to decompose a possibly complex task domain into different regions of specialization, but also to multitask DRL, in order to reuse previously learned skills across tasks \citep{andreas_modular_2017,shu_hierarchical_2018,wu_model_2019}.
Tasks may be annotated by handcrafted instructions (as ``policy sketches'' in \shortcite{andreas_modular_2017}), and symbolic subtasks may be associated with subpolicies which a full task-specific policy aims to successfully combine. Distinctively, \cite{shu_hierarchical_2018} proposes to train---or provide as a prior---a stochastic temporal grammar (STG), in order to capture temporal transitions between the tasks; a STG encodes priorities of sub-tasks over others and enables to better learn how to switch between base or augmented subpolicies.
\npw{The work in \cite{wu_model_2019} seeks} to learn a mixture of subpolicies by modeling the environment assuming given a set of (imperfect) models specialized by regions, referred to as \textit{model primitives}.
Each policy is specialized on those regions and the weights in the mixture correspond to the posterior probability of a model given the current state.
Echoing the hierarchical abstractions involved in our human decision-making, such modularity and task-decomposability---referred to as \textit{high-level modularity} as previously---would arguably  participate to make decision-\pw{making} more intelligible. 

\paragraph{Symbolic Planning}

A specific line of work within HRL has emerged trying to fuse symbolic planning \pw{(SP)} \citep{cimatti_automated_2008} with RL (SP+RL), to guide the agent’s task execution and learning \citep{leonetti_synthesis_2016}.
In classical SP, an agent uses a symbolic planner to generate a sequence of symbolic actions---\textit{plan}---based on its symbolic knowledge.
\cg{Yet, this naive pre-defined notion of planning seems unfit to most RL or real-world domains which present both domain uncertainties and execution failures. 
Recent work thereupon usually interleaves RL and SP, aiming to send feedback signals to the planner in order to handle such scenarios.
\cg{Planning agents often carry prior knowledge of the high-level dynamics, typically hand-designed, and assumingly consistent with the low-level environment.\footnote{In contrast, the work in RL+SP mentioned in \Cref{sec:extracting symbolic} does not assume similar HL domain knowledge, and aims to learn the mapping from the low-level domain to high-level symbols.}}

Framing SP in the context of automatic option discovery, in PEORL \citep{yang_peorl_2018},
a constraint answer set solver generates a symbolic plan, which is then turned into a sequence of options to guide the reward-based learning. Unlike earlier work in SP-RL} (e.g., \shortcite{leonetti_synthesis_2016}), RL is intertwined with SP, such that more suitable options could be selected.
Some work has extended PEORL, with two planning-RL loops for more robust and adaptive task-motion planning \citep{jiang_integrating_2018}, or with an additional meta-controller in charge of subtasks evaluation in order to propose new intrinsic goals to the planner \citep{lyu_sdrl_2019}.

Other work has focused on taskable RL in hierarchical RL, as in \cite{illanes_symbolic_2020}, which bypasses the problem of learning relevant goals for the planner. 

\paragraph{Declarative Domain Knowledge}

\cg{Declarative and common sense knowledge has been incorporated in RL frameworks in diverse ways to guide exploration, such as to filter out unreasonable or risky actions with finite state automaton \citep{li_formal_2019} or high-level rules \citep{zhang_faster_2019}.
\pw{\npw{By contrast}, \cite{furelos-blanco_induction_2021} proposes to learn a finite-state automaton for the higher-level with an ILP method and solve the lower-level with an RL method. There, the automaton is used to generate subgoals for the lower level.}
A deeper integration of knowledge representation and reasoning with model-based \pw{RL} has been advocated in \cite{lu_robot_2018}, where the learned dynamics are fed into the logical-probabilistic reasoning module to help it select a task for the controller.}
In a different direction, exploiting declarative knowledge to construct actions sequences can also help reward shaping to find the optimal policy, as a few studies \citep{camacho_ltl_2019,grzes_plan-based_2008} have demonstrated.
We refer to the survey \pw{by \cite{zhang_survey_2020}} for further examples of studies both in probabilistic planning and RL aiming to reason with declarative domain knowledge.

\paragraph{Conclusion}

Symbolic knowledge and reasoning elements have been consistently incorporated \pw{to} (symbolic) planning or hierarchical decision-making, in models which may reach a certain high-level interpretability, albeit neglecting action-level interpretability.
Moreover, most work still typically relies on manually-crafted symbolic knowledge---or has to rely \mz{on} a (pretrained or jointly-trained) perception model for symbol grounding---assuming a pre-given symbolic structure hand-engineered by \pw{a} human expert.
Due to the similarities shared by the majority of discrete dynamic domains, some researchers \citep{lyu_sdrl_2019} have argued that a laborious crafting of symbolic model is not always necessary, as the symbolic formulation could adapt to different problems, by instantiating new types of objects and a few additional rules for each new task; \cg{such claim would still need to be backed by further work in order to demonstrate such flexibility.} 
\cg{When adopting a symbolic or logical high-level framework, we also have to handle a new trade-off between expressivity and complexity of the symbolic representation.}

\section{Interpretable Transition and Preference Models} \label{sec:model}

\pw{
In this section, we overview the work that focuses on} exploiting \pw{an interpretable model of the environment or task.
This model can take the form of a transition model (\Cref{sec:transition model}) or preference model (\Cref{sec:preference model}).
Such interpretable models can help an RL agent reason about its decision-making, but also help humans understand and explain its decision-making.
As such, they can be used in an interpretable RL algorithm, but also in a post-hoc procedure to explain the agent's decision-making.
}
Note that those models may be learned, or not, and, when not learned, may possibly be fully provided to the RL agent or not.
For instance, the reward function, which is one typical way of defining the preference model, is generally specified by the system designer in order to guide the RL agent to learn and perform a specific task.
This function is usually not learned directly by the RL agent (except in inverse RL, \shortcite{ng_algorithms_2000}), but still has to be intelligible in some sense, otherwise the agent's decision-making may be based on spurious reasons and the agent may not accomplish the desired task \npw{since a non-intelligible reward function is hard to verify and may be incorrectly specified}.

\subsection{Transition Model} \label{sec:transition model}


Interpretable transition models can help discover the structure and potential decomposition in a problem that are useful for more data-efficient RL (via e.g., more effective exploration), but also allow for larger generalizability and better transfer learning.
Various interpretable representations have been considered for learning transition models, such as decision trees or graphical models for probabilistic models, physics-based or graph for deterministic models, or neural networks with architectural inductive bias.
\pw{The neural network-based approaches, which are more recent, are presented separately to emphasize them. 
We provide an overview of all the methods for interpretable transition models in \Cref{tab:transition_models}.}

\begin{table}[t]
\footnotesize
    \centering
    \caption{Overview of approaches for interpretable transition models}
    \label{tab:transition_models}
    
    \begin{tabular}{@{}llll@{}}
    \toprule
        Type & Model & Interpretability & Refs \\
    \midrule
        Probabilistic & Decision tree & Simulatable & \citep{degris_learning_2006} \\
        Probabilistic & Noisy deictic rule & Decomposable & \citep{pasula_learning_2007} \\
        Probabilistic & First-order rule & Decomposable & \citep{walker_building_2008} \\
        Probabilistic & Relational action schema & Decomposable & \citep{walsh_efficient_2010} \\
        Probabilistic & Relational planning operator 
        & Decomposable & \citep{martinez_learning_2016,martinez_relational_2017} \\
        Probabilistic & Weighted labeled multigraph & Decomposable & \citep{metzen_learning_2013} \\
        Probabilistic & Graphical model & Decomposable & \citep{kansky_schema_2017} \\
        Probabilistic & Gaussian process & Decomposable & \citep{kaiser_interpretable_2019} \\
        
        Deterministic & Object-oriented representation & Decomposable & \citep{diuk_object-oriented_2008} \\
        Deterministic & Deictic object-oriented & Decomposable & \citep{marom_zero-shot_2018} \\
        & representation & & \\
        Deterministic & Physics engine
        & Decomposable & \citep{scholz_physics-based_2014} \\
        Deterministic & Graph of transitions between & Decomposable & \citep{zhang_composable_2018} \\
        & user-defined Boolean attributes & & \\
        Deterministic & State-space graph 
        & Decomposable & \citep{eysenbach_search_2019} \\
    \midrule
        Neural & Graph neural networks & Partial decomp. & \citep{battaglia_interaction_2016} \\
        & & & \citep{sanchez-gonzalez_graph_2018} \\
        Neural & Object dynamic predictor  & Partial decomp. & \citep{finn_unsupervised_2016} \\
        & from pixels & & \citep{finn_deep_2017} \\
        Neural & Object dynamic predictor  & Partial decomp. & \citep{zhu_object-oriented_2018,zhu_object-oriented_2020} \\
        & with relations & & \\
        Neural & Object-level dynamics model & Partial decomp. & \citep{agnew_unsupervised_2018} \\
        & with unsupervised learning & & \\
        Neural & Models for entity grounding, & Partial decomp. & \citep{veerapaneni_entity_2020} \\
        & dynamics, and & & \\
        & observation distrib. & & \\
    \bottomrule
    \end{tabular}
\end{table}

\paragraph{Probabilistic Models} While the setting of factored MDPs generally assumes that the structure is given, \cite{degris_learning_2006} proposes a general model-based RL approach that can both learn the structure of the environment and its dynamics.
This method is instantiated with decision trees to represent the environment model.
Statistical $\chi^2$ tests are used to decide for the structural decomposition.


Various approaches are based on generative models under the form of graphical models.
\npw{In \cite{metzen_learning_2013}, a graph-based representation of the transition model is learned} in continuous domains.
\npw{In \cite{kansky_schema_2017}, schema networks are proposed as generative models} to represent the transition and reward models in a problem described in terms of entities (e.g., objects or pixels) and their attributes.
As a graphical model, the authors explain how to learn its structure and how to use it for planning using inference.
\npw{The work in \cite{kaiser_interpretable_2019} learns} via variational inference an interpretable transition model by encoding high-level knowledge in the structure of a graphical model.

In the relational setting, a certain number of studies explored the idea of learning a relational probabilistic model for representing the effects of actions (see \shortcite{walsh_efficient_2010} for discussions of older work).
\npw{In summary, those} approaches are either based on batch learning (e.g., \shortcite{pasula_learning_2007,walker_building_2008}) or online methods (e.g., \shortcite{walsh_efficient_2010}) with or without guarantees using more or less expressive relational languages.
Some recent work \cite{martinez_learning_2016,martinez_relational_2017} proposes to learn a relational probabilistic model via inductive logic programming (ILP) and uses optimization to select the best planning operators.

\paragraph{Deterministic Models} 
\npw{In \cite{diuk_object-oriented_2008}, an efficient model-based approach is proposed for an object-oriented representation of the world.}
\pw{This approach is extended \cite{marom_zero-shot_2018} to} 
deictic object-oriented representations, which use partially grounded predicates, in the KWIK framework \citep{walsh_efficient_2010}.

Alternatively, \cite{scholz_physics-based_2014} explores the use of a physics engine as a parametric model for representing the \pw{deterministic} dynamics of the environment.
The parameters of this engine are learned by a Bayesian learning approach.
Finally, the control problem is solved using the A* algorithm.

In a hierarchical setting, several recent studies have proposed to rely on search algorithms on state-space graphs or planning algorithms for the higher-level policy.
Given a mapping from the state space to a set of binary high-level attributes, \cite{zhang_composable_2018} proposes to learn a model of the environment predicting if a low-level policy would successfully transition from an initial set of binary attributes to another set.
The low-level policy observes the current state and the desired set of attributes to reach.
Once the transition model and the policy \pw{are} learned, a planning module can be applied to reach \pw{the} specified high-level goals.
Thus, the high-level plan is interpretable, but the low-level policy is not.

\npw{In \cite{eysenbach_search_2019}, the authors} extract a state-space graph from a replay buffer and apply the Dijkstra algorithm to find a shortest path to reach a goal.
This graph represents the state space for the high level, while the low level is dealt with a goal-conditioned policy.
The approach is validated in navigation problems with high-dimensional inputs.

\paragraph{Neural Network-based Model} 
Various recent propositions have tried to learn dynamics model using neural networks with specific architectural inductive bias taking \pw{graphs as inputs}
\citep{battaglia_interaction_2016,sanchez-gonzalez_graph_2018}.
While this line of work can provide high-fidelity simulators, the learned model \pw{may} suffer from \cg{a lack of} interpretability.

The final set of work we would like to mention aims at learning object-based dynamics models from low-level inputs (e.g., frames) using neural networks.
In that sense, they can be understood as an extension of the work in relational domain where the input is now generally high-level. 
One early work \citep{finn_unsupervised_2016,finn_deep_2017} tries to take into account moving objects.
However, the model predicts pixels and does not take into account relations between objects, which limits its generalizability.
\npw{In \cite{zhu_object-oriented_2018}, a novel neural network is proposed}, which can be trained in an unsupervised way, for object detection and object dynamic prediction conditioned on actions and object relations.
While this work focuses on one dynamic object, it has been generalized to multiple ones \citep{zhu_object-oriented_2020}.


\npw{Another work \cite{agnew_unsupervised_2018} proposes} an unsupervised method called Object-Level Reinforcement Learner (OLRL), which detects objects from pixels and learns a compact object-level dynamics model.
The method works according to the following steps.
Frames are first segmented into blobs of pixels, which are then tracked over time.
An object is defined as blobs having similar dynamics.
Dynamics of those objects are then predicted with a gradient boosting decision tree.


\npw{In \cite{veerapaneni_entity_2020} an end-to-end object-centric perception, prediction, and planning (OP3) framework is developed. 
The} model has different components jointly-trained: for (dynamic) entity grounding, for modeling the dynamics and for modeling the observation distribution. 
The variable binding problem is treated as an inference problem: being able to infer the posterior distribution of the entity variables given a sequence of observations and actions. 
One further specificity of \npw{this} work is to model a scene not globally but locally, i.e., for each entity and its local interactions (locally-scoped entity-centric functions), avoiding the complexity to work with the full combinatorial space, and enabling generalization to various configurations and number of objects. 

\paragraph{Conclusion}
Diverse approaches have been considered for learning an interpretable transition model.
They are designed to represent either deterministic or stochastic environments.
Recent work is based on neural networks in order to process high-dimensional inputs (e.g., images) and has adopted an object-centric approach.
While neural networks hinder the intelligibility of the method, the decomposition into object dynamics can help scale and add transparency to the transition model.

\subsection{Preference Model} \label{sec:preference model}

\pw{In RL, the usual approach to describe the task to be learned or performed by an agent consists in defining suitable rewards, which is often a hard problem to solve for the system designer.}
The difficulty of reward specification has been recognized early \citep{russell_learning_1998}.
Indeed, careless reward engineering can lead to undesired learned behavior \citep{randlov_learning_1998} and to value misalignment \citep{arnold_value_2017}.
Different approaches have been proposed to circumvent or tackle this difficulty: 
imitation learning (or behavior cloning) \citep{pomerleau_alvinn_1989,osa_algorithmic_2018}, inverse reinforcement learning \citep{ng_algorithms_2000,arora_survey_2018}, 
learning from human advice \citep{maclin_creating_1996,kunapuli_guiding_2013}, or 
preference elicitation  \citep{rothkopf_preference_2011,weng_interactive_2013}.
\Cref{tab:reward_models} summarizes the related work for interpretable preference models.
As a side note, some of the previous approaches for learning a transition model apply here as well (e.g., \shortcite{walker_building_2008}).

\begin{table}[t]
\footnotesize
    \centering
    \caption{Overview of approaches for interpretable preference models}
    \label{tab:reward_models}
    
    \begin{tabular}{@{}llll@{}}
    \toprule
        Model & Approach & Interpretability & Refs \\
    \midrule
        Relational & Inverse RL & Simulatable & \citep{munzer_inverse_2015} \\
        Relational & Active learning & Decomposable & \citep{martinez_relational_2017-1} \\
        Deep decision trees & Batch adversarial  & Partial decomp. & \citep{srinivasan_interpretable_2020} \\
         &  inverse RL &  &  \\
    
        Linear Temporal Logic & Pre-specified & Simulatable & \citep{aksaray_q-learning_2016} \\
        Geometric LTL & Pre-specified & Simulatable & \citep{littman_environment-independent_2017} \\
        Truncated LTL & Pre-specified & Simulatable & \citep{li_reinforcement_2017,li_formal_2019} \\
        LTL & Pre-specified & Simulatable & \citep{toro_icarte_teaching_2018} \\
        
        Finite-state machine & Pre-specified & Simulatable & \citep{toro_icarte_using_2018} \\
        Formal languages & Pre-specified & Simulatable & \citep{camacho_ltl_2019} \\
        LTL & Pre-specified & Simulatable & \citep{hasanbeig_deep_2020} \\
        Finite-state machine & Local search & Simulatable & \citep{toro_icarte_learning_2019} \\
        Finite-state machine & Automata learning & Simulatable & \citep{xu_joint_2020} \\
        Finite-state machine & Automata learning & Simulatable & \citep{gaon_reinforcement_2020} \\
        Boolean task algebra & Pre-specified & Simulatable & \citep{tasse_boolean_2020} \\
    \bottomrule
    \end{tabular}
\end{table}

Closer to interpretable RL, \cite{munzer_inverse_2015} extends inverse RL to relational RL\npw{, while
\cite{martinez_relational_2017-1} learns} from demonstration\mz{s} in relational domains by active learning.
\npw{In \cite{srinivasan_interpretable_2020}, more interpretable rewards are learned} using deep decision trees \citep{yang_deep_2018}.

Recent work has started to investigate the use of temporal logic (or variants) to specify an RL task \citep{aksaray_q-learning_2016,littman_environment-independent_2017,li_reinforcement_2017,li_formal_2019}.
Related to this direction, \cite{kasenberg_interpretable_2017} investigates the problem of learning from demonstration 
an interpretable description of an RL task under the form of linear temporal logic (LTL) specifications.

\npw{Another related work \cite{toro_icarte_using_2018} proposes} to specify and represent a reward function as a finite-state machine, called reward machine, which clarifies the reward function structure.
Reward machines can be specified using inputs in a formal language, such as linear temporal logic \citep{camacho_ltl_2019,hasanbeig_deep_2020}.
Reward machines can also be learned by local search \citep{toro_icarte_learning_2019} or with various automata learning techniques  \citep{xu_joint_2020,gaon_reinforcement_2020}.

\pw{
Using a different approach, \cite{tasse_boolean_2020} shows how a Boolean task algebra, if such structure holds for a problem, can be exploited to generate solutions for new tasks by task composition \citep{todorov_compositionality_2009}.
Such approach can arguably provide interpretability of the solutions thus obtained.
}

\paragraph{Conclusion}
As can be seen, research in interpretable preference representation is less developed than for transition models.
However, we believe interpretability of preference models is as important if not more than interpretability for describing the environment dynamics, when trying to understand the action selection of an RL agent.
Thus, more work is needed in this direction to obtain more transparent systems based on RL.

\section{Interpretable Decision-Making} \label{sec:decision-making}

\begin{figure*}[t]
\centering
\includegraphics[trim=0 125pt 0 0, clip, scale=.08]{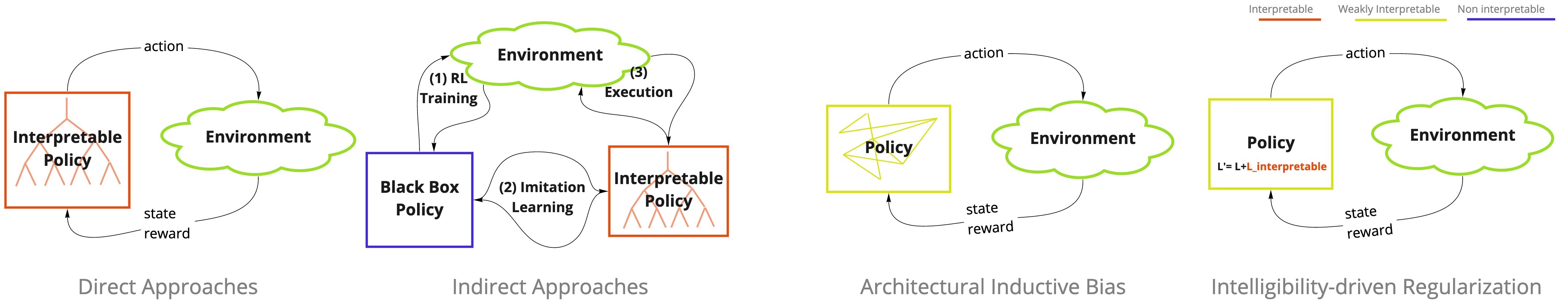}
\caption{Illustrations of the different approaches for interpretable decision-making. From left to right: Direct Approach, Indirect Approach, Architectural Inductive Bias, Intelligibility-driven Regularization.}
\label{fig:interpretable decision making}
\end{figure*}

\pw{We now turn to the main part of this survey paper, which deals with the question of interpretable decision-making.
Among the various approaches that have been explored to obtain interpretable policies, we distinguish four main families (see \Cref{fig:interpretable decision making}).
Interpretable policies can be learned directly (\Cref{sec:direct}) or indirectly (\Cref{sec:indirect}), \npw{and in addition,} in deep RL, interpretability can also be enforced or favored at the architectural level (\Cref{sec:architectural inductive bias}) or via regularization (\Cref{sec:preference bias}).}

\subsection{Direct Approach} \label{sec:direct}

\pw{Work in the direct approach} aims at directly searching for a policy in \pw{a policy space \npw{chosen and} accepted as interpretable \npw{by the system designer}}.
They can be categorized according to their search space and the method to search in this space (Table \ref{tab:direct_approach}).
Several models have been considered in the literature:

\begin{table}[t]
\footnotesize
    \centering
    \caption{Overview of direct approaches.}
    \label{tab:direct_approach}
    \begin{tabular}{@{}llll@{}}
    \toprule
        Type & Approach & Interpretability & Refs \\
    \midrule
        Decision tree & Tree-based algorithms & Simulatable & 
        \citep{ernst_tree-based_2005} \\
        Decision tree & Gradient descent & Simulatable & \citep{likmeta_combining_2020,silva_optimization_2020,silva_neural-encoding_2020,topin_iterative_2021,gupta_policy_2015} \\
        Formulas & Multi-armed bandit & Decomposable & \citep{maes_policy_2012,maes_automatic_2012} \\
         & with depth search & &  \\
        Formulas & Genetic algorithms & Decomposable & \citep{hein_interpretable_2018,hein_generating_2019} \\
        Formulas & Gradient descent & Decomposable & \citep{ault_learning_2020} \\
        Fuzzy controllers & Gradient descent & Decomposable & \citep{akrour_towards_2019} \\
        Fuzzy controllers & Particle swarm & Decomposable & 
        \citep{hein_particle_2017} \\
        Logic Rules & Gradient Descent & Decomposable &  \citep{evans_learning_2018,jiang_neural_2019,payani_inductive_2019,payani_learning_2019,payani_incorporating_2020,zimmer_differentiable_2021} \\
        Chained Logic Rules & Gradient descent & Decomposable & \citep{yang_learn_2019,ma_interpretable_2020} \\
        Programs & Gradient descent & Decomposable & \citep{verma_imitation-projected_2019,anderson_neurosymbolic_2020} \\
        Graphical Models & Gradient descent  & Decomposable & \citep{chen_interpretable_2020,levine_reinforcement_2018} \\
        
    \bottomrule
    \end{tabular}
\end{table}

\paragraph{Decision Trees}
A decision tree is a directed acyclic graph where the nodes can be categorized into decision nodes and leaf nodes.
It is interpretable by nature, but learning it can be computationally expensive.
The decision nodes will determine the path to follow in the tree until a leaf node is reached, this selection is mostly done according to the state features.
\pw{Decision trees can be used to represent value functions or policies.}
\npw{For instance, some older work} 
\citep{ernst_tree-based_2005} uses a decision tree
to represent the Q-value function where each leaf node represents the Q-value of an action \pw{in a state}.
Their optimization method is based on \pw{decision} tree-based supervised learning methods which do not rely on differentiability.

\pw{In contrast,} \cite{likmeta_combining_2020} proposes to learn parameterized decision nodes.
In this approach, the policy instead of the value function is represented by the decision tree \npw{where a leaf represents the action to take.}
\npw{The structure of the tree is assumed to be} given by experts.
To update the tree parameters, policy gradient with parameter-based exploration is employed.
Similarly, \cite{silva_optimization_2020} designs a method to discretize differentiable decision trees such that policy gradient can be used during learning. 
Therefore, the whole structure of the tree can be learned.
\npw{The analysis in \cite{silva_optimization_2020}} also suggests that representing the policy instead of the value function with a decision tree was more beneficial.
Expert or other prior knowledge may also bootstrap the learning process, as demonstrated by \cite{silva_neural-encoding_2020}, where the policy tree is initialised from human-provided knowledge, before being dynamically learned.
\npw{In addition}, \cite{topin_iterative_2021} introduces a method defining a meta-MDP from a base MDP with additional actions \npw{where}  any policy in the meta-MDP can be transformed in a decision tree policy in the base MDP.
In this way, the meta-MDP can be solved by classic deep reinforcement learning algorithms.

\pw{In a different approach}, \cite{gupta_policy_2015} proposes to learn a binary decision tree where each leaf is itself a parametric policy. 
Linear Gibbs softmax policies are learned in discrete action spaces, while in continuous action spaces Gaussian distributions are learned.
These parametric policies remain interpretable since their parameters are directly interpretable (probabilities for Gibbs softmax policies, mean and standard deviation for Gaussian distribution) and do not interact with the state.  
Hence, the composition of the decision tree with the parametric policies is interpretable.
Policy gradient is employed to update the parametric policies and to choose how the tree should grow.

\paragraph{Formulas}

\npw{The work of \cite{maes_policy_2012} proposes} to represent the Q-value function with a simple closed-form formula.
The policy is then defined greedily with respect to this value function.
The considered binary operations are addition, subtraction, multiplication, division, minimum, and maximum.
The possible unary operations are the square root, logarithm, absolute value, negation, and inverse.
The possible variables in the formulas contain all the components of the states and the actions.
Because of the combinatorial explosion, the total number of operators, constants, and variables occurring in a formula was limited to 6 in their experiments.
To search among this space, the authors formulate a multi-armed bandit problem and used a depth-limited search approach \citep{maes_automatic_2012}.

\pw{\npw{In \cite{hein_interpretable_2018,hein_generating_2019},} a formula is used to directly represent a policy.}
\npw{In this work, expressivity is improved} by adding more operators (tanh, if, and, or) and deeper formulas (a maximum depth of 5 and around 30 possible variables).
\npw{Genetic programming is used to search for the formula when a batch of RL transitions is available.}.

\pw{A different approach is proposed in the context of traffic light control by \cite{ault_learning_2020} who design} a dedicated interpretable polynomial function where the parameters are learned by a variant of DQN.
\pw{This function is then used similarly to a Q-value function to derive a policy.}

\paragraph{Fuzzy controllers}
Fuzzy controllers define the policy as a set of fuzzy ``if-then'' rules of the form:
$$ \text{IF } fuzzy\_condition(state) \text{ DO } action. $$
\npw{In \cite{akrour_towards_2019}, it is assumed that} the state can be categorized in a discrete number of clusters with a fuzzy membership function.
Hence, $fuzzy\_condition(state)$ is defined as a distance to a centroid. 
The policy is defined as a Gaussian distribution such that the closer a state is to a centroid, the more the mean associated to the centroid is taken into account for the global mean.
They learn the mean associated to each cluster with policy gradient given a non-interpretable critic.
Similarly, \cite{hein_particle_2017} 
learns fuzzy rules for deterministic policies with particle swarm optimization in continuous action domains.
In both approaches, the number of rules (and clusters) are adapted automatically.


\paragraph{Logic Rules}
Neural Logic Reinforcement Learning (NLRL) \cite{jiang_neural_2019} aims at representing policies by first-order logic. 
NLRL combines policy gradient methods with a new differentiable inductive logic programming architecture adapted from \cite{evans_learning_2018}. 
All the possible rules are generated given rule templates provided by experts.
To represent the importance of the rule in the deduction, a weight is associated to each rule.
As all the rules are applied with a softmax over their weights, the resulting predicate takes its value over the continuous interval $[0; 1]$ during learning.
Such an approach is able to generalize to domains with more objects than it was trained on, but it is costly to compute all the applications of all the possible rules during training.

In the previous approach, to limit the number of possible rules, the templates are generally formulated such that the number of atoms in the body of a rule is restricted to two.
To overcome this limitation,
\cite{payani_inductive_2019} and \cite{payani_learning_2019} design an alternative model\pw{, enforcing formulas to be in disjunctive normal form,} where weights are associated to atoms (instead of rules) in a clause and extend it to RL \citep{payani_incorporating_2020}.
Similarly, \cite{zimmer_differentiable_2021} also defines weights associated to atoms.
\pw{However, the architecture proposed by \cite{dong_neural_2019} is adapted to enforce interpretability and} rely on a Gumbel-Softmax distribution to select the arguments in a predicate.
\pw{This approach can be more interpretable than previous similar work, since it can learn a logic program instead of a weighted combination of logic formulas.}

Alternatively, \cite{yang_learn_2019} proposes a differentiable ILP method extending the multi-hop reasoning framework \citep{lao_relational_2010,yang_differentiable_2017}. 
Instead of performing forward-chaining on predefined templates, weights are associated to every possible relational paths where each path corresponds to a multi-step chain-like logic formula.
Compared to the previous work, it is \pw{less expressive since it is} not able to represent full Horn clauses, but has a better scalability.
\npw{This approach is further extended \cite{ma_interpretable_2020}} to the reinforcement learning setting.

\paragraph{Programs}
\npw{In \cite{verma_imitation-projected_2019}, a novel approach is proposed} to learn directly a policy written as a program.
Their approach can be seen as inspired by (constrained functional) mirror descent.
Indeed, their algorithm iteratively updates the current policy using a gradient step in the continuous policy space that mixes neural and programmatic representations, then projects the resulting policy in the space of programmatic policies via imitation learning.
\npw{This approach is extended \cite{anderson_neurosymbolic_2020} to} safe reinforcement learning in order to avoid unsafe states during exploration with formal verification.

\pw{
\paragraph{Graphical Models}
Most previously-discussed work uses a deterministic interpretable representation.
However, one may also argue that probabilistic graphical models are also interpretable.
Thus, for instance, in the context of autonomous driving, \cite{chen_interpretable_2020} solves the corresponding DRL problem as a probabilistic inference problem \citep{levine_reinforcement_2018}: for the RL model and policy, they learn probabilistic graphical models with hidden states, which are trained to be interpretable by enforcing semantic meanings available at training time.
The drawback of this approach is that it can only provide interpretability to the learned latent space.
}




\paragraph{Conclusion}
Using the direct approach 
to find interpretable policies in RL problems is complex since we must be able to solve two \pw{potentially conflicting problems at the same time}: 
(1) finding a good policy for the given (PO)MDP and 
(2) keeping that policy interpretable.
These two objectives become more and more contradictory when the RL problems are \npw{larger}, resulting in a scalability issue with the direct approach.
Most work in this section, with a fully interpretable policy, focuses only on small toy problems.

The direct approach is related to discrete optimization where the objective function is not differentiable and looking for a policy in such a space is very difficult.
Another limitation of these approaches is their poor robustness to noise.
To overcome those issues, several approaches use a continuous relaxation to make the objective function differentiable (\pw{i.e.,} search in \pw{a} smoother space) and more robust to noise, but the scalability issue remains open.

\subsection{Indirect Approach} \label{sec:indirect}


In contrast to the direct approach, the indirect approach follows two steps:
first train a non-interpretable policy with any efficient RL algorithm, then transfer this trained policy to an interpretable one.
Thus, this approach is related to imitation learning \citep{hussein_imitation_2017} and policy distillation \citep{rusu_policy_2016}.
\npw{Note a similar two-step approach can be found in post-hoc explainability for RL. 
However, a key difference concerns how the obtained interpretable policy is used, either as a final controller or as a policy that explains a black-box controller, which leads to different considerations about how to learn and evaluate such interpretable policy (see Section~\ref{sec:XRL}).}
Similarly to RL algorithms that can be subdivided into value-oriented methods and policy-oriented methods, the focus in the indirect approach may be to obtain an interpretable representation of either a learned Q-value function (which provides an implicit representation of a policy) or a learned policy (often called \textit{oracle}), although most work focuses on the latter case.
Regarding the types of interpretable policies, decision trees or their variants are often chosen due to their interpretability, however other representations like programs have also been considered.

\paragraph{Decision Trees and Variants}
The work by \cite{liu_toward_2018} is a representative recent work among the value-oriented methods using decision trees.
The authors introduce Linear \npw{Model} U-trees (LMUTs) to approximate Q-functions estimated by neural networks in DRL.
LMUTs is based on U-tree \citep{maes_learning_1996}, which is a tree-structured representation specifically designed to approximate a value function.
A U-Tree, whose structure and parameters are learned online, can be viewed as a compact decision tree where each arc corresponds to the selection of the feature of a current or past observation, and each path from the root to a leaf represents a cluster of observation histories having the same Q-values.
LMUTs extend U-Trees by having in each leaf a linear model, which is trained by stochastic gradient descent.
Although LMUT is undoubtedly a more interpretable model than a neural network, it shows its limit when dealing with high-dimensional features spaces (e.g., \pw{images}).
\npw{In \cite{liu_toward_2018}, rules extraction and super-pixels \citep{ribeiro_why_2016} are used} to explain the decision-making of the resulting LMUT-based agent.

Many policy-oriented methods propose to learn a decision tree policy.
The difficulty of this approach is that a high-fidelity policy may require a large-sized decision tree.
To overcome this difficulty, 
\cite{bastani_verifiable_2018} presents a method called VIPER that builds on DAGGER \citep{ross_reduction_2011}, a state-of-the-art imitation learning algorithm, but exploits the available learned Q-function. 
The authors show that their proposition can achieve comparable performance to the original non-interpretable policy, and is amenable to verification.
As an alternative approach to control the decision tree size, \cite{roth_conservative_2019} proposes to increase its size only if the novel decision tree increases sufficiently the performance. 
As an improvement to work (like VIPER) using only one decision tree, 
\npw{In \cite{vasic_moet_2019}, a mixture of Expert Trees (MOET) is proposed}. 
The approach is based on a gating function that partitions the state space and then within each partition, a decision tree expert (via VIPER) approximates the policy.  

For completeness, we mention a few other relevant studies, mostly based on imitation learning:
\cite{natarajan_imitation_2011} learns a set of relational regression trees in relational domains by functional gradient boosting;
\cite{cichosz_imitation_2014} learns decision tree (and random forest) policies for car driving; 
\cite{nageshrao_interpretable_2019} extracts a set of fuzzy rules from a neural oracle.

\paragraph{Programs}
As an alternative to decision trees, 
\cite{verma_programmatically_2018} introduce a framework,  Programmatically Interpretable Reinforcement Learning (PIRL), that generates policies represented in a high-level, domain-specific programming language. 
In order to find a program that can reproduce the performance of a neural oracle, they propose a new method, Neurally Directed Program Search (NDPS). 
NPDS performs a local search over the non-smooth space of programmatic policies in order to minimize a distance from this neural oracle computed over a set of adaptively chosen inputs. 
To restrict the search space, a policy sketch is assumed to be given.
Unlike the imitation learning setting where the goal is to match the expert demonstrations perfectly, a key feature of NPDS is that the expert trajectories only guide the local program search in the program space to find a good policy. 

\npw{In \cite{zhu_inductive_2019}, a search technique is also proposed} to find a program to mimic a trained neural network policy for verification and shielding \citep{alshiekh_safe_2018}.
The novelty in their approach is to exploit the information of safe states, assumed to be given.
If a generated program is found to be unsafe from an initial state, this information is used to guide the generation of subsequent programs.

\npw{In \cite{burke_explanation_2019}, a method is proposed} to learn a program from demonstration for robotics tasks that are solvable by applying a sequence of low-level proportional controllers.
In a first step, the method fits a sequence of such controllers to a demonstration using a generative switching controller task model.
This sequence is then clustered to generate a symbolic trace, which is then used to generate a programmatic representation by a program induction method.

\pw{
Finally, although \npw{strictly speaking} not a program,  \cite{Koul_learning_2019} proposes to extract from a trained recurrent neural network policy a finite-state representation (i.e., Moore machine) that can approximate the trained policy and possibly match its performance by fine-tuning if needed.
This representation is arguably more interpretable than the original neural network.
}

\paragraph{Conclusion}
\pw{As mentioned previously,} the direct approach requires tackling \cg{simultaneously} two difficulties: (1) solve the RL problem and (2) obtain an interpretable policy.
In contrast to the direct approach, the indirect approach circumvents the first above-mentioned difficulty \pw{at the cost of solving two consecutive (hopefully easier) problems: (1) solve the RL problem with any efficient RL algorithm, (2) mimic the good learned policy with an interpretable one by solving a supervised learning problem}.
Therefore, any imitation learning \citep{hussein_imitation_2017} and policy distillation \citep{rusu_policy_2016} methods could be applied \pw{to obtain an interpretable policy} in the indirect approach.
However, the indirect approach is a more flexible setting than \pw{the standard} imitation learning \pw{setting} because of the unrestricted access to (1) an already-trained expert policy using the same observation/action spaces, and (2) its value function as well. 
The teacher-student framework \citep{torrey_teaching_2013} fits particularly well this setting.
As such, it would be worthwhile to investigate the applications of techniques proposed for this framework (e.g., \shortcite{zimmer_teacher-student_2014}) to the indirect approach.

\pw{In addition, t}he work by \cite{verma_imitation-projected_2019} seems a promising approach to combine the direct and indirect approaches.
While the authors show that the performance of their proposition outperforms NDPS, it is currently still not completely clear which of a direct method or an indirect one should be preferred to learn good interpretable policies.

\subsection{Architectural Inductive Bias} \label{sec:architectural inductive bias}


To favor interpretable decision-making, specific architectural choices may be adopted for the policy network \pw{or value function}, may it be through relational, logical, or attention-based bias; some examples are presented \pw{in \Cref{tab:architectural_bias}}.

\begin{table}[t]
\footnotesize
    \centering
    \caption{Overview of approaches for Architectural Inductive Bias}
    \label{tab:architectural_bias}
    
    \begin{tabular}{@{}llll@{}}
    \toprule
        Inductive bias & Model & Intelligibity & Refs \\
    \midrule
        Relational & Graph neural network & Partial decomp & \citep{wang_nervenet_2018} \\

        Logical & Modular architecture: MLPs  & Weak partial decomp.  & \citep{dong_neural_2019} \\
        & wired w/ tensor operators & & \\
        Attention &  self-attention bottleneck,  & Partial decomp.  & \citep{tang_neuroevolution_2020} \\
        &  LSTM controller & Partial explainab. & \\
        
        Attention & ConvLSTM, attention module, & Partial decomp. &  \citep{mott_towards_2019} \\
        &LSTM controller & Partial explainab. & \\
        
        Attention & DQN architecture  & Partial decomp. &   \citep{annasamy_towards_2019} \\
        & w/ attention & Partial explainab. & \\

    \bottomrule
    \end{tabular}
\end{table}

\paragraph{Relational Inductive Bias}
\pw{Such bias} refers to inductive bias imposing constraints on relationships and interactions among entities in a learning process. 
Its nuances range from convolutional neural networks to Graph Neural Networks (GNN) as mentioned in \pw{S}ection \ref{sec:extracting symbolic} for representation learning, here designed for the policy network.
A representative example is NerveNet \citep{wang_nervenet_2018} which aim\pw{s} \pw{at learning} a structured \pw{policy}---parametrized as a GNN, and executing some graph propagation steps.


\paragraph{Logical Inductive Bias}
For instance, Neural Logic Machine (\pw{NLM}) \citep{dong_neural_2019} is an end-to-end differentiable neural-symbolic architecture for inductive learning and logic reasoning.
Predicates are represented by probabilistic tensors, i.e.\pw{,} grounded on any possible combination of objects.
\pw{From a set of premises (base predicates), the forward pass in NLM, mimicking a sequence of forward chaining steps, outputs some conclusive tensors.}
Some logical architectural bias is embedded, as through the explicit wiring among the neural modules to realize the logical existential quantifiers as tensorial operations. 
\pw{Such approach can be seen as learning on a continuous relaxation of logic programs.}
Some undeniable advantages of NLM compared with the neuro-symbolic literature is the improved inference time, and that it does not rely on hand-engineered rule templates. However, what it gains in scalability, it \pw{loses} in interpretability.


\paragraph{Attention-based Inductive Bias}
Another 
intelligibility incentive is the use of selective attention mechanisms for the policy network \citep{tang_neuroevolution_2020,mott_towards_2019}, or the Q-network \citep{annasamy_towards_2019}.
\npw{The work of \cite{tang_neuroevolution_2020} evolves} RL agents which are encouraged to attend to a small fraction of its visual input, by selecting which spatial patches of the input representation they feed to the LSTM controller.
\cg{\npw{Similarly, \cite{mott_towards_2019} presents} a soft, top-down, spatial attention mechanism applied to the visual input, while allegedly uncovering part of the underlying decision process, in terms of space (``where'') and content (``what''). 
Although the authors argue \cg{that} these attentions mechanisms yield more informative and reliable explanations than other methods for analyzing saliency, the correlation between attention and explainability has been both supported \citep{wiegreffe_attention_2019} and disputed \citep{jain_attention_2019,brunner_identifiability_2020} in further work along different scenarios.}
\pw{For related work focusing on explainability, see \Cref{sec:XRL}.}


\paragraph{Conclusion}
As recent work \pw{suggests}, relational or logical inductive bias can foster reasoning and generalization over structured data, may it be \mz{a} graph or predicates, and can improve learning efficiency and robustness,
while still benefiting from the flexibility of statistical learning, \pw{in contrast to pure} symbolic approaches.
Although, as soon as the environment and dynamics are complex, these learned relational, logical or attentive representations would not be sufficient to non-ambiguously make sense of the decision-making process.

It is worth mentioning some collateral deep learning work, which has used logical background knowledge as a way to shape the neural architecture itself, such as 
\cite{franca_fast_2014}, whose neural ILP-solver builds recursive neural networks, made with AND-OR type of networks.
However, strong architectural bias may drastically decreases the model's expressivity.



\subsection{\pw{Intelligibility-Driven Regularization}} \label{sec:preference bias}

An alternative to structural bias is to encompass a \pw{soft bias} on the hypothesis space, through some additional cost function favoring a certain notion of interpretability.
This additional \pw{term}, as ultimately aiming \pw{at improving} generalization error over training error, can be interpreted as a regularization technique.
\npw{Although this approach is natural and has received some attention in the broader scope of machine learning, it is relatively less explored in deep RL.
For this reason, we also discuss some non-RL studies in that direction, which could potentially be fruitfully adapted to RL.
\cg{We gather the RL work presented below in \Cref{tab:preference_bias}}
and the non-RL work in \Cref{tab:preference_bias_nonRL}.

\begin{table}[t]
\footnotesize
    \centering
    \caption{Overview of RL approaches for Soft Interpretability Bias}
    \label{tab:preference_bias}
    
    \begin{tabular}{@{}llll@{}}
    \toprule
        Regularizer & Model & Interpretability & Refs \\
    \midrule
        Smoothness reg. & Neural network & Weak & \citep{jia_advanced_2019} \\
        
        Alignment reg. & Model-based model-free,
 & Weak &  \citep{francois-lavet_combined_2019} \\
 
         & w/ double DQN &  & \\

    \bottomrule
    \end{tabular}
\end{table}

\begin{table}[t]
\footnotesize
    \centering
    \caption{Overview of non-RL approaches for Soft Interpretability Bias}
    \label{tab:preference_bias_nonRL}
    
    \begin{tabular}{@{}llll@{}}
    \toprule
        Regularizer & Model & Interpretability & Refs \\
    \midrule
        Model compression & Neural network  & Weak &  \citep{bucilua_model_2006}\\
        L$1-$Reg. & Neural network & Weak & \citep{zhang_1-regularized_2016} \\
        
         Legibility/Predictability Reg. & - & Weak &  \citep{dragan_legibility_2013}\\
         Tree-Reg.  & Neural network & Weak &   \citep{wu_optimizing_2019} \\       

    \bottomrule
    \end{tabular}
\end{table}
}

\cg{Classical regularization methods in deep learning which foster lower complexity should be beneficial for interpretability, although far from being sufficient; e.g. L1-regularization \cite{zhang_1-regularized_2016} encouraging sparsity 
or \textit{model compression} \cite{bucilua_model_2006}.}

Other interpretability-oriented penalty formulations have been proposed such as erratic-behavior penalties to improve smoothness \citep{jia_advanced_2019}, or objectives targeting legible \npw{or} predictable motions \citep{dragan_legibility_2013};
another example is given by \cite{francois-lavet_combined_2019} which introduce an additional loss term based on cosine similarity to encourage the predicted abstract state change to align with a chosen embedding vector. This regularization arguably drives the abstract state to be more meaningful and generalisable, and thereupon may enable more efficient planning.

More endemic interpretability-oriented regularizers have been proposed, with first-order logic (in DL, with \cite{serafini_logic_2016}), or tree-regularizers \citep{wu_optimizing_2019}.
The (regional) tree-regularization \npw{proposed by} \cite{wu_optimizing_2019} aims to specifically learn deep policy networks whose decision boundaries are well approximated by small decision tree(s), hence targets human-\pw{simulatability}.
By considering interpretability \pw{from} the very start---\pw{in contrast to} indirect approaches aiming at approximating a black-box policy network with a decision tree a posteriori---it should be more accessible to reach both good performance and simulatability, due to the \textit{multiple optima} property of deep network. \cg{Indeed, indirect approaches may be unreliable as the original unregularized black box NN has no incentive to be simulatable or decomposable.}

\paragraph{Conclusion}
Embedding \pw{interpretability} bias through regularizers has the advantage to be easily integrated with any optimization algorithm\pw{, such as stochastic gradient descent, if the hypothesis class is made differentiable.} 
As deep models have---infamously---multiple optima of similar predictive accuracy \citep{goodfellow_deep_2016}, we can hope that using interpretability-oriented regularizers may not impact much the performance, if convex.
However, since such approaches do not restrict the search space per se, they do not provide interpretability guarantee.

There are a few noticeable studies in deep learning aiming to distil logical knowledge through loss functions and regularizers during the neural network training, \pw{such as} \citep{donadello_logic_2017,serafini_logic_2016,diligenti_semantic-based_2017,wang_integrating_2019,rocktaschel_injecting_2015,demeester_lifted_2016,xu_semantic_2018}\footnote{For instance,  \cite{serafini_logic_2016} uses FOL-based loss-function to \pw{constrain} the learned semantic representations to be logically consistent.}
or \citep{minervini_adversarial_2017} with adversarial training. \cg{Bridging the gap between XRL and interpretable literature, \cite{plumb_regularizing_2020} proposes some explanability-regularisers, differentiable, and model agnostic, which would encourage the learned models, trained end-to-end, to be well explainable.}
Although intelligibility-enhancers seem numerous, the question of defining specific regularizers leading to a reasonably-interpretable decision-making in complex \pw{environments} is far from being obvious.

\begin{figure}[t]
\centering

\includegraphics[bb=0 0 128 96]{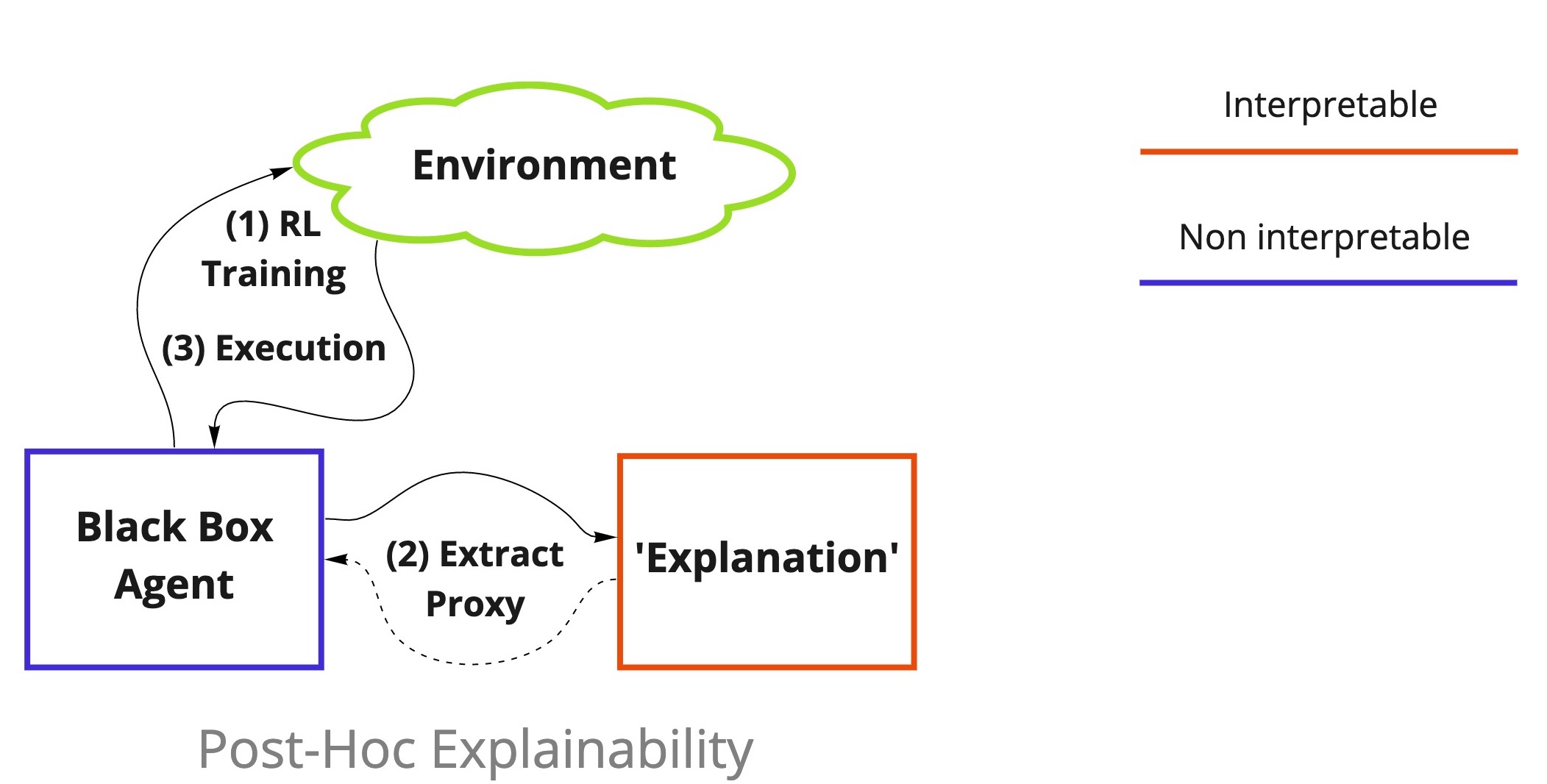}
\caption{Illustration for explainable RL approaches. Dashed  lines represent optional links, depending on the methods.}
\label{fig:explainable RL}
\end{figure}

\section{Explainable RL} \label{sec:XRL}

Although the focus of this survey is on interpretable RL, we also provide a \cg{succinct} overview of explainable RL (XRL) for completeness and \pw{in order} to contrast it with the work in interpretable RL.
\Cref{fig:explainable RL} illustrates the high-level procedure in XRL, which can be contrasted with the approaches for interpretable decision-making described in \Cref{fig:interpretable decision making}.
A more thorough discussion on XRL can be found in the recent surveys by \cite{alharin_reinforcement_2020} or \cite{heuillet_explainability_2021}.
\pw{A summary of the work discussed in this section can be found in \Cref{tab:xrl}.}

\begin{table}[t]
\footnotesize
    \centering
    \caption{Overview of approaches in explainable RL}
    \label{tab:xrl}
    
    \begin{tabular}{@{}llll@{}}
    \toprule
        Type & Approach & Refs \\
    \midrule
        Visual & t-SNE, saliency map from Jacobian & \citep{zahavy_graying_2016} \\
        Visual & Saliency map from perturbation &  \citep{greydanus_visualizing_2018} \\
        Visual & Saliency map by balancing specificity and relevance & \citep{gupta_explain_2020} \\
        Visual & SHAP & \citep{wang_attribution-based_2020} \\
        Visual & Attention mask & \citep{shi_self-supervised_2020} \\
        Visual & Attention mask with information bottleneck & \citep{kim_attentional_2020} \\
        Visual & Summary from history & \citep{sequeira_interestingness_2020} \\
        
        Textual & State predicates & \citep{hayes_improving_2017} \\
        Textual & State and outcome predicates & \citep{van_der_waa_contrastive_2018} \\
        Textual & Reuse of provided instructions & \citep{fukuchi_autonomous_2017} \\
        
        Causal & Causal model & \citep{madumal_explainable_2020} \\
        Causal & Opportunity chain & \citep{madumal_distal_2020} \\
        
        Policy & Soft decision tree & \citep{coppens_distilling_2019} \\
        Policy & Decision tree & \citep{bewley_tripletree_2021} \\
        
        Other & Reward decomposition & \citep{juozapaitis_explainable_2019} \\
        Other & Markov chain on abstract state space & \citep{topin_generation_2019} \\
        Other & Probability of success, \# steps to reach goal & \citep{cruz_memory-based_2019} \\
    \bottomrule
    \end{tabular}
\end{table}

The goal in XRL is to provide some explanations \pw{regarding} an RL agent's decisions, \cg{e.g., \pw{highlighting} the main features which influenced a decision and their importance.}
This is commonly done via a post-hoc \cg{and often model-agnostic} 
procedure after a black-\pw{box} model is already trained, \cg{which usually only aims to offer a functional understanding.}
\cg{Many contextual parameters should be taken into consideration when defining what constitutes a "good" explanation 
for a scenario, e.g., the background knowledge and levels of expertise of the addressee of this explanation, their needs and expectations, but also---often neglected---the time available to them.}
Explanations can take various forms:

\paragraph{Visual explanation} 
Using the DQN algorithm, \cite{zahavy_graying_2016} builds two graphical representations in order to analyze the decisions made by the DQN network: (1) t-SNE maps \citep{maaten_visualizing_2008} from the activations of the last hidden layer of the network and (2) saliency maps from the Jacobian of the network. 
Motivated by the limitations of Jacobian saliency maps, \cite{greydanus_visualizing_2018} proposes to build saliency maps using a perturbation-based approach, which provides information about the importance of a perturbed region.
Continuing this line of research, \cite{gupta_explain_2020} introduces the idea of balancing specificity and relevance in order to build saliency maps to highlight more relevant regions.
In order to take into account non-visual inputs as well, \cite{wang_attribution-based_2020} extends a generic explanation technique called SHAP  (SHapley  Additive  exPlanation)  \citep{lundberg_unified_2017} to select important features for RL.
Another approach is based on attention mechanisms. \cite{shi_self-supervised_2020} propose to learn attention masks in a self-supervised way to highlight information important for a decision.
\npw{In \cite{kim_attentional_2020}, attention is further combined} with an information bottleneck mechanism in order to generate sparser attention maps.
Using a different kind of explanation, \cite{sequeira_interestingness_2020} investigates the use of visual summaries extracted from histories to explain an agent's behavior.

\paragraph{Textual explanation} 
\npw{The work of \cite{hayes_improving_2017} generates} explanations for choosing an action by finding state predicates that co-occur with that action.
Inspired by that approach, \cite{van_der_waa_contrastive_2018} extend it by introducing outcome predicates and provide contrastive explanations using both state and outcome predicates.
In a setting where the agent learns from instructions given by a human tutor, \cite{fukuchi_autonomous_2017} proposes to explain the agent's decisions by reusing the provided instructions.

\paragraph{Causal explanation}
\npw{In the proposition of \cite{madumal_explainable_2020}, a causal model is learned} from a given graph of causal relation in order to generate contrastive explanations of action choices.
Building on this work, \cite{madumal_distal_2020} instead generates explanations based on potential future actions using the concept of opportunity chains, which include information of what is enabled or caused by an action.

\paragraph{Interpretable policy}
Some work tries to obtain a more intelligible policy in order to explain a trained RL agent using, e.g.,
soft decision trees \citep{coppens_distilling_2019}, or
decision trees
\citep{bewley_tripletree_2021}.
Note that the indirect approach for interpretability \cg{(as presented in \Cref{sec:indirect})} should not be confused with this approach for post-hoc explainability.
In the latter case, a more intelligible policy is learned to explain a black-box policy that is used as the proper controller.
In contrast, in the former case, a more interpretable policy is learned to be used as the final controller that replaces the intermediate black-box policy, \pw{which therefore does not need to be explained anymore}.
Therefore, in the latter case, it is important that the intelligible policy mimics the black-box policy well, while in the former, the performance of the interpretable policy is more important that its ability to mimic the black-box policy.
\cg{When learning such an interpretative policy, there is a tradeoff between the intelligibility of the explanatory policy and the fidelity of the approximation which has to be balanced.
One common drawback is that such an approximation may be valid only on a restricted domain.
}

\paragraph{Other}
\npw{Besides,} \cite{juozapaitis_explainable_2019} proposes to learn a vector Q-function, where each component corresponds to a given attribute called reward type.
This decomposition of the Q-function is then used to explain preferences between actions.
In contrast, \npw{in \cite{topin_generation_2019}}, a policy is explained with a Markov chain built on an abstract state space.
In addition, in goal-oriented RL, \cite{cruz_memory-based_2019} justifies an action choice based on its probability of success and the number of time steps to reach the goal.

\paragraph{Conclusion}
Most work we discussed takes the target audience of the explanations to be the end-user.
Even in this case, explanations can take multiple forms.
Thus, it can be presented to the user in different modes (e.g., visual, textual, tabular,...) and it can be either local or global.
\cg{Beyond their forms, explanations may also answer intelligibility queries of different nature and granularity: certainty, contextual, case-based or analogies
, contrastive, counterfactual (“what if"), simulation-based (consequences), trace/steps, why not, etc
 (e.g., \shortcite{chari_directions_2020,lim_why_2019,mittelstadt_explaining_2019}).}
\pw{Hence}, an explanation can be used to clarify, justify, or rationalize an action choice.
We recommend that future work on explainable RL make those aspects clear\cg{, s}ince this information would impact how an explanation technique should be evaluated \cg{and taken into consideration}.


\pw{One} issue with post-hoc explanation approaches is that while the generated explanation may seem to make sense, it may in fact be specious \cg{(e.g., \shortcite{atrey_exploratory_2020} for saliency maps)} and may not reflect the true inner working of the model.
While this may not be an issue if the explanation is used as a tool to justify an action choice to a user, this is problematic for understanding the decision-making process.
Note this issue does not occur if an interpretable policy is used for decision-making. 

\cg{While the explainable and interpretable literature refers to usually divergent approaches, some recent work aimed at bridging this gap, (e.g., \shortcite{plumb_regularizing_2020} previously mentioned in \Cref{sec:preference bias}). Through regularizers, it gracefully integrates explanability considerations during the training of the model. It stands at odds with traditional XRL literature, assuming they could extract a posteriori explanations, without any incentive for the model to be intelligible.}

\section{Open Problems \pw{and Research Directions}} \label{sec:open}



Before concluding this survey, we discuss a selection of open problems, which we \cg{regard as essential within the quest for interpretable RL}.

\paragraph{Full Interpretability in RL}
The work we have reviewed falls in various intermediate levels on the interpretability scale, some being more interpretable than others for different RL components.
Moreover, few deep RL \pw{approaches accepting} high-dimensional inputs, if any, can achieve full interpretability, i.e., interpretable inputs, interpretable models, and interpretable decision-making.
Designing a fully-interpretable RL method with a high-degree of interpretability seems not to be achievable with the current methods, especially for complex tasks like autonomous driving, although such tasks calls for such methods.
Thus, for deep RL to be considered as a practical method to solve those difficult tasks, fully interpretable RL methods must be developed for all the RL components.
Given the complexity of those tasks, this may only be achievable by abstraction and composition in the programming language sense, where interpretable methods can be composed to solve more difficult problems.

\paragraph{Interpretability vs Performance} 
A common held opinion is that using a more transparent model or approach impacts negatively the final performance \citep{ribeiro_model-agnostic_2016}.
In the light of impressive results achieved by deep learning methods, this opinion seems hard to be challenged.
However, some different voices \citep{rudin_secrets_2019} suggest that black-box models like those based on deep learning may not always be needed and that in some domains simple models should be favored and can obtain excellent performance without the drawbacks of deep learning methods.
Similar remarks have also been made in deep \pw{RL} by \cite{mania_simple_2018}, who showed that simple linear models with stochastic search can fare well against more advanced deep RL methods.
Extrapolating those observations, one may wonder if this could be achieved with all aspects of RL (inputs, models, decision-making) and if interpretability can be considered as a regularization technique, which would bring more transparency obviously, but also larger generalizability.

\paragraph{Interpretability vs Scalability}
\cg{In addition to the challenge of designing a fully-interpre\-table RL method}, running such a method \pw{in order to learn a fully-interpretable solution} would probably be also very costly in terms of computation. 
Indeed, for decision-making for instance, learning an interpretable policy corresponds to a task similar to program synthesis \citep{Gulwani_program_2017}, which is known to be an NP-hard problem.
Therefore, there may be a trade-off between the degree of interpretability one may want to achieve and the scalability of the interpretable algorithm.
This question is crucial to investigate as the research moves to more and more interpretable methods, critically needed for high-stake tasks.

\paragraph{Evaluation of Interpretability and Explainability}
We finish this discussion by a more classic question that has been \cg{frequently} raised within XAI, but that we mention here due to its importance.
Given the various meanings of interpretability and explainability and more precisely the various purposes they can serve, there is no common ground for the definition of good evaluation metrics for XAI in general, but also for interpretable and explainable RL.
For interpretability, is there a good metric for deciding if one model is more interpretable than another?
For explainability, is it possible to evaluate what constitutes a good 
explanation \cg{in a specific context}?
This state of affairs prevents a comparative evaluation of the different methods that have been proposed, which also impedes the rapid progress in this research direction.
While achieving more precise definitions for interpretability and explainability can help, evaluation \pw{metrics and} protocols could be proposed depending on precise goals regarding ethical, legal, operational, or usability concerns, which may help them to be adopted by the research community.


\section{Conclusion} \label{sec:conclusion}

We surveyed recent work in reinforcement learning (RL) related to the important concern of interpretability \pw{(and its related notion of explainability).}
We proposed a definition of interpretability in RL, which contrary to the general setting of explainable artificial intelligence, leads to different levels of transparency in the components that play a role in RL.
In particular, we first discussed the studies that focus on interpretable inputs (e.g., observations, but possibly other types of information).
Moreover, we provided an overview of approaches that deal with learning an interpretable transition model, which is significant for interpretable model-based RL, but also those that deal with learning an interpretable preference model, which is fundamental to justify action selection.
Then, we surveyed the papers on learning interpretable policies, which constitute arguably the most critical part of interpretable RL.
\pw{For completeness}, we also provided a short review of work related to post-hoc explainability.
\pw{Finally, we highlighted a few open problems and future research directions that we deemed as particularly relevant.}

\cg{Although concerns around the ethical implications of algorithmic and automation deployment are nothing new \citep{wiener_human_1954}, the field of AI ethics \pw{still} seems \pw{at} its infancy as we begin to witness the extent of the influence and impact that these systems may have on our societal fabric when deployed.
In this regard, a responsible practice for the design, implementation, use, and monitoring/auditing of AI-driven systems is greatly impeded by the non-intelligibity of current algorithms.}
As RL-based systems become more widespread, questions related to interpretability \cg{become consequently increasingly pressing}.
One could \cg{even} contend that interpretable RL is one of the key deadlocks to overcome to make RL a more \cg{functional} method for being deployed in real-life
While it may be hard to achieve a fully intelligible RL model, one may envision hierarchical RL approaches where some parts may not be completely transparent\cg{---e.g., at the low-level---}but a maximum of other parts\cg{---e.g., at the subgoal level---}are thoroughly interpretable.

\cg{While the act of opening up the blackbox do not suffice to instantly disclose a thorough understanding of its social implications---since we \textit{``need to look across the system, rather than merely inside''} (as noted by \shortcite{ananny_seeing_2018})---algorithmic intelligibility appears as a promising step towards further \textit{algorithmic accountability} and more trustworthy AI. We encourage the curious reader to look further at the generous work of other researchers investigating these tangent questions (such as \shortcite{crawford_ai_2016,raji_closing_2020,daly_artificial_2019,yu_building_2018,european_commission_ethics_2019} to mention only a few)}.

\footnotesize
\bibliographystyle{IEEEtranN}
\bibliography{InterpretableRL}

\begin{thebibliography}{261}
\providecommand{\natexlab}[1]{#1}
\providecommand{\url}[1]{#1}
\csname url@samestyle\endcsname
\providecommand{\newblock}{\relax}
\providecommand{\bibinfo}[2]{#2}
\providecommand{\BIBentrySTDinterwordspacing}{\spaceskip=0pt\relax}
\providecommand{\BIBentryALTinterwordstretchfactor}{4}
\providecommand{\BIBentryALTinterwordspacing}{\spaceskip=\fontdimen2\font plus
\BIBentryALTinterwordstretchfactor\fontdimen3\font minus
  \fontdimen4\font\relax}
\providecommand{\BIBforeignlanguage}[2]{{%
\expandafter\ifx\csname l@#1\endcsname\relax
\typeout{** WARNING: IEEEtranN.bst: No hyphenation pattern has been}%
\typeout{** loaded for the language `#1'. Using the pattern for}%
\typeout{** the default language instead.}%
\else
\language=\csname l@#1\endcsname
\fi
#2}}
\providecommand{\BIBdecl}{\relax}
\BIBdecl

\bibitem[Sutton and Barto(2018)]{sutton_reinforcement_2018}
R.~S. Sutton and A.~G. Barto, ``Reinforcement learning: An introduction,''
  2018.

\bibitem[Silver et~al.(2017)Silver, Schrittwieser, Simonyan, Antonoglou, Huang,
  Guez, Hubert, Baker, Lai, Bolton, Chen, Lillicrap, Hui, Sifre, {van den
  Driessche}, Graepel, and Hassabis]{silver_mastering_2017}
D.~Silver, J.~Schrittwieser, K.~Simonyan, I.~Antonoglou, A.~Huang, A.~Guez,
  T.~Hubert, L.~Baker, M.~Lai, A.~Bolton, Y.~Chen, T.~Lillicrap, F.~Hui,
  L.~Sifre, G.~{van den Driessche}, T.~Graepel, and D.~Hassabis, ``Mastering
  the game of {{Go}} without human knowledge,'' \emph{Nature}, vol. 550, pp.
  354--359, 2017.

\bibitem[Vinyals et~al.(2019)Vinyals, Babuschkin, Czarnecki, Mathieu, Dudzik,
  Chung, Choi, Powell, Ewalds, Georgiev, Oh, Horgan, Kroiss, Danihelka, Huang,
  Sifre, Cai, Agapiou, Jaderberg, Vezhnevets, Leblond, Pohlen, Dalibard,
  et~al.]{vinyals_grandmaster_2019}
O.~Vinyals, I.~Babuschkin, W.~M. Czarnecki, M.~Mathieu, A.~Dudzik, J.~Chung,
  D.~H. Choi, R.~Powell, T.~Ewalds, P.~Georgiev, J.~Oh, D.~Horgan, M.~Kroiss,
  I.~Danihelka, A.~Huang, L.~Sifre, T.~Cai, J.~P. Agapiou, M.~Jaderberg, A.~S.
  Vezhnevets, R.~Leblond, T.~Pohlen, V.~Dalibard \emph{et~al.}, ``Grandmaster
  level in {StarCraft II} using multi-agent reinforcement learning,''
  \emph{Nature}, vol. 575, no. 7782, pp. 350--354, 2019.

\bibitem[OpenAI et~al.(2019)OpenAI, Akkaya, Andrychowicz, Chociej, Litwin,
  McGrew, Petron, Paino, Plappert, Powell, Ribas, Schneider, Tezak, Tworek,
  Welinder, Weng, Yuan, Zaremba, and Zhang]{openai_solving_2019}
OpenAI, I.~Akkaya, M.~Andrychowicz, M.~Chociej, M.~Litwin, B.~McGrew,
  A.~Petron, A.~Paino, M.~Plappert, G.~Powell, R.~Ribas, J.~Schneider,
  N.~Tezak, J.~Tworek, P.~Welinder, L.~Weng, Q.~Yuan, W.~Zaremba, and L.~Zhang,
  ``Solving {{Rubik}}'s {{Cube}} with a {{Robot Hand}},'' \emph{arXiv:
  1910.07113}, 2019.

\bibitem[Zahavy et~al.(2016)Zahavy, {Ben-Zrihem}, and
  Mannor]{zahavy_graying_2016}
T.~Zahavy, N.~{Ben-Zrihem}, and S.~Mannor, ``Graying the black box:
  {{Understanding DQNs}},'' in \emph{ICML}, 2016.

\bibitem[Henderson et~al.(2018)Henderson, Islam, Bachman, Pineau, Precup, and
  Meger]{henderson_deep_2018}
P.~Henderson, R.~Islam, P.~Bachman, J.~Pineau, D.~Precup, and D.~Meger, ``Deep
  reinforcement learning that matters,'' in \emph{AAAI}, 2018.

\bibitem[Zhang et~al.(2018{\natexlab{a}})Zhang, Vinyals, Munos, and
  Bengio]{zhang_study_2018}
C.~Zhang, O.~Vinyals, R.~Munos, and S.~Bengio, ``A {{Study}} on {{Overfitting}}
  in {{Deep Reinforcement Learning}},'' \emph{arXiv: 1804.06893}, 2018.

\bibitem[Huang et~al.(2017)Huang, Papernot, Goodfellow, Duan, and
  Abbeel]{huang_adversarial_2017}
S.~Huang, N.~Papernot, I.~Goodfellow, Y.~Duan, and P.~Abbeel, ``Adversarial
  {{Attacks}} on {{Neural Network Policies}},'' in \emph{ICLR Workshop}, 2017.

\bibitem[Crawford et~al.(2016)Crawford, Dobbe, Dryer, Fried, Green, Kaziunas,
  Kak, Mathur, McElroy, S{\'a}nchez, Raji, Rankin, Richardson, Schultz, West,
  and Whittaker]{crawford_ai_2016}
K.~Crawford, R.~Dobbe, T.~Dryer, G.~Fried, B.~Green, E.~Kaziunas, A.~Kak,
  V.~Mathur, E.~McElroy, A.~N. S{\'a}nchez, D.~Raji, J.~L. Rankin,
  R.~Richardson, J.~Schultz, S.~M. West, and M.~Whittaker, ``{{AI Now
  Report}},'' {AI Now Institute}, Tech. Rep., 2016.

\bibitem[Yu et~al.(2018)Yu, Shen, Miao, Leung, Lesser, and
  Yang]{yu_building_2018}
H.~Yu, Z.~Shen, C.~Miao, C.~Leung, V.~R. Lesser, and Q.~Yang, ``Building ethics
  into artificial intelligence,'' in \emph{IJCAI}, 2018.

\bibitem[Leslie(2020)]{leslie_understanding_2020}
D.~Leslie, ``Understanding artificial intelligence ethics and safety: {{A}}
  guide for the responsible design and implementation of {{AI}} systems in the
  public sector,'' \emph{SSRN Electronic Journal}, 2020.

\bibitem[Morley et~al.(2020)Morley, Floridi, Kinsey, and
  Elhalal]{morley_what_2020}
J.~Morley, L.~Floridi, L.~Kinsey, and A.~Elhalal, ``From what to how: {{An}}
  initial review of publicly available {{AI}} ethics tools, methods and
  research to translate principles into practices,'' \emph{Science and
  Engineering Ethics}, vol.~26, no.~4, 2020.

\bibitem[Lo~Piano(2020)]{lo_piano_ethical_2020}
S.~Lo~Piano, ``Ethical principles in machine learning and artificial
  intelligence: Cases from the field and possible ways forward,''
  \emph{Humanities and Social Sciences Communications}, vol.~7, no.~1, pp.
  1--7, 2020.

\bibitem[Dwork et~al.(2012)Dwork, Hardt, Pitassi, Reingold, and
  Zemel]{dwork_fairness_2012}
C.~Dwork, M.~Hardt, T.~Pitassi, O.~Reingold, and R.~Zemel, ``Fairness through
  awareness,'' in \emph{ICTS}, 2012.

\bibitem[Friedler et~al.(2021)Friedler, Scheidegger, and
  Venkatasubramanian]{friedler_impossibility_2021}
S.~A. Friedler, C.~Scheidegger, and S.~Venkatasubramanian, ``The
  ({{Im}})possibility of fairness: Different value systems require different
  mechanisms for fair decision making,'' \emph{Communications of the ACM},
  vol.~64, no.~4, pp. 136--143, 2021.

\bibitem[Mehrabi et~al.(2019)Mehrabi, Morstatter, Saxena, Lerman, and
  Galstyan]{mehrabi_survey_2019}
N.~Mehrabi, F.~Morstatter, N.~Saxena, K.~Lerman, and A.~Galstyan, ``A
  {{Survey}} on {{Bias}} and {{Fairness}} in {{Machine Learning}},''
  \emph{arXiv: 1908.09635}, 2019.

\bibitem[Amodei et~al.(2016)Amodei, Olah, Steinhardt, Christiano, Schulman, and
  Man{\'e}]{amodei_concrete_2016}
D.~Amodei, C.~Olah, J.~Steinhardt, P.~Christiano, J.~Schulman, and D.~Man{\'e},
  ``Concrete {{Problems}} in {{AI Safety}},'' \emph{arXiv: 1606.06565}, 2016.

\bibitem[{Doshi-Velez} et~al.(2019){Doshi-Velez}, Kortz, Budish, Bavitz,
  Gershman, O'Brien, Scott, Schieber, Waldo, Weinberger, Weller, and
  Wood]{doshi-velez_accountability_2019}
F.~{Doshi-Velez}, M.~Kortz, R.~Budish, C.~Bavitz, S.~Gershman, D.~O'Brien,
  K.~Scott, S.~Schieber, J.~Waldo, D.~Weinberger, A.~Weller, and A.~Wood,
  ``Accountability of {{AI Under}} the {{Law}}: {{The Role}} of
  {{Explanation}},'' \emph{arXiv: 1711.01134}, 2019.

\bibitem[Commission(2019)]{european_commission_ethics_2019}
E.~Commission, ``Ethics guidelines for trustworthy {{AI}},''
  https://ec.europa.eu/digital-single-market/en/news/ethics-guidelines-trustworthy-ai,
  2019.

\bibitem[Horvitz and Mulligan(2015)]{horvitz_data_2015}
E.~Horvitz and D.~Mulligan, ``Data, privacy, and the greater good,''
  \emph{Science}, vol. 349, no. 6245, pp. 253--255, 2015.

\bibitem[Bonnefon et~al.(2019)Bonnefon, Shariff, and
  Rahwan]{bonnefon_trolley_2019}
J.~Bonnefon, A.~Shariff, and I.~Rahwan, ``The {{Trolley}}, {{The Bull Bar}},
  and {{Why Engineers Should Care About The Ethics}} of {{Autonomous Cars}}
  [point of view],'' \emph{Proceedings of the IEEE}, vol. 107, no.~3, 2019.

\bibitem[Mohseni et~al.(2020)Mohseni, Zarei, and
  Ragan]{mohseni_multidisciplinary_2020}
S.~Mohseni, N.~Zarei, and E.~D. Ragan, ``A {{Multidisciplinary Survey}} and
  {{Framework}} for {{Design}} and {{Evaluation}} of {{Explainable AI
  Systems}},'' \emph{arXiv: 1811.11839}, 2020.

\bibitem[Whittlestone et~al.(2021)Whittlestone, Arulkumaran, and
  Crosby]{whittlestone_societal_2021}
J.~Whittlestone, K.~Arulkumaran, and M.~Crosby, ``The {{Societal Implications}}
  of {{Deep Reinforcement Learning}},'' \emph{JAIR}, vol.~70, pp. 1003--1030,
  2021.

\bibitem[Puiutta and Veith(2020)]{puiutta_explainable_2020}
E.~Puiutta and E.~M. Veith, ``Explainable reinforcement learning: {{A}}
  survey,'' in \emph{LNCS}, 2020.

\bibitem[Alharin et~al.(2020)Alharin, Doan, and
  Sartipi]{alharin_reinforcement_2020}
A.~Alharin, T.-N. Doan, and M.~Sartipi, ``Reinforcement {{Learning
  Interpretation Methods}}: {{A Survey}},'' \emph{IEEE Access}, vol.~8, pp.
  171\,058--171\,077, 2020.

\bibitem[Heuillet et~al.(2021)Heuillet, Couthouis, and
  {D{\'i}az-Rodr{\'i}guez}]{heuillet_explainability_2021}
A.~Heuillet, F.~Couthouis, and N.~{D{\'i}az-Rodr{\'i}guez}, ``Explainability in
  deep reinforcement learning,'' \emph{Knowledge-Based Systems}, vol. 214,
  2021.

\bibitem[Puterman(1994)]{puterman94}
M.~Puterman, \emph{Markov decision processes: discrete stochastic dynamic
  programming}, 1994.

\bibitem[Mnih et~al.(2015)Mnih, Kavukcuoglu, Silver, Rusu, Veness, Bellemare,
  Graves, Riedmiller, Fidjeland, and {Others}]{mnih_human-level_2015}
V.~Mnih, K.~Kavukcuoglu, D.~Silver, A.~A. Rusu, J.~Veness, M.~G. Bellemare,
  A.~Graves, M.~Riedmiller, A.~K. Fidjeland, and {Others}, ``Human-level
  control through deep reinforcement learning,'' \emph{Nature}, vol. 518, pp.
  529--533, 2015.

\bibitem[Schulman et~al.(2017)Schulman, Wolski, Dhariwal, Radford, and
  Klimov]{schulman_proximal_2017}
J.~Schulman, F.~Wolski, P.~Dhariwal, A.~Radford, and O.~Klimov, ``Proximal
  {{Policy Optimization Algorithms}},'' \emph{arXiv: 1707.06347}, 2017.

\bibitem[Fujimoto et~al.(2018)Fujimoto, Hoof, and
  Meger]{fujimoto_addressing_2018}
S.~Fujimoto, H.~Hoof, and D.~Meger, ``Addressing {{Function Approximation
  Error}} in {{Actor}}-{{Critic Methods}},'' in \emph{ICML}, 2018.

\bibitem[Haarnoja et~al.(2018)Haarnoja, Zhou, Abbeel, and
  Levine]{haarnoja_soft_2018}
T.~Haarnoja, A.~Zhou, P.~Abbeel, and S.~Levine, ``Soft {{Actor}}-{{Critic}}:
  {{Off}}-{{Policy Maximum Entropy Deep Reinforcement Learning}} with a
  {{Stochastic Actor}},'' in \emph{ICML}, 2018.

\bibitem[{Francois-Lavet} et~al.(2019){Francois-Lavet}, Bengio, Precup, and
  Pineau]{francois-lavet_combined_2019}
V.~{Francois-Lavet}, Y.~Bengio, D.~Precup, and J.~Pineau, ``Combined
  reinforcement learning via abstract representations,'' in \emph{AAAI}, 2019.

\bibitem[Veerapaneni et~al.(2020)Veerapaneni, {Co-Reyes}, Chang, Janner, Finn,
  Wu, Tenenbaum, and Levine]{veerapaneni_entity_2020}
R.~Veerapaneni, J.~D. {Co-Reyes}, M.~Chang, M.~Janner, C.~Finn, J.~Wu,
  J.~Tenenbaum, and S.~Levine, ``Entity abstraction in visual model-based
  reinforcement learning,'' in \emph{CoRL}, 2020.

\bibitem[Scholz et~al.(2014)Scholz, Levihn, Isbell, Wingate, and
  Wingate]{scholz_physics-based_2014}
J.~Scholz, M.~Levihn, C.~L. Isbell, D.~Wingate, and D.~Wingate, ``A
  {{Physics}}-{{Based Model Prior}} for {{Object}}-{{Oriented MDPs}},'' in
  \emph{ICML}, 2014.

\bibitem[Barto and Mahadevan(2003)]{barto_recent_2003}
A.~G. Barto and S.~Mahadevan, ``Recent {{Advances}} in {{Hierarchical
  Reinforcement Learning}},'' \emph{Discrete Event Dynamic Systems}, 2003.

\bibitem[Ribeiro et~al.(2016{\natexlab{a}})Ribeiro, Singh, and
  Guestrin]{ribeiro_model-agnostic_2016}
M.~T. Ribeiro, S.~Singh, and C.~Guestrin, ``Model-{{Agnostic Interpretability}}
  of {{Machine Learning}},'' in \emph{ICML Workshop on {{Human
  Interpretability}} in {{ML}}}, 2016.

\bibitem[Miller(2019)]{miller_explanation_2019}
T.~Miller, ``Explanation in {{Artificial Intelligence}}: {{Insights}} from the
  {{Social Sciences}},'' \emph{Artificial Intelligence}, 2019.

\bibitem[Molnar(2019)]{molnar_interpretable_2019}
C.~Molnar, ``Interpretable {{Machine Learning}}: {{A Guide}} for {{Making Black
  Box Models Explainable}},'' 2019.

\bibitem[Lipton(2017)]{lipton_mythos_2017}
Z.~C. Lipton, ``The {{Mythos}} of {{Model Interpretability}},''
  \emph{arXiv:1606.03490}, 2017.

\bibitem[Barredo~Arrieta et~al.(2020)Barredo~Arrieta, {D{\'i}az-Rodr{\'i}guez},
  Ser, Bennetot, Tabik, Barbado, Garcia, {Gil-Lopez}, Molina, Benjamins,
  Chatila, and Herrera]{barredo_arrieta_explainable_2020}
A.~Barredo~Arrieta, N.~{D{\'i}az-Rodr{\'i}guez}, J.~D. Ser, A.~Bennetot,
  S.~Tabik, A.~Barbado, S.~Garcia, S.~{Gil-Lopez}, D.~Molina, R.~Benjamins,
  R.~Chatila, and F.~Herrera, ``Explainable {{Artificial Intelligence}}
  ({{XAI}}): {{Concepts}}, {{Taxonomies}}, {{Opportunities}} and {{Challenges}}
  toward {{Responsible AI}},'' \emph{Information Fusion}, vol.~58, pp. 82--115,
  2020.

\bibitem[Chari et~al.(2020)Chari, Gruen, Seneviratne, and
  McGuinness]{chari_directions_2020}
S.~Chari, D.~M. Gruen, O.~Seneviratne, and D.~L. McGuinness, ``Directions for
  {{Explainable Knowledge}}-{{Enabled Systems}},'' \emph{arXiv: 2003.07523},
  2020.

\bibitem[Gilpin et~al.(2019)Gilpin, Bau, Yuan, Bajwa, Specter, and
  Kagal]{gilpin_explaining_2019}
L.~H. Gilpin, D.~Bau, B.~Z. Yuan, A.~Bajwa, M.~Specter, and L.~Kagal,
  ``Explaining {{Explanations}}: {{An Overview}} of {{Interpretability}} of
  {{Machine Learning}},'' in \emph{DSAA}, 2019.

\bibitem[P{\'a}ez(2019)]{paez_pragmatic_2019}
A.~P{\'a}ez, ``The {{Pragmatic Turn}} in {{Explainable Artificial
  Intelligence}} ({{XAI}}),'' \emph{Minds and Machines}, vol.~29, no.~3, pp.
  441--459, 2019.

\bibitem[Rudin(2019)]{rudin_stop_2019}
C.~Rudin, ``Stop explaining black box machine learning models for high stakes
  decisions and use interpretable models instead,'' \emph{Nature Machine
  Intelligence}, vol.~1, no.~5, pp. 206--215, 2019.

\bibitem[Penkov and Ramamoorthy(2019)]{penkov_learning_2019}
S.~Penkov and S.~Ramamoorthy, ``Learning {{Programmatically Structured
  Representations}} with {{Perceptor Gradients}},'' in \emph{ICLR}, 2019.

\bibitem[Degris et~al.(2006)Degris, Sigaud, and
  Wuillemin]{degris_learning_2006}
T.~Degris, O.~Sigaud, and P.~H. Wuillemin, ``Learning the structure of factored
  {{Markov}} decision processes in reinforcement learning problems,'' in
  \emph{ICML}, 2006.

\bibitem[Dzeroski et~al.(1998)Dzeroski, Raedt, and
  Blockeel]{dzeroski_relational_1998}
S.~Dzeroski, L.~D. Raedt, and H.~Blockeel, ``Relational reinforcement
  learning,'' in \emph{{{ICML}}}, 1998.

\bibitem[Barwise(1977)]{barwise_introduction_1977}
J.~Barwise, ``An introduction to first-order logic,'' in \emph{Studies in Logic
  and the Foundations of Mathematics}, 1977, vol.~90, pp. 5--46.

\bibitem[Swain(2013)]{swain_knowledge_2013}
M.~Swain, ``Knowledge {{Representation}},'' in \emph{Encyclopedia of {{Systems
  Biology}}}, 2013, pp. 1082--1084.

\bibitem[Slaney and Thi{\'e}baux(2001)]{slaney_blocks_2001}
J.~Slaney and S.~Thi{\'e}baux, ``Blocks {{World Revisited}},'' \emph{Artificial
  Intelligence}, vol. 125, no. 1--2, pp. 119--153, 2001.

\bibitem[Guestrin et~al.(2003)Guestrin, Koller, Gearhart, and
  Kanodia]{guestrin_generalizing_2003}
C.~Guestrin, D.~Koller, C.~Gearhart, and N.~Kanodia, ``Generalizing plans to
  new environments in relational {{MDPs}},'' in \emph{IJCAI}, 2003.

\bibitem[D{\v z}eroski et~al.(2001)D{\v z}eroski, De~Raedt, and
  Driessens]{dzeroski_relational_2001}
S.~D{\v z}eroski, L.~De~Raedt, and K.~Driessens, ``Relational {{Reinforcement
  Learning}},'' \emph{Machine Learning}, vol.~43, no.~1, pp. 7--52, 2001.

\bibitem[Driessens and Blockeel(2001)]{driessens_learning_2001}
Driessens and H.~Blockeel, ``Learning {{Digger}} using {{Hierarchical
  Reinforcement Learning}} for {{Concurrent Goals}}.'' in \emph{{EWRL}}, 2001.

\bibitem[Cole et~al.(2003)Cole, Lloyd, and Ng]{cole_symbolic_2003}
J.~Cole, J.~Lloyd, and K.~S. Ng, ``Symbolic {{Learning}} for {{Adaptive
  Agents}},'' in \emph{Annual {{Partner Conference}}}, 2003.

\bibitem[Walker et~al.(2004)Walker, Shavlik, and
  Maclin]{walker_relational_2004}
T.~Walker, J.~Shavlik, and R.~Maclin, ``Relational {{Reinforcement Learning}}
  via {{Sampling}} the {{Space}} of {{First}}-{{Order Conjunctive Features}},''
  in \emph{{{ICML}} Workshop on {{Relational Reinforcement Learning}}}, 2004.

\bibitem[Sanner(2005)]{sanner_simultaneous_2005}
S.~Sanner, ``Simultaneous {{Learning}} of {{Structure}} and {{Value}} in
  {{Relational Reinforcement Learning}},'' in \emph{ICML Workshop on {{Rich
  Representations}} for RL}, 2005.

\bibitem[Driessens et~al.(2006)Driessens, Ramon, and
  Gartner]{gartner_graph_2006}
K.~Driessens, J.~Ramon, and T.~Gartner, ``Graph {{Kernels}} and {{Gaussian
  Processes}} for {{Relational Reinforcement Learning}},'' \emph{Machine
  Learning}, 2006.

\bibitem[Garg et~al.(2020)Garg, Bajpai, and Mausam]{garg_symbolic_2020}
S.~Garg, A.~Bajpai, and Mausam, ``Symbolic {{Network}}: {{Generalized Neural
  Policies}} for {{Relational MDPs}},'' \emph{arXiv:2002.07375}, 2020.

\bibitem[Janisch et~al.(2021)Janisch, Pevn{\'y}, and
  Lis{\'y}]{janisch_symbolic_2021}
J.~Janisch, T.~Pevn{\'y}, and V.~Lis{\'y}, ``Symbolic {{Relational Deep
  Reinforcement Learning}} based on {{Graph Neural Networks}},''
  \emph{arXiv:2009.12462}, 2021.

\bibitem[Boutilier et~al.(2000)Boutilier, Dearden, and
  Goldszmidt]{boutilier_stochastic_2000}
C.~Boutilier, R.~Dearden, and M.~Goldszmidt, ``Stochastic dynamic programming
  with factored representations,'' \emph{Artificial Intelligence}, vol. 121,
  no. 1-2, pp. 49--107, 2000.

\bibitem[Koller(1999)]{koller_probabilistic_1999}
D.~Koller, ``Probabilistic relational models,'' in \emph{Inductive Logic
  Programming}, 1999, pp. 3--13.

\bibitem[Dean and Kanazawa(1990)]{dean_model_1990}
T.~Dean and K.~Kanazawa, ``A model for reasoning about persistence and
  causation,'' \emph{Computational Intelligence}, vol.~5, no.~3, pp. 142--150,
  1990.

\bibitem[Sanner(2011)]{sanner_relational_2011}
S.~Sanner, ``Relational {{Dynamic Influence Diagram Language}} ({{RDDL}}):
  {{Language Description}},'' in \emph{International Planning Competition},
  2011.

\bibitem[Younes and Littman(2004)]{younes_ppddl10_2004}
Younes and Littman, ``{{PPDDL1}}.0: {{The}} language for the probabilistic part
  of {{IPC}}-4,'' 2004.

\bibitem[Diuk et~al.(2008)Diuk, Cohen, and Littman]{diuk_object-oriented_2008}
C.~Diuk, A.~Cohen, and M.~L. Littman, ``An object-oriented representation for
  efficient reinforcement learning,'' in \emph{ICML}, 2008.

\bibitem[Marom and Rosman(2018)]{marom_zero-shot_2018}
O.~Marom and B.~Rosman, ``Zero-{{Shot Transfer}} with {{Deictic
  Object}}-{{Oriented Representation}} in {{Reinforcement Learning}},'' in
  \emph{{NeurIPS}}, 2018.

\bibitem[Otterlo(2005)]{otterlo_survey_2005}
M.~V. Otterlo, ``A survey of reinforcement learning in relational domains,''
  {CTIT Technical Report Series}, Tech. Rep., 2005.

\bibitem[Vinyals et~al.(2017)Vinyals, Ewalds, Bartunov, Georgiev, Vezhnevets,
  Yeo, Makhzani, K{\"u}ttler, Agapiou, Schrittwieser, Quan, Gaffney, Petersen,
  Simonyan, Schaul, {van Hasselt}, Silver, Lillicrap, Calderone, Keet,
  et~al.]{vinyals_starcraft_2017}
O.~Vinyals, T.~Ewalds, S.~Bartunov, P.~Georgiev, A.~S. Vezhnevets, M.~Yeo,
  A.~Makhzani, H.~K{\"u}ttler, J.~Agapiou, J.~Schrittwieser, J.~Quan,
  S.~Gaffney, S.~Petersen, K.~Simonyan, T.~Schaul, H.~{van Hasselt}, D.~Silver,
  T.~Lillicrap, K.~Calderone, P.~Keet \emph{et~al.}, ``{{StarCraft II}}: {{A
  New Challenge}} for {{Reinforcement Learning}},'' \emph{arXiv:1708.04782},
  2017.

\bibitem[Otterlo(2009)]{otterlo_book_2009}
M.~Otterlo, \emph{The Logic of Adaptive Behavior Knowledge Representation and
  Algorithms for Adaptive Sequential Decision Making under Uncertainty in First
  Order and Relational Domains}, 01 2009.

\bibitem[van Otterlo(2012)]{otterlo_wiering_2012}
M.~van Otterlo, ``Solving relational and first order logical markov decision
  processes: A survey,'' in \emph{Reinforcement Learning}, M.~Wiering and
  M.~van Otterlo, Eds.\hskip 1em plus 0.5em minus 0.4em\relax Springer Berlin
  Heidelberg, 2012, vol.~12, pp. 253--292.

\bibitem[Mart{\'i}nez et~al.(2017{\natexlab{a}})Mart{\'i}nez, Aleny{\`a}, and
  Torras]{martinez_relational_2017}
D.~Mart{\'i}nez, G.~Aleny{\`a}, and C.~Torras, ``Relational reinforcement
  learning with guided demonstrations,'' \emph{Artificial Intelligence}, vol.
  247, pp. 295--312, 2017.

\bibitem[Harnad(1990)]{harnad_symbol_1990}
S.~Harnad, ``The symbol grounding problem,'' \emph{Physica D-nonlinear
  Phenomena}, vol.~42, pp. 335--346, 1990.

\bibitem[Konidaris et~al.(2014)Konidaris, Kaelbling, and
  {Lozano-Perez}]{konidaris_constructing_2014}
G.~Konidaris, L.~P. Kaelbling, and T.~{Lozano-Perez}, ``Constructing {{Symbolic
  Representations}} for {{High}}-{{Level Planning}},'' in \emph{AAAI}, 2014.

\bibitem[Konidaris et~al.(2015)Konidaris, Kaelbling, and
  {Lozano-Perez}]{konidaris_symbol_2015}
------, ``Symbol {{Acquisition}} for {{Probabilistic High}}-{{Level
  Planning}},'' in \emph{IJCAI}, 2015.

\bibitem[Andersen and Konidaris(2017)]{andersen_active_2017}
G.~Andersen and G.~Konidaris, ``Active {{Exploration}} for {{Learning Symbolic
  Representations}},'' in \emph{{NeurIPS}}, 2017.

\bibitem[Konidaris et~al.(2018)Konidaris, Kaelbling, and
  {Lozano-Perez}]{konidaris_skills_2018}
G.~Konidaris, L.~P. Kaelbling, and T.~{Lozano-Perez}, ``From {{Skills}} to
  {{Symbols}}: {{Learning Symbolic Representations}} for {{Abstract
  High}}-{{Level Planning}},'' \emph{JAIR}, vol.~61, pp. 215--289, 2018.

\bibitem[Sutton et~al.(1999)Sutton, Precup, and Singh]{sutton_between_1999}
R.~S. Sutton, D.~Precup, and S.~Singh, ``Between {{MDPs}} and semi-{{MDPs}}:
  {{A}} framework for temporal abstraction in reinforcement learning,''
  \emph{Artificial Intelligence}, vol. 112, no. 1-2, pp. 181--211, 1999.

\bibitem[Kulick et~al.(2013)Kulick, Toussaint, Lang, and
  Lopes]{kulick_active_2013}
J.~Kulick, M.~Toussaint, T.~Lang, and M.~Lopes, ``Active {{Learning}} for
  {{Teaching}} a {{Robot Grounded Relational Symbols}}.'' in \emph{IJCAI},
  2013.

\bibitem[Sun et~al.(2020)Sun, Wu, and Lim]{sun_program_2020}
S.-H. Sun, T.-L. Wu, and J.~J. Lim, ``Program {{Guided Agent}},'' in
  \emph{ICLR}, 2020.

\bibitem[Li et~al.(2017{\natexlab{a}})Li, Sycara, and
  Iyer]{li_object-sensitive_2017}
Y.~Li, K.~Sycara, and R.~Iyer, ``Object-sensitive {{Deep Reinforcement
  Learning}},'' in \emph{Global {{Conference}} on {AI}}, 2017.

\bibitem[Garnelo et~al.(2016)Garnelo, Arulkumaran, and
  Shanahan]{garnelo_towards_2016}
M.~Garnelo, K.~Arulkumaran, and M.~Shanahan, ``Towards {{Deep Symbolic
  Reinforcement Learning}},'' in \emph{NeurIPS Workshop on DRL}, 2016.

\bibitem[Goel et~al.(2018)Goel, Weng, and Poupart]{goel_unsupervised_2018}
V.~Goel, J.~Weng, and P.~Poupart, ``Unsupervised video object segmentation for
  deep reinforcement learning,'' in \emph{NeurIPS}, 2018.

\bibitem[Adjodah et~al.(2018)Adjodah, Klinger, and
  Joseph]{adjodah_symbolic_2018}
D.~Adjodah, T.~Klinger, and J.~Joseph, ``Symbolic {{Relation Networks}} for
  {{Reinforcement Learning}},'' in \emph{NeurIPS Workshop on Representation
  Learning}, 2018.

\bibitem[Zambaldi et~al.(2019)Zambaldi, Raposo, Santoro, Bapst, Li, Babuschkin,
  Tuyls, Reichert, Lillicrap, Lockhart, Shanahan, Langston, Pascanu, Botvinick,
  Vinyals, and Battaglia]{zambaldi_deep_2019}
V.~Zambaldi, D.~Raposo, A.~Santoro, V.~Bapst, Y.~Li, I.~Babuschkin, K.~Tuyls,
  D.~Reichert, T.~Lillicrap, E.~Lockhart, M.~Shanahan, V.~Langston, R.~Pascanu,
  M.~Botvinick, O.~Vinyals, and P.~Battaglia, ``Deep reinforcement learning
  with relational inductive biases,'' in \emph{ICLR}, 2019.

\bibitem[Brunelli(2009)]{brunelli_template_2009}
R.~Brunelli, \emph{Template Matching Techniques in Computer Vision: Theory and
  Practice}.\hskip 1em plus 0.5em minus 0.4em\relax Wiley Publishing, 2009.

\bibitem[Viola and Jones(2001)]{viola_robust_2001}
P.~Viola and M.~Jones, ``Robust real-time object detection,'' in
  \emph{International Journal of Computer Vision}, 2001.

\bibitem[Iyer et~al.(2018)Iyer, Li, Li, Lewis, Sundar, and
  Sycara]{iyer_transparency_2018}
R.~Iyer, Y.~Li, H.~Li, M.~Lewis, R.~Sundar, and K.~Sycara, ``Transparency and
  {{Explanation}} in {{Deep Reinforcement Learning Neural Networks}},'' in
  \emph{AIES}, 2018.

\bibitem[Bader and Hitzler(2005)]{bader_dimensions_2005}
S.~Bader and P.~Hitzler, ``Dimensions of neural-symbolic integration
  \textemdash{} a structured survey,'' in \emph{We Will Show Them: {{Essays}}
  in Honour of Dov Gabbay}, 2005.

\bibitem[Dutra and {d'Avila Garcez}(2017)]{dutra_comparison_2017}
A.~R. Dutra and A.~S. {d'Avila Garcez}, ``A {{Comparison}} between deep
  {{Q}}-networks and deep symbolic reinforcement learning,'' in \emph{{{CEUR
  Workshop Proceedings}}}, 2017.

\bibitem[{d'Avila Garcez} et~al.(2018){d'Avila Garcez}, Dutra, and
  Alonso]{davila_garcez_towards_2018}
A.~{d'Avila Garcez}, A.~R.~R. Dutra, and E.~Alonso, ``Towards {{Symbolic
  Reinforcement Learning}} with {{Common Sense}},'' \emph{arXiv: 1804.08597},
  2018.

\bibitem[Battaglia et~al.(2018)Battaglia, Hamrick, Bapst, Zambaldi, Malinowski,
  Tacchetti, Raposo, Santoro, Faulkner, Gulcehre, Song, Ballard, Gilmer, Dahl,
  Vaswani, Allen, Nash, Langston, Dyer, Heess,
  et~al.]{battaglia_relational_2018}
P.~W. Battaglia, J.~B. Hamrick, V.~Bapst, V.~Zambaldi, M.~Malinowski,
  A.~Tacchetti, D.~Raposo, A.~Santoro, R.~Faulkner, C.~Gulcehre, F.~Song,
  A.~Ballard, J.~Gilmer, G.~Dahl, A.~Vaswani, K.~Allen, C.~Nash, V.~Langston,
  C.~Dyer, N.~Heess \emph{et~al.}, ``Relational inductive biases, deep
  learning, and graph networks,'' \emph{arXiv: 1806.01261}, 2018.

\bibitem[Scarselli et~al.(2009)Scarselli, Gori, {Ah Chung Tsoi}, Hagenbuchner,
  and Monfardini]{scarselli_graph_2009}
F.~Scarselli, M.~Gori, {Ah Chung Tsoi}, M.~Hagenbuchner, and G.~Monfardini,
  ``The {{Graph Neural Network Model}},'' \emph{IEEE Transactions on Neural
  Networks}, vol.~20, no.~1, pp. 61--80, 2009.

\bibitem[Gilmer et~al.(2017)Gilmer, Schoenholz, Riley, Vinyals, and
  Dahl]{gilmer_neural_2017}
J.~Gilmer, S.~S. Schoenholz, P.~F. Riley, O.~Vinyals, and G.~E. Dahl, ``Neural
  message passing for quantum chemistry,'' in \emph{ICML}, 2017.

\bibitem[Cranmer et~al.(2020)Cranmer, Sanchez~Gonzalez, Battaglia, Xu, Cranmer,
  Spergel, and Ho]{cranmer_discovering_2020}
M.~Cranmer, A.~Sanchez~Gonzalez, P.~Battaglia, R.~Xu, K.~Cranmer, D.~Spergel,
  and S.~Ho, ``Discovering symbolic models from deep learning with inductive
  biases,'' in \emph{NeurIPS}, 2020.

\bibitem[{Sanchez-Gonzalez} et~al.(2018){Sanchez-Gonzalez}, Heess,
  Springenberg, Merel, Riedmiller, Hadsell, and
  Battaglia]{sanchez-gonzalez_graph_2018}
A.~{Sanchez-Gonzalez}, N.~Heess, J.~T. Springenberg, J.~Merel, M.~Riedmiller,
  R.~Hadsell, and P.~Battaglia, ``Graph networks as learnable physics engines
  for inference and control,'' in \emph{ICML}, 2018.

\bibitem[Li et~al.(2017{\natexlab{b}})Li, Tarlow, Brockschmidt, and
  Zemel]{li_gated_2017}
Y.~Li, D.~Tarlow, M.~Brockschmidt, and R.~Zemel, ``Gated {{Graph Sequence
  Neural Networks}},'' in \emph{ICLR}, 2017.

\bibitem[Vaswani et~al.(2017)Vaswani, Shazeer, Parmar, Uszkoreit, Jones, Gomez,
  Kaiser, and Polosukhin]{vaswani_attention_2017}
A.~Vaswani, N.~Shazeer, N.~Parmar, J.~Uszkoreit, L.~Jones, A.~N. Gomez,
  {\L}.~Kaiser, and I.~Polosukhin, ``Attention is all you need,'' in
  \emph{NeurIPS}, 2017.

\bibitem[Wang et~al.(2018)Wang, Liao, and Fidler]{wang_nervenet_2018}
T.~Wang, R.~Liao, and S.~Fidler, ``{{NerveNet}}: {{Learning Structured Policy}}
  with {{Graph Neural Networks}},'' in \emph{ICLR}, 2018.

\bibitem[Santoro et~al.(2017)Santoro, Raposo, Barrett, Malinowski, Pascanu,
  Battaglia, and Lillicrap]{santoro_simple_2017}
A.~Santoro, D.~Raposo, D.~G.~T. Barrett, M.~Malinowski, R.~Pascanu,
  P.~Battaglia, and T.~Lillicrap, ``A simple neural network module for
  relational reasoning,'' in \emph{NeurIPS}, 2017.

\bibitem[Chang et~al.(2017)Chang, Ullman, Torralba, and
  Tenenbaum]{chang_compositional_2017}
M.~B. Chang, T.~Ullman, A.~Torralba, and J.~B. Tenenbaum, ``A {{Compositional
  Object}}-{{Based Approach}} to {{Learning Physical Dynamics}},'' in
  \emph{ICLR}, 2017.

\bibitem[Mandal and {Adu-Gyamfi}(2020)]{mandal_object_2020}
V.~Mandal and Y.~{Adu-Gyamfi}, ``Object {{Detection}} and {{Tracking
  Algorithms}} for {{Vehicle Counting}}: {{A Comparative Analysis}},''
  \emph{arXiv:2007.16198}, 2020.

\bibitem[Bear et~al.(2020)Bear, Fan, Mrowca, Li, Alter, Nayebi, Schwartz,
  {Fei-Fei}, Wu, Tenenbaum, and Yamins]{bear_learning_2020}
D.~Bear, C.~Fan, D.~Mrowca, Y.~Li, S.~Alter, A.~Nayebi, J.~Schwartz, L.~F.
  {Fei-Fei}, J.~Wu, J.~Tenenbaum, and D.~L. Yamins, ``Learning physical graph
  representations from visual scenes,'' in \emph{NeurIPS}, 2020.

\bibitem[Donadello et~al.(2017)Donadello, Serafini, and
  D'Avila~Garcez]{donadello_logic_2017}
I.~Donadello, L.~Serafini, and A.~D'Avila~Garcez, ``Logic tensor networks for
  semantic image interpretation,'' in \emph{IJCAI}, 2017.

\bibitem[Serafini and d'Avila Garcez(2016)]{serafini_logic_2016}
L.~Serafini and A.~d'Avila Garcez, ``Logic {{Tensor Networks}}: {{Deep
  Learning}} and {{Logical Reasoning}} from {{Data}} and {{Knowledge}},'' in
  \emph{{{CEUR Workshop}}}, 2016.

\bibitem[Higgins et~al.(2018)Higgins, Amos, Pfau, Racaniere, Matthey, Rezende,
  and Lerchner]{higgins_towards_2018}
I.~Higgins, D.~Amos, D.~Pfau, S.~Racaniere, L.~Matthey, D.~Rezende, and
  A.~Lerchner, ``Towards a {{Definition}} of {{Disentangled
  Representations}},'' \emph{arXiv:1812.02230}, 2018.

\bibitem[Hengst()]{Hengst2010}
B.~Hengst, ``Hierarchical reinforcement learning,'' in \emph{Encyclopedia of
  machine learning}, C.~Sammut and G.~I. Webb, Eds.\hskip 1em plus 0.5em minus
  0.4em\relax Springer {US}, pp. 495--502.

\bibitem[Beyret et~al.(2019)Beyret, Shafti, and Faisal]{beyret_dot--dot_2019}
B.~Beyret, A.~Shafti, and A.~A. Faisal, ``Dot-to-dot: {{Explainable}}
  hierarchical reinforcement learning for robotic manipulation,'' in
  \emph{IROS}, 2019.

\bibitem[Sridharan et~al.(2019)Sridharan, Gelfond, Zhang, and
  Wyatt]{sridharan_reba_2019}
M.~Sridharan, M.~Gelfond, S.~Zhang, and J.~Wyatt, ``{{REBA}}: {{A
  Refinement}}-{{Based Architecture}} for {{Knowledge Representation}} and
  {{Reasoning}} in {{Robotics}},'' \emph{JAIR}, vol.~65, pp. 87--180, 2019.

\bibitem[Andreas et~al.(2017)Andreas, Klein, and Levine]{andreas_modular_2017}
J.~Andreas, D.~Klein, and S.~Levine, ``Modular multitask reinforcement learning
  with policy sketches,'' in \emph{ICML}, 2017.

\bibitem[Shu et~al.(2018)Shu, Xiong, and Socher]{shu_hierarchical_2018}
T.~Shu, C.~Xiong, and R.~Socher, ``Hierarchical and {{Interpretable Skill
  Acquisition}} in {{Multi}}-task {{Reinforcement Learning}},'' in \emph{ICLR},
  2018.

\bibitem[Wu et~al.(2019{\natexlab{a}})Wu, Gupta, and
  Kochenderfer]{wu_model_2019}
B.~Wu, J.~K. Gupta, and M.~J. Kochenderfer, ``Model primitive hierarchical
  lifelong reinforcement learning,'' in \emph{AAMAS}, 2019.

\bibitem[Leonetti et~al.(2016)Leonetti, Iocchi, and
  Stone]{leonetti_synthesis_2016}
M.~Leonetti, L.~Iocchi, and P.~Stone, ``A synthesis of automated planning and
  reinforcement learning for efficient, robust decision-making,''
  \emph{Artificial Intelligence}, vol. 241, pp. 103--130, 2016.

\bibitem[Yang et~al.(2018{\natexlab{a}})Yang, Lyu, Liu, and
  Gustafson]{yang_peorl_2018}
F.~Yang, D.~Lyu, B.~Liu, and S.~Gustafson, ``{{PEORL}}: {{Integrating Symbolic
  Planning}} and {{Hierarchical Reinforcement Learning}} for {{Robust
  Decision}}-{{Making}},'' in \emph{IJCAI}, 2018.

\bibitem[Jiang et~al.(2018)Jiang, Yang, Zhang, and
  Stone]{jiang_integrating_2018}
Y.~Jiang, F.~Yang, S.~Zhang, and P.~Stone, ``Integrating {{Task}}-{{Motion
  Planning}} with {{Reinforcement Learning}} for {{Robust Decision Making}} in
  {{Mobile Robots}},'' in \emph{ICAPS}, 2018.

\bibitem[Lyu et~al.(2019)Lyu, Yang, Liu, and Gustafson]{lyu_sdrl_2019}
D.~Lyu, F.~Yang, B.~Liu, and S.~Gustafson, ``{{SDRL}}: {{Interpretable}} and
  {{Data}}-{{Efficient Deep Reinforcement Learning Leveraging Symbolic
  Planning}},'' in \emph{AAAI}, 2019.

\bibitem[{Furelos-Blanco} et~al.(2021){Furelos-Blanco}, Law, Jonsson, Broda,
  and Russo]{furelos-blanco_induction_2021}
D.~{Furelos-Blanco}, M.~Law, A.~Jonsson, K.~Broda, and A.~Russo, ``Induction
  and {{Exploitation}} of {{Subgoal Automata}} for {{Reinforcement
  Learning}},'' \emph{JAIR}, vol.~70, pp. 1031--1116, 2021.

\bibitem[Li et~al.(2019)Li, Serlin, Yang, and Belta]{li_formal_2019}
X.~Li, Z.~Serlin, G.~Yang, and C.~Belta, ``A formal methods approach to
  interpretable reinforcement learning for robotic planning,'' \emph{Science
  Robotics}, vol.~4, no.~37, 2019.

\bibitem[Zhang et~al.(2019)Zhang, Gao, Zhou, Zhang, Wu, and
  Lin]{zhang_faster_2019}
H.~Zhang, Z.~Gao, Y.~Zhou, H.~Zhang, K.~Wu, and F.~Lin, ``Faster and {{Safer
  Training}} by {{Embedding High}}-{{Level Knowledge}} into {{Deep
  Reinforcement Learning}},'' \emph{arXiv: 1910.09986}, 2019.

\bibitem[Lu et~al.(2018)Lu, Zhang, Stone, and Chen]{lu_robot_2018}
K.~Lu, S.~Zhang, P.~Stone, and X.~Chen, ``Robot {{Representation}} and
  {{Reasoning}} with {{Knowledge}} from {{Reinforcement Learning}},''
  \emph{arXiv: 1809.11074}, 2018.

\bibitem[Camacho et~al.(2019)Camacho, Toro~Icarte, Klassen, Valenzano, and
  McIlraith]{camacho_ltl_2019}
A.~Camacho, R.~Toro~Icarte, T.~Q. Klassen, R.~Valenzano, and S.~A. McIlraith,
  ``{{LTL}} and {{Beyond}}: {{Formal Languages}} for {{Reward Function
  Specification}} in {{Reinforcement Learning}},'' in \emph{IJCAI}, 2019.

\bibitem[Grzes and Kudenko(2008)]{grzes_plan-based_2008}
M.~Grzes and D.~Kudenko, ``Plan-based reward shaping for reinforcement
  learning,'' in \emph{{Int. Conference Intelligent Systems}}, 2008.

\bibitem[Cimatti et~al.(2008)Cimatti, Pistore, and
  Traverso]{cimatti_automated_2008}
A.~Cimatti, M.~Pistore, and P.~Traverso, ``Automated planning,'' in
  \emph{Handbook of Knowledge Representation}, 2008.

\bibitem[Illanes et~al.(2020)Illanes, Yan, Icarte, and
  McIlraith]{illanes_symbolic_2020}
L.~Illanes, X.~Yan, R.~T. Icarte, and S.~A. McIlraith, ``Symbolic {{Plans}} as
  {{High}}-{{Level Instructions}} for {{Reinforcement Learning}},'' in
  \emph{ICAPS}, 2020.

\bibitem[Zhang and Sridharan(2020)]{zhang_survey_2020}
S.~Zhang and M.~Sridharan, ``A {{Survey}} of {{Knowledge}}-based {{Sequential
  Decision Making}} under {{Uncertainty}},'' \emph{arXiv: 2008.08548}, 2020.

\bibitem[Ng and Russell(2000)]{ng_algorithms_2000}
A.~Y. Ng and S.~Russell, ``Algorithms for {{Inverse Reinforcement Learning}},''
  in \emph{ICML}, 2000.

\bibitem[Pasula et~al.(2007)Pasula, Zettlemoyer, and
  Kaelbling]{pasula_learning_2007}
H.~M. Pasula, L.~S. Zettlemoyer, and L.~P. Kaelbling, ``Learning symbolic
  models of stochastic domains,'' \emph{JAIR}, 2007.

\bibitem[Walker et~al.(2008)Walker, Torrey, Shavlik, and
  MacLin]{walker_building_2008}
T.~Walker, L.~Torrey, J.~Shavlik, and R.~MacLin, ``Building relational world
  models for reinforcement learning,'' in \emph{LNCS}, 2008.

\bibitem[Walsh(2010)]{walsh_efficient_2010}
J.~Walsh, ``Efficient {{Learning}} of {{Relational Models}} for {{Sequential
  Decision Making}},'' Ph.D. dissertation, Rutgers, 2010.

\bibitem[Mart{\'i}nez et~al.(2016)Mart{\'i}nez, Aleny{\`a}, Torras, Ribeiro,
  and Inoue]{martinez_learning_2016}
D.~Mart{\'i}nez, G.~Aleny{\`a}, C.~Torras, T.~Ribeiro, and K.~Inoue, ``Learning
  relational dynamics of stochastic domains for planning,'' in \emph{ICAPS},
  2016.

\bibitem[Metzen(2013)]{metzen_learning_2013}
J.~H. Metzen, ``Learning {{Graph}}-{{Based Representations}} for {{Continuous
  Reinforcement Learning Domains}},'' in \emph{ECML}, 2013.

\bibitem[Kansky et~al.(2017)Kansky, Silver, M{\'e}ly, Eldawy,
  {L{\'a}zaro-Gredilla}, Lou, Dorfman, Sidor, Phoenix, and
  George]{kansky_schema_2017}
K.~Kansky, T.~Silver, D.~A. M{\'e}ly, M.~Eldawy, M.~{L{\'a}zaro-Gredilla},
  X.~Lou, N.~Dorfman, S.~Sidor, S.~Phoenix, and D.~George, ``Schema
  {{Networks}}: {{Zero}}-shot {{Transfer}} with a {{Generative Causal Model}}
  of {{Intuitive Physics}},'' in \emph{ICML}, 2017.

\bibitem[Kaiser et~al.(2019)Kaiser, Otte, Runkler, and
  Ek]{kaiser_interpretable_2019}
M.~Kaiser, C.~Otte, T.~Runkler, and C.~H. Ek, ``Interpretable dynamics models
  for data-efficient reinforcement learning,'' in \emph{ESANN}, 2019.

\bibitem[Zhang et~al.(2018{\natexlab{b}})Zhang, Sukhbaatar, Lerer, Szlam, and
  Fergus]{zhang_composable_2018}
A.~Zhang, S.~Sukhbaatar, A.~Lerer, A.~Szlam, and R.~Fergus, ``Composable
  {{Planning}} with {{Attributes}},'' in \emph{ICML}, 2018.

\bibitem[Eysenbach et~al.(2019)Eysenbach, Salakhutdinov, and
  Levine]{eysenbach_search_2019}
B.~Eysenbach, R.~R. Salakhutdinov, and S.~Levine, ``Search on the replay
  buffer: {{Bridging}} planning and reinforcement learning,'' in
  \emph{NeurIPS}, 2019.

\bibitem[Battaglia et~al.(2016)Battaglia, Pascanu, Lai, Rezende, and
  Kavukcuoglu]{battaglia_interaction_2016}
P.~Battaglia, R.~Pascanu, M.~Lai, D.~Rezende, and K.~Kavukcuoglu, ``Interaction
  networks for learning about objects, relations and physics,'' in
  \emph{{NeurIPS}}, 2016.

\bibitem[Finn et~al.(2016)Finn, Goodfellow, and Levine]{finn_unsupervised_2016}
C.~Finn, I.~Goodfellow, and S.~Levine, ``Unsupervised learning for physical
  interaction through video prediction,'' in \emph{NeurIPS}, 2016.

\bibitem[Finn and Levine(2017)]{finn_deep_2017}
C.~Finn and S.~Levine, ``Deep visual foresight for planning robot motion,'' in
  \emph{ICRA}, 2017.

\bibitem[Zhu et~al.(2018)Zhu, Huang, and Zhang]{zhu_object-oriented_2018}
G.~Zhu, Z.~Huang, and C.~Zhang, ``Object-oriented dynamics predictor,'' in
  \emph{NeurIPS}, 2018.

\bibitem[Zhu et~al.(2020)Zhu, Wang, Ren, Lin, and
  Zhang]{zhu_object-oriented_2020}
G.~Zhu, J.~Wang, Z.~Ren, Z.~Lin, and C.~Zhang, ``Object-{{Oriented Dynamics
  Learning}} through {{Multi}}-{{Level Abstraction}},'' in \emph{AAAI}, 2020.

\bibitem[Agnew and Domingos(2018)]{agnew_unsupervised_2018}
W.~Agnew and P.~Domingos, ``Unsupervised {{Object}}-{{Level Deep Reinforcement
  Learning}},'' in \emph{NeurIPS Workshop on Deep RL}, 2018.

\bibitem[Russell(1998)]{russell_learning_1998}
S.~Russell, ``Learning {{Agents}} for {{Uncertain Environments}},'' in
  \emph{COLT}, 1998.

\bibitem[Randlov and Alstrom(1998)]{randlov_learning_1998}
J.~Randlov and P.~Alstrom, ``Learning to drive a bicycle using reinforcement
  learning and shaping,'' in \emph{ICML}, 1998.

\bibitem[Arnold et~al.(2017)Arnold, Kasenberg, and Scheutz]{arnold_value_2017}
T.~Arnold, D.~Kasenberg, and M.~Scheutz, ``Value {{Alignment}} or
  {{Misalignment}} \textendash{} {{What Will Keep Systems Accountable}}?'' in
  \emph{AAAI Workshop}, 2017.

\bibitem[Pomerleau(1989)]{pomerleau_alvinn_1989}
D.~Pomerleau, ``Alvinn: {{An}} autonomous land vehicle in a neural network,''
  in \emph{{NeurIPS}}, 1989.

\bibitem[Osa et~al.(2018)Osa, Pajarinen, Neumann, Bagnell, Abbeel, and
  Peters]{osa_algorithmic_2018}
T.~Osa, J.~Pajarinen, G.~Neumann, J.~A. Bagnell, P.~Abbeel, and J.~Peters,
  ``Algorithmic {{Perspective}} on {{Imitation Learning}},'' \emph{Foundations
  and Trends in Robotics}, vol.~7, no. 1\textendash 2, pp. 1--179, 2018.

\bibitem[Arora and Doshi(2018)]{arora_survey_2018}
S.~Arora and P.~Doshi, ``A survey of inverse reinforcement learning:
  {{Challenges}}, methods and progress,'' \emph{arXiv: 1806.06877}, 2018.

\bibitem[Maclin and Shavlik(1996)]{maclin_creating_1996}
R.~Maclin and J.~W. Shavlik, ``Creating advice-taking reinforcement learners,''
  \emph{Machine Learning}, vol.~22, pp. 251--282, 1996.

\bibitem[Kunapuli et~al.(2013)Kunapuli, Odom, Shavlik, and
  Natarajan]{kunapuli_guiding_2013}
G.~Kunapuli, P.~Odom, J.~W. Shavlik, and S.~Natarajan, ``Guiding {{Autonomous
  Agents}} to {{Better Behaviors}} through {{Human Advice}},'' in \emph{ICDM},
  2013.

\bibitem[Rothkopf and Dimitrakakis(2011)]{rothkopf_preference_2011}
C.~A. Rothkopf and C.~Dimitrakakis, ``Preference {{Elicitation}} and {{Inverse
  Reinforcement Learning}},'' in \emph{ECML}, 2011.

\bibitem[Weng et~al.(2013)Weng, {Busa-Fekete}, and
  H{\"u}llermeier]{weng_interactive_2013}
P.~Weng, R.~{Busa-Fekete}, and E.~H{\"u}llermeier, ``Interactive
  {{Q}}-{{Learning}} with {{Ordinal Rewards}} and {{Unreliable Tutor}},'' in
  \emph{ECML Workshop on {{RL}} with {{Generalized Feedback}}}, 2013.

\bibitem[Munzer et~al.(2015)Munzer, Piot, Geist, Pietquin, and
  Lopes]{munzer_inverse_2015}
T.~Munzer, B.~Piot, M.~Geist, O.~Pietquin, and M.~Lopes, ``Inverse
  reinforcement learning in relational domains,'' in \emph{IJCAI}, 2015.

\bibitem[Mart{\'i}nez et~al.(2017{\natexlab{b}})Mart{\'i}nez, Aleny{\`a},
  Ribeiro, Inoue, and Torras]{martinez_relational_2017-1}
D.~Mart{\'i}nez, G.~Aleny{\`a}, T.~Ribeiro, K.~Inoue, and C.~Torras,
  ``Relational reinforcement learning for planning with exogenous effects,''
  \emph{Journal of Machine Learning Research}, vol.~18, no.~78, pp. 1--44,
  2017.

\bibitem[Srinivasan and {Doshi-Velez}(2020)]{srinivasan_interpretable_2020}
S.~Srinivasan and F.~{Doshi-Velez}, ``Interpretable batch {{IRL}} to extract
  clinician goals in {{ICU}} hypotension management,'' in \emph{AMIA Joint
  Summits on Translational Science}, 2020.

\bibitem[Aksaray et~al.(2016)Aksaray, Jones, Kong, Schwager, and
  Belta]{aksaray_q-learning_2016}
D.~Aksaray, A.~Jones, Z.~Kong, M.~Schwager, and C.~Belta, ``Q-{{Learning}} for
  robust satisfaction of signal temporal logic specifications,'' in \emph{CDC},
  2016.

\bibitem[Littman et~al.(2017)Littman, Topcu, Fu, Isbell, Wen, and
  MacGlashan]{littman_environment-independent_2017}
M.~L. Littman, U.~Topcu, J.~Fu, C.~Isbell, M.~Wen, and J.~MacGlashan,
  ``Environment-{{Independent Task Specifications}} via {{GLTL}},'' 2017.

\bibitem[Li et~al.(2017{\natexlab{c}})Li, Vasile, and
  Belta]{li_reinforcement_2017}
X.~Li, C.~I. Vasile, and C.~Belta, ``Reinforcement learning with temporal logic
  rewards,'' in \emph{IROS}, 2017.

\bibitem[Toro~Icarte et~al.(2018{\natexlab{a}})Toro~Icarte, Klassen, Valenzano,
  and McIlraith]{toro_icarte_teaching_2018}
R.~Toro~Icarte, T.~Q. Klassen, R.~Valenzano, and S.~A. McIlraith, ``Teaching
  {{Multiple Tasks}} to an {{RL Agent}} using {{LTL}},'' in \emph{AAMAS}, 2018.

\bibitem[Toro~Icarte et~al.(2018{\natexlab{b}})Toro~Icarte, Klassen, Valenzano,
  and McIlraith]{toro_icarte_using_2018}
R.~Toro~Icarte, T.~Klassen, R.~Valenzano, and S.~McIlraith, ``Using {{Reward
  Machines}} for {{High}}-{{Level Task Specification}} and {{Decomposition}} in
  {{Reinforcement Learning}},'' in \emph{ICML}, 2018.

\bibitem[Hasanbeig et~al.(2020)Hasanbeig, Kroening, and
  Abate]{hasanbeig_deep_2020}
M.~Hasanbeig, D.~Kroening, and A.~Abate, ``Deep {{Reinforcement Learning}} with
  {{Temporal Logics}},'' in \emph{Formal {{Modeling}} and {{Analysis}} of
  {{Timed Systems}}}, 2020.

\bibitem[Toro~Icarte et~al.(2019)Toro~Icarte, Waldie, Klassen, Valenzano,
  Castro, and McIlraith]{toro_icarte_learning_2019}
R.~Toro~Icarte, E.~Waldie, T.~Klassen, R.~Valenzano, M.~Castro, and
  S.~McIlraith, ``Learning {{Reward Machines}} for {{Partially Observable
  Reinforcement Learning}},'' in \emph{{NeurIPS}}, 2019.

\bibitem[Xu et~al.(2020)Xu, Gavran, Ahmad, Majumdar, Neider, Topcu, and
  Wu]{xu_joint_2020}
Z.~Xu, I.~Gavran, Y.~Ahmad, R.~Majumdar, D.~Neider, U.~Topcu, and B.~Wu,
  ``Joint {{Inference}} of {{Reward Machines}} and {{Policies}} for
  {{Reinforcement Learning}},'' in \emph{ICAPS}, 2020.

\bibitem[Gaon and Brafman(2020)]{gaon_reinforcement_2020}
M.~Gaon and R.~I. Brafman, ``Reinforcement {{Learning}} with
  {{Non}}-{{Markovian Rewards}},'' in \emph{{{AAAI}}}, 2020.

\bibitem[Tasse et~al.(2020)Tasse, James, and Rosman]{tasse_boolean_2020}
G.~N. Tasse, S.~James, and B.~Rosman, ``A {{Boolean Task Algebra}} for
  {{Reinforcement Learning}},'' in \emph{{{NeurIPS}}}, 2020.

\bibitem[Yang et~al.(2018{\natexlab{b}})Yang, Morillo, and
  Hospedales]{yang_deep_2018}
Y.~Yang, I.~G. Morillo, and T.~M. Hospedales, ``Deep {{Neural Decision
  Trees}},'' in \emph{ICML Workshop on {{Human Interpretability}} in {ML}},
  2018.

\bibitem[Kasenberg and Scheutz(2017)]{kasenberg_interpretable_2017}
D.~Kasenberg and M.~Scheutz, ``Interpretable {{Apprenticeship Learning}} with
  {{Temporal Logic Specifications}},'' in \emph{CDC}, 2017.

\bibitem[Todorov(2009)]{todorov_compositionality_2009}
E.~Todorov, ``Compositionality of optimal control laws,'' in \emph{NeurIPS},
  2009.

\bibitem[Ernst et~al.(2005)Ernst, Geurts, and Wehenkel]{ernst_tree-based_2005}
D.~Ernst, P.~Geurts, and L.~Wehenkel, ``Tree-based batch mode reinforcement
  learning,'' \emph{JMLR}, vol.~6, no. Apr, pp. 503--556, 2005.

\bibitem[Likmeta et~al.(2020)Likmeta, Metelli, Tirinzoni, Giol, Restelli, and
  Romano]{likmeta_combining_2020}
A.~Likmeta, A.~M. Metelli, A.~Tirinzoni, R.~Giol, M.~Restelli, and D.~Romano,
  ``Combining reinforcement learning with rule-based controllers for
  transparent and general decision-making in autonomous driving,''
  \emph{Robotics and Autonomous Systems}, 2020.

\bibitem[Silva et~al.(2020)Silva, Gombolay, Killian, Jimenez, and
  Son]{silva_optimization_2020}
A.~Silva, M.~Gombolay, T.~Killian, I.~Jimenez, and S.-H. Son, ``Optimization
  {{Methods}} for {{Interpretable Differentiable Decision Trees Applied}} to
  {{Reinforcement Learning}},'' in \emph{AISTATS}, 2020.

\bibitem[Silva and Gombolay(2020)]{silva_neural-encoding_2020}
A.~Silva and M.~Gombolay, ``Neural-encoding {{Human Experts}}' {{Domain
  Knowledge}} to {{Warm Start Reinforcement Learning}},'' \emph{arXiv:
  1902.06007}, 2020.

\bibitem[Topin et~al.(2021)Topin, Milani, Fang, and
  Veloso]{topin_iterative_2021}
N.~Topin, S.~Milani, F.~Fang, and M.~Veloso, ``Iterative {{Bounding MDPs}}:
  {{Learning Interpretable Policies}} via {{Non}}-{{Interpretable Methods}},''
  in \emph{AAAI}, 2021.

\bibitem[Gupta et~al.(2015)Gupta, Talvitie, and Bowling]{gupta_policy_2015}
U.~D. Gupta, E.~Talvitie, and M.~Bowling, ``Policy tree: Adaptive
  representation for policy gradient,'' in \emph{AAAI}, 2015.

\bibitem[Maes et~al.(2012{\natexlab{a}})Maes, Fonteneau, Wehenkel, and
  Ernst]{maes_policy_2012}
F.~Maes, R.~Fonteneau, L.~Wehenkel, and D.~Ernst, ``Policy {{Search}} in a
  {{Space}} of {{Simple Closed}}-form {{Formulas}}: {{Towards
  Interpretability}} of {{Reinforcement Learning}},'' in \emph{Discovery
  {{Science}}}, 2012.

\bibitem[Maes et~al.(2012{\natexlab{b}})Maes, Wehenkel, and
  Ernst]{maes_automatic_2012}
F.~Maes, L.~Wehenkel, and D.~Ernst, ``Automatic {{Discovery}} of {{Ranking
  Formulas}} for {{Playing}} with {{Multi}}-armed {{Bandits}},'' in
  \emph{Recent {{Advances}} in {{Reinforcement Learning}}}, 2012.

\bibitem[Hein et~al.(2018)Hein, Udluft, and Runkler]{hein_interpretable_2018}
D.~Hein, S.~Udluft, and T.~A. Runkler, ``Interpretable policies for
  reinforcement learning by genetic programming,'' \emph{Engineering
  Applications of AI}, vol.~76, pp. 158--169, 2018.

\bibitem[Hein et~al.(2019)Hein, Udluft, and Runkler]{hein_generating_2019}
------, ``Generating interpretable reinforcement learning policies using
  genetic programming,'' in \emph{GECCO}, 2019.

\bibitem[Ault et~al.(2020)Ault, Hanna, and Sharon]{ault_learning_2020}
J.~Ault, J.~P. Hanna, and G.~Sharon, ``Learning an {{Interpretable Traffic
  Signal Control Policy}},'' in \emph{AAMAS}, 2020.

\bibitem[Akrour et~al.(2019)Akrour, Tateo, and Peters]{akrour_towards_2019}
R.~Akrour, D.~Tateo, and J.~Peters, ``Towards reinforcement learning of human
  readable policies,'' in \emph{Workshop on Deep Continuous-Discrete Machine
  Learning}, 2019.

\bibitem[Hein et~al.(2017)Hein, Hentschel, Runkler, and
  Udluft]{hein_particle_2017}
D.~Hein, A.~Hentschel, T.~Runkler, and S.~Udluft, ``Particle swarm optimization
  for generating interpretable fuzzy reinforcement learning policies,''
  \emph{Engineering Applications of AI}, vol.~65, pp. 87--98, 2017.

\bibitem[Evans and Grefenstette(2018)]{evans_learning_2018}
R.~Evans and E.~Grefenstette, ``Learning explanatory rules from noisy data,''
  \emph{Journal of Artificial Intelligence Research}, vol.~61, 2018.

\bibitem[Jiang and Luo(2019)]{jiang_neural_2019}
Z.~Jiang and S.~Luo, ``Neural {{Logic Reinforcement Learning}},'' in
  \emph{ICML}, 2019.

\bibitem[Payani and Fekri(2019{\natexlab{a}})]{payani_inductive_2019}
A.~Payani and F.~Fekri, ``Inductive {{Logic Programming}} via {{Differentiable
  Deep Neural Logic Networks}},'' \emph{arXiv: 1906.03523}, 2019.

\bibitem[Payani and Fekri(2019{\natexlab{b}})]{payani_learning_2019}
------, ``Learning {{Algorithms}} via {{Neural Logic Networks}},'' \emph{arXiv:
  1904.01554}, 2019.

\bibitem[Payani and Fekri(2020)]{payani_incorporating_2020}
------, ``Incorporating {{Relational Background Knowledge}} into
  {{Reinforcement Learning}} via {{Differentiable Inductive Logic
  Programming}},'' \emph{arXiv: 2003.10386}, 2020.

\bibitem[Zimmer et~al.(2021)Zimmer, Feng, Glanois, Jiang, Zhang, Weng, Jianye,
  Dong, and Wulong]{zimmer_differentiable_2021}
M.~Zimmer, X.~Feng, C.~Glanois, Z.~Jiang, J.~Zhang, P.~Weng, H.~Jianye,
  L.~Dong, and L.~Wulong, ``Differentiable logic machines,'' \emph{arXiv:
  2102.11529}, 2021.

\bibitem[Yang and Song(2019)]{yang_learn_2019}
Y.~Yang and L.~Song, ``Learn to {{Explain Efficiently}} via {{Neural Logic
  Inductive Learning}},'' in \emph{ICLR}, 2019.

\bibitem[Ma et~al.(2020)Ma, Zhuang, Weng, Li, Shao, Liu, Zhuo, and
  Hao]{ma_interpretable_2020}
Z.~Ma, Y.~Zhuang, P.~Weng, D.~Li, K.~Shao, W.~Liu, H.~H. Zhuo, and J.~Hao,
  ``Interpretable {{Reinforcement Learning With Neural Symbolic Logic}},''
  \emph{{{arxiv}} 2103.08228}, 2020.

\bibitem[Verma et~al.(2019)Verma, M.~Le, Yue, and
  Chaudhuri]{verma_imitation-projected_2019}
A.~Verma, H.~M.~Le, Y.~Yue, and S.~Chaudhuri, ``Imitation-{{Projected
  Programmatic Reinforcement Learning}},'' in \emph{NeurIPS}, 2019.

\bibitem[Anderson et~al.(2020)Anderson, Verma, Dillig, and
  Chaudhuri]{anderson_neurosymbolic_2020}
G.~Anderson, A.~Verma, I.~Dillig, and S.~Chaudhuri, ``Neurosymbolic
  {{Reinforcement Learning}} with {{Formally Verified Exploration}},'' in
  \emph{{NeurIPS}}, 2020.

\bibitem[Chen et~al.(2020)Chen, Li, and Tomizuka]{chen_interpretable_2020}
J.~Chen, S.~E. Li, and M.~Tomizuka, ``Interpretable {{End}}-to-end {{Urban
  Autonomous Driving}} with {{Latent Deep Reinforcement Learning}},'' in
  \emph{ICML Workshop on AI for Autonomous Driving}, 2020.

\bibitem[Levine(2018)]{levine_reinforcement_2018}
S.~Levine, ``Reinforcement {{Learning}} and {{Control}} as {{Probabilistic
  Inference}}: {{Tutorial}} and {{Review}},'' \emph{arXiv: 1805.00909}, 2018.

\bibitem[Dong et~al.(2019)Dong, Mao, Lin, Wang, Li, and Zhou]{dong_neural_2019}
H.~Dong, J.~Mao, T.~Lin, C.~Wang, L.~Li, and D.~Zhou, ``Neural {{Logic
  Machines}},'' in \emph{ICLR}, 2019.

\bibitem[Lao and Cohen(2010)]{lao_relational_2010}
N.~Lao and W.~W. Cohen, ``Relational retrieval using a combination of
  path-constrained random walks,'' in \emph{Machine Learning}, 2010.

\bibitem[Yang et~al.(2017)Yang, Yang, and Cohen]{yang_differentiable_2017}
F.~Yang, Z.~Yang, and W.~W. Cohen, ``Differentiable {{Learning}} of {{Logical
  Rules}} for {{Knowledge Base Reasoning}},'' in \emph{{NeurIPS}}, 2017.

\bibitem[Hussein et~al.(2017)Hussein, Gaber, Elyan, and
  Jayne]{hussein_imitation_2017}
A.~Hussein, M.~M. Gaber, E.~Elyan, and C.~Jayne, ``Imitation {{Learning}}: {{A
  Survey}} of {{Learning Methods}},'' \emph{ACM Computing Surveys}, vol.~50,
  no.~2, pp. 21:1--21:35, 2017.

\bibitem[Rusu et~al.(2016)Rusu, Colmenarejo, G{\"u}l{\c c}ehre, Desjardins,
  Kirkpatrick, Pascanu, Mnih, Kavukcuoglu, and Hadsell]{rusu_policy_2016}
A.~A. Rusu, S.~G. Colmenarejo, {\c C}.~G{\"u}l{\c c}ehre, G.~Desjardins,
  J.~Kirkpatrick, R.~Pascanu, V.~Mnih, K.~Kavukcuoglu, and R.~Hadsell, ``Policy
  distillation,'' in \emph{ICLR}, 2016.

\bibitem[Liu et~al.(2018)Liu, Schulte, Zhu, and Li]{liu_toward_2018}
G.~Liu, O.~Schulte, W.~Zhu, and Q.~Li, ``Toward {{Interpretable Deep
  Reinforcement Learning}} with {{Linear Model U}}-{{Trees}},'' in \emph{ECML},
  2018.

\bibitem[Maes et~al.(1996)Maes, Mataric, Meyer, Pollack, and
  Wilson]{maes_learning_1996}
P.~Maes, M.~J. Mataric, J.~A. Meyer, J.~Pollack, and S.~W. Wilson, ``Learning
  to use selective attention and short-term memory in sequential tasks,'' in
  \emph{Int. Conf. on Simulation of Adaptive Behavior}, 1996.

\bibitem[Ribeiro et~al.(2016{\natexlab{b}})Ribeiro, Singh, and
  Guestrin]{ribeiro_why_2016}
M.~T. Ribeiro, S.~Singh, and C.~Guestrin, ``"{{Why Should I Trust You}}?":
  {{Explaining}} the {{Predictions}} of {{Any Classifier}},'' in \emph{KDD},
  2016.

\bibitem[Bastani et~al.(2018)Bastani, Pu, and
  {Solar-Lezama}]{bastani_verifiable_2018}
O.~Bastani, Y.~Pu, and A.~{Solar-Lezama}, ``Verifiable {{Reinforcement
  Learning}} via {{Policy Extraction}},'' in \emph{{NeurIPS}}, 2018.

\bibitem[Ross et~al.(2011)Ross, Gordon, and Bagnell]{ross_reduction_2011}
S.~Ross, G.~J. Gordon, and J.~A. Bagnell, ``A {{Reduction}} of {{Imitation
  Learning}} and {{Structured Prediction}} to {{No}}-regret {{Online
  Learning}},'' in \emph{{{AISTATS}}}, 2011.

\bibitem[Roth et~al.(2019)Roth, Topin, Jamshidi, and
  Veloso]{roth_conservative_2019}
A.~M. Roth, N.~Topin, P.~Jamshidi, and M.~Veloso, ``Conservative
  {{Q}}-{{Improvement}}: {{Reinforcement Learning}} for an {{Interpretable
  Decision}}-{{Tree Policy}},'' \emph{arXiv: 1907.01180}, 2019.

\bibitem[Vasic et~al.(2019)Vasic, Petrovic, Wang, Nikolic, Singh, and
  Khurshid]{vasic_moet_2019}
M.~Vasic, A.~Petrovic, K.~Wang, M.~Nikolic, R.~Singh, and S.~Khurshid,
  ``{{MoET}}: {{Interpretable}} and {{Verifiable Reinforcement Learning}} via
  {{Mixture}} of {{Expert Trees}},'' \emph{arXiv: 1906.06717}, 2019.

\bibitem[Natarajan et~al.(2011)Natarajan, Joshi, Tadepalli, Kersting, and
  Shavlik]{natarajan_imitation_2011}
S.~Natarajan, S.~Joshi, P.~Tadepalli, K.~Kersting, and J.~Shavlik, ``Imitation
  learning in relational domains: {{A}} functional-gradient boosting
  approach,'' in \emph{IJCAI}, 2011.

\bibitem[Cichosz and Pawe{\l}czak(2014)]{cichosz_imitation_2014}
P.~Cichosz and L.~Pawe{\l}czak, ``Imitation learning of car driving skills with
  decision trees and random forests,'' \emph{International Journal of Applied
  Mathematics and Computer Science}, 2014.

\bibitem[Nageshrao et~al.(2019)Nageshrao, Costa, and
  Filev]{nageshrao_interpretable_2019}
S.~Nageshrao, B.~Costa, and D.~Filev, ``Interpretable approximation of a deep
  reinforcement learning agent as a set of if-then rules,'' in \emph{ICMLA},
  2019.

\bibitem[Verma et~al.(2018)Verma, Murali, Singh, Kohli, and
  Chaudhuri]{verma_programmatically_2018}
A.~Verma, V.~Murali, R.~Singh, P.~Kohli, and S.~Chaudhuri, ``Programmatically
  interpretable reinforcement learning,'' in \emph{ICML}, 2018.

\bibitem[Zhu et~al.(2019)Zhu, Magill, Xiong, and
  Jagannathan]{zhu_inductive_2019}
H.~Zhu, S.~Magill, Z.~Xiong, and S.~Jagannathan, ``An inductive synthesis
  framework for verifiable reinforcement learning,'' in \emph{{ACM SIGPLAN
  Conference} on {PLDI}}, 2019.

\bibitem[Alshiekh et~al.(2018)Alshiekh, Bloem, Ehlers, K{\"o}nighofer, Niekum,
  and Topcu]{alshiekh_safe_2018}
M.~Alshiekh, R.~Bloem, R.~Ehlers, B.~K{\"o}nighofer, S.~Niekum, and U.~Topcu,
  ``Safe reinforcement learning via shielding,'' in \emph{{AAAI}}, 2018.

\bibitem[Burke et~al.(2019)Burke, Penkov, and
  Ramamoorthy]{burke_explanation_2019}
M.~Burke, S.~Penkov, and S.~Ramamoorthy, ``From explanation to synthesis:
  {{Compositional}} program induction for learning from demonstration,'' in
  \emph{RSS}, 2019.

\bibitem[Koul et~al.(2019)Koul, Greydanus, and Fern]{Koul_learning_2019}
A.~Koul, S.~Greydanus, and A.~Fern, ``Learning {{Finite State Representations}}
  of {{Recurrent Policy Networks}},'' in \emph{ICLR}, 2019.

\bibitem[Torrey and Taylor(2013)]{torrey_teaching_2013}
L.~Torrey and M.~E. Taylor, ``Teaching on a {{Budget}}: {{Agents Advising
  Agents}} in {{Reinforcement Learning}},'' in \emph{AAMAS}, 2013.

\bibitem[Zimmer et~al.(2014)Zimmer, Viappiani, and
  Weng]{zimmer_teacher-student_2014}
M.~Zimmer, P.~Viappiani, and P.~Weng, ``Teacher-{{Student Framework}}: A
  {{Reinforcement Learning Approach}},'' in \emph{AAMAS Workshop on
  {{Autonomous Robots}} and {{Multirobot Systems}}}, 2014.

\bibitem[Tang et~al.(2020)Tang, Nguyen, and Ha]{tang_neuroevolution_2020}
Y.~Tang, D.~Nguyen, and D.~Ha, ``Neuroevolution of {{Self}}-{{Interpretable
  Agents}},'' in \emph{GECCO}, 2020.

\bibitem[Mott et~al.(2019)Mott, Zoran, Chrzanowski, Wierstra, and
  Rezende]{mott_towards_2019}
A.~Mott, D.~Zoran, M.~Chrzanowski, D.~Wierstra, and D.~J. Rezende, ``Towards
  {{Interpretable Reinforcement Learning Using Attention Augmented Agents}},''
  in \emph{NeurIPS}, 2019.

\bibitem[Annasamy and Sycara(2019)]{annasamy_towards_2019}
R.~M. Annasamy and K.~Sycara, ``Towards {{Better Interpretability}} in {{Deep
  Q}}-{{Networks}},'' in \emph{AAAI}, 2019.

\bibitem[Wiegreffe and Pinter(2019)]{wiegreffe_attention_2019}
S.~Wiegreffe and Y.~Pinter, ``Attention is not not {{Explanation}},'' in
  \emph{EMNLP}, 2019.

\bibitem[Jain and Wallace(2019)]{jain_attention_2019}
S.~Jain and B.~C. Wallace, ``Attention is not {{Explanation}},'' in
  \emph{{{NAACL}}}, 2019.

\bibitem[Brunner et~al.(2020)Brunner, Liu, Pascual, Richter, Ciaramita, and
  Wattenhofer]{brunner_identifiability_2020}
G.~Brunner, Y.~Liu, D.~Pascual, O.~Richter, M.~Ciaramita, and R.~Wattenhofer,
  ``On {{Identifiability}} in {{Transformers}},'' in \emph{ICLR}, 2020.

\bibitem[Franca et~al.(2014)Franca, Zaverucha, and Garcez]{franca_fast_2014}
M.~V.~M. Franca, G.~Zaverucha, and A.~Garcez, ``Fast relational learning using
  bottom clause propositionalization with artificial neural networks,''
  \emph{Machine Learning}, vol.~94, no.~1, pp. 81--104, Jan. 2014.

\bibitem[Jia et~al.(2019)Jia, Jin, Sun, Hong, and Spanos]{jia_advanced_2019}
R.~Jia, M.~Jin, K.~Sun, T.~Hong, and C.~Spanos, ``Advanced building control via
  deep reinforcement learning,'' in \emph{Energy Procedia}, 2019.

\bibitem[Bucilu{\v a} et~al.(2006)Bucilu{\v a}, Caruana, and
  {Niculescu-Mizil}]{bucilua_model_2006}
C.~Bucilu{\v a}, R.~Caruana, and A.~{Niculescu-Mizil}, ``Model compression,''
  in \emph{{KDD}}, 2006.

\bibitem[Zhang et~al.(2016)Zhang, Lee, and Jordan]{zhang_1-regularized_2016}
Y.~Zhang, J.~D. Lee, and M.~I. Jordan, ``L1-regularized {{Neural Networks}} are
  {{Improperly Learnable}} in {{Polynomial Time}},'' in \emph{ICML}, 2016.

\bibitem[Dragan et~al.(2013)Dragan, Lee, and Srinivasa]{dragan_legibility_2013}
A.~D. Dragan, K.~C. Lee, and S.~S. Srinivasa, ``Legibility and predictability
  of robot motion,'' in \emph{{HRI}}, 2013.

\bibitem[Wu et~al.(2019{\natexlab{b}})Wu, Parbhoo, Hughes, Roth, and
  {Doshi-Velez}]{wu_optimizing_2019}
M.~Wu, S.~Parbhoo, M.~C. Hughes, V.~Roth, and F.~{Doshi-Velez}, ``Optimizing
  for {{Interpretability}} in {{Deep Neural Networks}} with {{Tree
  Regularization}},'' \emph{arXiv: 1908.05254}, 2019.

\bibitem[Goodfellow et~al.(2016)Goodfellow, Bengio, and
  Courville]{goodfellow_deep_2016}
I.~Goodfellow, Y.~Bengio, and A.~Courville, \emph{Deep Learning}, 2016.

\bibitem[Diligenti et~al.(2017)Diligenti, Gori, and
  Sacc{\`a}]{diligenti_semantic-based_2017}
M.~Diligenti, M.~Gori, and C.~Sacc{\`a}, ``Semantic-based regularization for
  learning and inference,'' \emph{Artificial Intelligence}, vol. 244, 2017.

\bibitem[Wang and Pan(2019)]{wang_integrating_2019}
W.~Wang and S.~J. Pan, ``Integrating {{Deep Learning}} with {{Logic Fusion}}
  for {{Information Extraction}},'' in \emph{AAAI}, 2019.

\bibitem[Rockt{\"a}schel et~al.(2015)Rockt{\"a}schel, Singh, and
  Riedel]{rocktaschel_injecting_2015}
T.~Rockt{\"a}schel, S.~Singh, and S.~Riedel, ``Injecting {{Logical Background
  Knowledge}} into {{Embeddings}} for {{Relation Extraction}},'' in
  \emph{{{Human Language Technologies}}}, 2015.

\bibitem[Demeester et~al.(2016)Demeester, Rockt{\"a}schel, and
  Riedel]{demeester_lifted_2016}
T.~Demeester, T.~Rockt{\"a}schel, and S.~Riedel, ``Lifted {{Rule Injection}}
  for {{Relation Embeddings}},'' in \emph{EMNLP}, 2016.

\bibitem[Xu et~al.(2018)Xu, Zhang, Friedman, Liang, and {Van den
  Broeck}]{xu_semantic_2018}
J.~Xu, Z.~Zhang, T.~Friedman, Y.~Liang, and G.~{Van den Broeck}, ``A semantic
  loss function for deep learning with symbolic knowledge,'' in \emph{ICML},
  2018.

\bibitem[Minervini et~al.(2017)Minervini, Demeester, Rockt{\"a}schel, and
  Riedel]{minervini_adversarial_2017}
P.~Minervini, T.~Demeester, T.~Rockt{\"a}schel, and S.~Riedel, ``Adversarial
  {{Sets}} for {{Regularising Neural Link Predictors}},'' in \emph{UAI}, 2017.

\bibitem[Plumb et~al.(2020)Plumb, {Al-Shedivat}, Cabrera, Perer, Xing, and
  Talwalkar]{plumb_regularizing_2020}
G.~Plumb, M.~{Al-Shedivat}, A.~A. Cabrera, A.~Perer, E.~Xing, and A.~Talwalkar,
  ``Regularizing {{Black}}-box {{Models}} for {{Improved Interpretability}},''
  \emph{arXiv:1902.06787}, 2020.

\bibitem[Greydanus et~al.(2018)Greydanus, Koul, Dodge, and
  Fern]{greydanus_visualizing_2018}
S.~Greydanus, A.~Koul, J.~Dodge, and A.~Fern, ``Visualizing and understanding
  atari agents,'' in \emph{ICML}, 2018.

\bibitem[Gupta et~al.(2020)Gupta, Puri, Verma, Kayastha, Deshmukh,
  Krishnamurthy, and Singh]{gupta_explain_2020}
P.~Gupta, N.~Puri, S.~Verma, D.~Kayastha, S.~Deshmukh, B.~Krishnamurthy, and
  S.~Singh, ``Explain {{Your Move}}: {{Understanding Agent Actions Using
  Focused Feature Saliency}},'' in \emph{ICLR}, 2020.

\bibitem[Wang et~al.(2020)Wang, Mase, and Egi]{wang_attribution-based_2020}
Y.~Wang, M.~Mase, and M.~Egi, ``Attribution-based {{Salience Method}} towards
  {{Interpretable Reinforcement Learning}},'' in \emph{Spring {{Symposium}} on
  {{Combining ML}} and {{Knowledge Engineering}} in {{Practice}}}, 2020.

\bibitem[Shi et~al.(2020)Shi, Huang, Song, Wang, Lin, and
  Wu]{shi_self-supervised_2020}
W.~Shi, G.~Huang, S.~Song, Z.~Wang, T.~Lin, and C.~Wu, ``Self-{{Supervised
  Discovering}} of {{Interpretable Features}} for {{Reinforcement Learning}},''
  \emph{arXiv: 2003.07069}, 2020.

\bibitem[Kim and Bansal(2020)]{kim_attentional_2020}
J.~Kim and M.~Bansal, ``Attentional {{Bottleneck}}: {{Towards}} an
  {{Interpretable Deep Driving Network}},'' in \emph{CVPR Workshop}, 2020.

\bibitem[Sequeira and Gervasio(2020)]{sequeira_interestingness_2020}
P.~Sequeira and M.~Gervasio, ``Interestingness {{Elements}} for {{Explainable
  Reinforcement Learning}}: {{Understanding Agents}}' {{Capabilities}} and
  {{Limitations}},'' \emph{Artificial Intelligence}, vol. 288, 2020.

\bibitem[Hayes and Shah(2017)]{hayes_improving_2017}
B.~Hayes and J.~A. Shah, ``Improving {{Robot Controller Transparency Through
  Autonomous Policy Explanation}},'' in \emph{Int. Conf. on HRI}, 2017.

\bibitem[{van der Waa} et~al.(2018){van der Waa}, {van Diggelen}, van~den
  Bosch, and Neerincx]{van_der_waa_contrastive_2018}
J.~{van der Waa}, J.~{van Diggelen}, K.~van~den Bosch, and M.~Neerincx,
  ``Contrastive {{Explanations}} for {{Reinforcement Learning}} in terms of
  {{Expected Consequences}},'' in \emph{{IJCAI} Workshop on {XAI}}, 2018.

\bibitem[Fukuchi et~al.(2017)Fukuchi, Osawa, Yamakawa, and
  Imai]{fukuchi_autonomous_2017}
Y.~Fukuchi, M.~Osawa, H.~Yamakawa, and M.~Imai, ``Autonomous
  {{Self}}-{{Explanation}} of {{Behavior}} for {{Interactive Reinforcement
  Learning Agents}},'' in \emph{{{Int. Conf.}} on {{Human Agent Interaction}}},
  2017.

\bibitem[Madumal et~al.(2020{\natexlab{a}})Madumal, Miller, Sonenberg, and
  Vetere]{madumal_explainable_2020}
P.~Madumal, T.~Miller, L.~Sonenberg, and F.~Vetere, ``Explainable
  {{Reinforcement Learning Through}} a {{Causal Lens}},'' in \emph{AAAI}, 2020.

\bibitem[Madumal et~al.(2020{\natexlab{b}})Madumal, Miller, Sonenberg, and
  Vetere]{madumal_distal_2020}
------, ``Distal {{Explanations}} for {{Model}}-free {{Explainable
  Reinforcement Learning}},'' \emph{arXiv: 2001.10284}, 2020.

\bibitem[Coppens et~al.(2019)Coppens, Efthymiadis, Lenaerts, Now{\'e}, Miller,
  Weber, and Magazzeni]{coppens_distilling_2019}
Y.~Coppens, K.~Efthymiadis, T.~Lenaerts, A.~Now{\'e}, T.~Miller, R.~Weber, and
  D.~Magazzeni, ``Distilling deep reinforcement learning policies in soft
  decision trees,'' in \emph{{IJCAI} Workshop on {XAI}}, 2019.

\bibitem[Bewley and Lawry(2021)]{bewley_tripletree_2021}
T.~Bewley and J.~Lawry, ``{{TripleTree}}: {{A Versatile Interpretable
  Representation}} of {{Black Box Agents}} and their {{Environments}},'' in
  \emph{AAAI}, 2021.

\bibitem[Juozapaitis et~al.(2019)Juozapaitis, Koul, Fern, Erwig, and
  {Doshi-Velez}]{juozapaitis_explainable_2019}
Z.~Juozapaitis, A.~Koul, A.~Fern, M.~Erwig, and F.~{Doshi-Velez}, ``Explainable
  reinforcement learning via reward decomposition,'' in \emph{{{IJCAI}}/{{ECAI
  Workshop}} on {{Explainable Artificial Intelligence}}}, 2019.

\bibitem[Topin and Veloso(2019)]{topin_generation_2019}
N.~Topin and M.~Veloso, ``Generation of {{Policy}}-{{Level Explanations}} for
  {{Reinforcement Learning}},'' in \emph{AAAI}, 2019.

\bibitem[Cruz et~al.(2019)Cruz, Dazeley, and Vamplew]{cruz_memory-based_2019}
F.~Cruz, R.~Dazeley, and P.~Vamplew, ``Memory-{{Based Explainable Reinforcement
  Learning}},'' in \emph{{{Advances}} in {{Artificial Intelligence}}}, 2019.

\bibitem[van~der Maaten and Hinton(2008)]{maaten_visualizing_2008}
L.~van~der Maaten and G.~Hinton, ``Visualizing {{Data}} using t-{{SNE}},''
  \emph{JMLR}, vol.~9, no.~86, pp. 2579--2605, 2008.

\bibitem[Lundberg and Lee(2017)]{lundberg_unified_2017}
S.~M. Lundberg and S.-I. Lee, ``A unified approach to interpreting model
  predictions,'' in \emph{NeurIPS}, 2017.

\bibitem[Lim et~al.(2019)Lim, Yang, Abdul, and Wang]{lim_why_2019}
B.~Y. Lim, Q.~Yang, A.~Abdul, and D.~Wang, ``Why these {{Explanations}}?
  {{Selecting Intelligibility Types}} for {{Explanation Goals}},'' in \emph{IUI
  Workshops}, 2019.

\bibitem[Mittelstadt et~al.(2019)Mittelstadt, Russell, and
  Wachter]{mittelstadt_explaining_2019}
B.~Mittelstadt, C.~Russell, and S.~Wachter, ``Explaining {{Explanations}} in
  {{AI}},'' in \emph{Conf. on Fairness, Accountability, and Transparency},
  2019.

\bibitem[Atrey et~al.(2020)Atrey, Clary, and Jensen]{atrey_exploratory_2020}
A.~Atrey, K.~Clary, and D.~Jensen, ``Exploratory {{Not Explanatory}}:
  {{Counterfactual Analysis}} of {{Saliency Maps}} for {{Deep Reinforcement
  Learning}},'' in \emph{ICLR}, 2020.

\bibitem[Rudin and Carlson(2019)]{rudin_secrets_2019}
C.~Rudin and D.~Carlson, ``The {{Secrets}} of {{Machine Learning}}: {{Ten
  Things You Wish You Had Known Earlier}} to {{Be More Effective}} at {{Data
  Analysis}},'' in \emph{Operations {{Research}} \& {{Management Science}} in
  the {{Age}} of {{Analytics}}}, 2019, pp. 44--72.

\bibitem[Mania et~al.(2018)Mania, Guy, and Recht]{mania_simple_2018}
H.~Mania, A.~Guy, and B.~Recht, ``Simple random search of static linear
  policies is competitive for reinforcement learning,'' in \emph{NeurIPS},
  2018.

\bibitem[Gulwani et~al.(2017)Gulwani, Polozov, and Singh]{Gulwani_program_2017}
S.~Gulwani, O.~Polozov, and R.~Singh, ``Program {{Synthesis}},''
  \emph{Foundations and Trends in Programming Languages}, vol.~4, no. 1-2, pp.
  1--119, 2017.

\bibitem[Wiener(1954)]{wiener_human_1954}
N.~Wiener, \emph{The {{Human Use}} of {{Human Beings}}}, 1954.

\bibitem[Ananny and Crawford(2018)]{ananny_seeing_2018}
M.~Ananny and K.~Crawford, ``Seeing without knowing: {{Limitations}} of the
  transparency ideal and its application to algorithmic accountability,''
  \emph{New Media and Society}, vol.~20, no.~3, 2018.

\bibitem[Raji et~al.(2020)Raji, Smart, White, Mitchell, Gebru, Hutchinson,
  {Smith-Loud}, Theron, and Barnes]{raji_closing_2020}
I.~D. Raji, A.~Smart, R.~N. White, M.~Mitchell, T.~Gebru, B.~Hutchinson,
  J.~{Smith-Loud}, D.~Theron, and P.~Barnes, ``Closing the {{AI Accountability
  Gap}}: {{Defining}} an {{End}}-to-{{End Framework}} for {{Internal
  Algorithmic Auditing}},'' \emph{arXiv:2001.00973}, 2020.

\bibitem[Daly et~al.(2019)Daly, Hagendorff, Li, Mann, Marda, Wagner, Wang, and
  Witteborn]{daly_artificial_2019}
A.~Daly, T.~Hagendorff, H.~Li, M.~Mann, V.~Marda, B.~Wagner, W.~W. Wang, and
  S.~Witteborn, ``Artificial {{Intelligence}}, {{Governance}} and {{Ethics}}:
  {{Global Perspectives}},'' {{SSRN Scholarly Paper}}, 2019.

\end{thebibliography}
\end{document}